\PassOptionsToPackage{table}{xcolor}
\documentclass[10pt]{article}

\usepackage[preprint]{tmlr}

\makeatletter
\newif\ifdeanonymized
\if@accepted\deanonymizedtrue\else\deanonymizedfalse\fi
\makeatother

\usepackage[utf8]{inputenc} 
\usepackage[T1]{fontenc}    
\definecolor{linkpurple}{RGB}{128,64,170} 
\usepackage[colorlinks=true,allcolors=linkpurple,backref=page,bookmarksnumbered]{hyperref}       
\renewcommand*{\backref}[1]{}
\renewcommand*{\backrefalt}[4]{\ifcase #1\or Cited on page #2.\else Cited on pages #2.\fi}
\hypersetup{pdftitle={Programs-of-Layers in LLMs through the Lens of Cortical Areas}}
\ifdeanonymized\hypersetup{pdfauthor={Justus Westerhoff, Stephan Olbrich, Hatem Oraby, Walter Senn, Matthew Evan Larkum, Felix Alexander Gers}}\fi
\usepackage{etoc}           
\usepackage{url}            
\usepackage{graphicx}       
\usepackage{subcaption}     
\usepackage{placeins}       
\usepackage{booktabs}       
\usepackage{multirow}       
\usepackage{amsfonts}       
\usepackage{amsmath}        
\usepackage{amssymb}        
\usepackage{bbm}            
\usepackage{nicefrac}       
\usepackage{microtype}      
\usepackage{xcolor}         
\usepackage{xspace}         
\usepackage{glossaries}     
\glsdisablehyper            
\usepackage{cleveref}       
\usepackage{algorithm}      
\usepackage{algpseudocode}  
\algrenewcommand{\algorithmiccomment}[1]{\hfill{\footnotesize $\triangleright$ #1}}
\usepackage{tikz}           
\usetikzlibrary{arrows.meta, positioning, calc, patterns}
\newcommand{\legswatch}[1]{\tikz[baseline=-0.42ex]{\fill[#1] (0,0) rectangle (0.85em,0.68em);}}

\newcommand{\legenditem}[2]{\legswatch{#1}~#2}

\definecolor{polD1}{HTML}{2A78D6}
\definecolor{polD2}{HTML}{EB6834}
\definecolor{polD3}{HTML}{4A3AA7}
\definecolor{polD4}{HTML}{2F9366}
\definecolor{polD5}{HTML}{CC5169}
\definecolor{polBaseline}{HTML}{2F9366}
\definecolor{polRescued}{HTML}{0E8F86}
\definecolor{polUnsolved}{HTML}{CC5169}
\definecolor{polIdentityGray}{HTML}{898781}
\definecolor{polSkip}{HTML}{2A78D6}
\definecolor{polRepeat}{HTML}{EB6834}
\definecolor{polBoth}{HTML}{4A3AA7}
\definecolor{polDepthRed}{HTML}{D62728}
\definecolor{polOrigPath}{HTML}{FF7F0E}
\definecolor{polDepth90}{HTML}{DBE9F6}
\definecolor{polDepth95}{HTML}{B6D3EE}
\definecolor{polDepth100}{HTML}{8EBDE6}
\definecolor{polDepth105}{HTML}{5FA2DB}
\definecolor{polDepth110}{HTML}{2F7FC4}
\definecolor{polDepth115}{HTML}{1C5490}
\definecolor{polQwen38b}{HTML}{D62728}
\definecolor{polQwen257b}{HTML}{9467BD}
\definecolor{polQwen253b}{HTML}{2CA02C}
\definecolor{polQwen332b}{HTML}{1C5CAB}
\definecolor{polLlama32}{HTML}{17BECF}
\definecolor{polQwen15moe}{HTML}{C07D17}
\usepackage{tcolorbox}      
\usepackage{pifont}        
\usepackage{wasysym}       

\newcommand{\projectpage}{\url{https://datexis.github.io/RE-PoLar/}}
\newcommand{\repolar}{\textsc{RE-PoLar}} 

\newtcbox{\errtag}{on line, arc=2.5pt, colback=black!6, colframe=black!35,
  boxrule=0.4pt, boxsep=0.5pt, left=3pt, right=3pt, top=0.5pt, bottom=0.5pt,
  fontupper=\sffamily\footnotesize}

\definecolor{findingBack}{HTML}{F1EBFB}
\definecolor{findingFrame}{HTML}{8B5CF6}
\newtcbox{\ftag}{on line, arc=2.5pt, colback=findingBack, colframe=findingFrame,
  boxrule=0.5pt, boxsep=0.5pt, left=3pt, right=3pt, top=0.5pt, bottom=0.5pt,
  fontupper=\sffamily\footnotesize\bfseries, coltext=findingFrame!80!black}
\newcommand{\finding}[1]{\phantomsection\label{finding:#1}\ftag{F#1}}

\colorlet{polgreen}{green!45!black}
\colorlet{polred}{red!60!black}
\definecolor{polyellow}{HTML}{B8860B} 

\newcommand{\reprodmark}{\textcolor{polgreen}{\ding{51}}}
\newcommand{\notreprodmark}{\textcolor{polred}{\lightning}}
\newcommand{\newfindingmark}{\textcolor{polyellow}{$\bigstar$}}

\newcommand{\Base}{\textsc{Base}\xspace}
\newcommand{\Router}{\textsc{Router}\xspace}
\xspaceaddexceptions{(}

\newcommand{\dg}[1]{\textcolor{polgreen}{\fontsize{5}{6}\selectfont(#1)}}
\newcommand{\dr}[1]{\textcolor{polred}{\fontsize{5}{6}\selectfont(#1)}}

\newacronym{llm}{LLM}{large language model}
\newacronym{mcts}{MCTS}{Monte Carlo Tree Search}
\newacronym{lora}{LoRA}{Low-Rank Adaptation}
\newacronym{ood}{OOD}{Out-of-Distribution}
\newacronym{kv}{KV}{Key-Value}
\newacronym{mlp}{MLP}{Multi-Layer Perceptron}
\newacronym{ml}{ML}{machine learning}
\newacronym{kl}{KL}{Kullback--Leibler}
\newacronym{ucb}{UCB}{Upper Confidence Bound}
\newacronym{uct}{UCT}{Upper Confidence bounds applied to Trees}
\newacronym{dag}{DAG}{directed acyclic graph}

\newacronym{polar}{\textsc{PoLar}}{program-of-layers}

\title{Programs-of-Layers in LLMs through the Lens of Cortical Areas}

\author{\name Justus Westerhoff\thanks{Shared first authorship; order chosen at random.} \email justus.westerhoff@bht-berlin.de \\
      \addr Berliner Hochschule f\"ur Technik (BHT), Berlin, Germany
      \AND
      \name Stephan Olbrich\footnotemark[1] \email stephan.olbrich@bht-berlin.de \\
      \addr Berliner Hochschule f\"ur Technik (BHT), Berlin, Germany
      \AND
      \name Hatem Oraby \email hatem.oraby@hu-berlin.de \\
      \addr Humboldt-Universit\"at zu Berlin, Berlin, Germany
      \AND
      \name Walter Senn \email walter.senn@unibe.ch \\
      \addr University of Bern, Bern, Switzerland
      \AND
      \name Matthew Evan Larkum \email matthew.larkum@hu-berlin.de \\
      \addr Humboldt-Universit\"at zu Berlin, Berlin, Germany
      \AND
      \name Felix Alexander Gers \email felixalexander.gers@bht-berlin.de \\
      \addr Berliner Hochschule f\"ur Technik (BHT), Berlin, Germany}

\begin{document}

\maketitle

\etocdepthtag.toc{mainbody}
\begin{abstract}
Inference in \glspl{llm} is conventionally a fixed-depth, fixed-order forward pass through every layer, regardless of how difficult the input is.
The human brain does not work this way. Using the thalamus as a central hub, it routes information flexibly to all regions of the cortex according to demand.
\citet{li2026polar} recently showed, with a system they call \gls{polar}, that transformers can be given an analogous flexibility if their layers are treated as a library of functions rather than a fixed sequence.
Performance improves over the standard forward pass when each input is dynamically routed through an adaptive sequence of skipped or repeated contiguous layer blocks.
Reconstructing \gls{polar}'s diagnostic \gls{mcts} in more detail across 6 models, we reproduce several of its findings: skipping outperforms the standard pass, repeating outperforms skipping, and combining both outperforms either alone.
Shorter programs suffice for easier questions, while harder questions require more layer repeats.
For their main claim, the learned router's gain, we find what drives it.
Under their setup, the router gains by repairing the format of the model's answers, which a single fixed program, selected on validation questions, does better.
In our setup, with a prompt that already fixes the format, a larger data split and our search, its top prediction collapses to the standard pass on every model, although its top-$5$ programs together still gain.
Beyond reproduction, we find in our setup, where the format no longer fails, that a small fixed menu of programs solves far more questions than the standard pass,
and that programs which correct errors are brittle: undoing a single edit typically breaks the correction.
Connecting this to the brain's routing mechanisms, \gls{polar} mirrors principles of thalamo-cortical coordination between cortical-area-like transformer layers,
although a working thalamus analogue, a router that picks a better program per input, remains to be found.
We release our reproduction, \repolar, with code and data at
\ifdeanonymized
\projectpage.
\else
[\emph{removed for blind review}].
\fi
\end{abstract}

\glsresetall
\begin{figure}[t]
  \centering
  \includegraphics[width=0.95\linewidth]{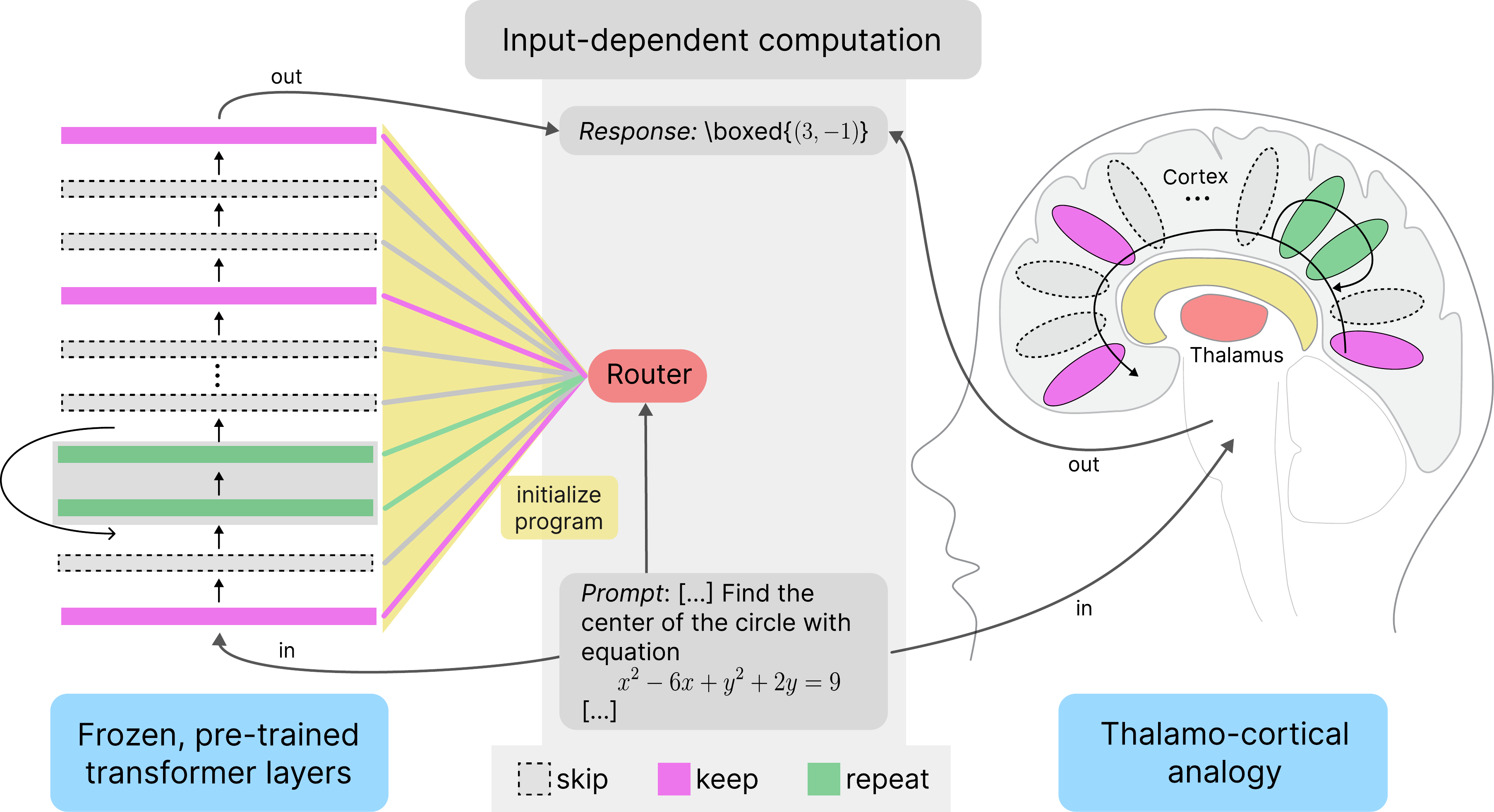}
  \caption{A \gls{polar}, predicted by a lightweight router,
  executes a schedule of \textsf{skip}, \textsf{keep}, and \textsf{repeat}
  operations over frozen, pretrained transformer layers (left).
  We read this coordinating role as \emph{functionally} analogous to the thalamus routing computation across cortical areas (right).
  A transformer's layer order is fixed only by training,
  not by necessity:
  a model \emph{trained} to run its layers in order can nonetheless be usefully \emph{executed} out of that order at inference time,
  a flexibility the brain already has by default.}
  \label{fig:overview}
\end{figure}

\section{Introduction}
\label{sec:intro}
\glsresetall

Most \glspl{llm} apply one fixed-depth architecture \citep{vaswani2017attention} to every input, regardless of how difficult it actually is.
By contrast, a conventional computer program is adaptive: program structure and complexity scale with the problem (e.g., through if-clauses and loops).
Similarly, the human brain, as a resource-constrained system, follows an adaptive principle, allocating different amounts and types of neural computation to different tasks.
This is reflected both in which brain circuits are recruited and in how long activity is sustained within them.
Difficult tasks can engage recurrent processing: activity reverberates within and between cortical circuits until sufficient evidence has accumulated \citep{wang2002,kar2019}.
Familiar tasks instead rely on faster, more automatic pathways \citep{Poldrack2005}.
While anatomy constrains where sensory information first arrives, it does not fix how that information is routed afterward: feedback loops within the cortex, and between the cortex and the thalamus, can amplify some pathways and suppress others depending on context and the brain's current behavioral state \citep{Munn2026}.

One axis of adaptivity \glspl{llm} do have at inference time is sequence length: a model can ``think longer'' by emitting more tokens \citep{wei2022chainofthought},
but the computation performed by any single forward pass through the network stays fixed regardless of how hard that step actually is.
This raises a basic question for transformer inference: \textbf{can a fixed forward pass be adapted to recover more of a model's latent capability, without retraining it?}

One family of answers changes the architecture and trains the model itself to be more recursive.
\citet{li2026polar} instead answer it without retraining the base model at all, treating the $D$ pretrained layers $f_0,\dots,f_{D-1}$ of an \gls{llm} as a library of functions rather than a fixed pipeline.
A per-input \gls{polar} $\pi=(i_1,\dots,i_K)$, where each $i_j$ is a layer index, selects a schedule of contiguous layer segments.
Each segment is \textsf{skipped}, \textsf{kept}, or \textsf{repeated}.
Their paper has two halves: a diagnostic \gls{mcts} over the \textsf{skip}/\textsf{repeat} program space, and the \gls{polar} predictor (a lightweight network that outputs a program in a single shot, replacing per-input search) itself.
The diagnostic search shows that better programs almost always exist, among other findings.
This paper's primary contribution is a careful reproduction of these findings alongside a public release of our search implementation and of all programs it discovers;
\citeauthor{li2026polar} release no search results.
On this basis, we build a detailed analysis and audit of the discovered programs,
together with a brain-inspired analogy for how such a program coordinates computation (\cref{fig:overview}).
\gls{polar} belongs to a broader family of cheap, training-free interventions that recover a model's latent capability without touching the weights (\cref{sec:related_work}).
Models can be ``under-elicited'': they compute a correct answer internally but fail to state it.
A \gls{polar} is one concrete way to recover such an answer, and the rest of this paper asks how reliably that recovery works and how recovered programs behave.
Concretely, this paper makes three contributions:
\begin{enumerate}
  \item We reproduce \citeauthor{li2026polar}'s diagnostic \gls{mcts} study and its router across 6 models with our own implementation of the search.\footnote{We implemented the search from their paper, which leaves it underspecified, before their search code was released.
  We keep ours for the main results because it searches each question on its own and more broadly (\cref{sec:appendix_mcts_design}).}
  We confirm the majority of \gls{polar}'s diagnostic findings (\cref{tab:findings_summary}).
  In our setup, with a prompt that keeps the answer format intact, the router's top prediction is the standard forward pass on every model, and no training recipe we tried makes it pick a better program per input (\cref{sec:router_collapse_results} and~\labelcref{sec:appendix_router_recipe}).
  We also run their released search on their data split and explain the router's gain there: it comes only from repairing the format of the model's answers, is smaller than that of a single fixed program, and depends strongly on the search seed (\cref{sec:appendix_their_split}).
  \item Building on our reproduction, we audit quantity and quality of the discovered programs.
  A small fixed menu of programs already solves most of what a far larger search finds, and under \citeauthor{li2026polar}'s setup, one fixed program applied to every input gains more than their router wherever the router gains.
  These programs are also fragile,
  since undoing a single edit inside a program that fixes a wrong answer breaks the fix more often than not.
  \item For the growing NeuroAI community \citep{zador2023neuroai}, we connect programs-of-layers to thalamo-cortical coordination,
  a reading that is new to our knowledge, and review it against the wider brain-inspired \gls{ml} literature.
\end{enumerate}
We publicly release our reproduction, \repolar, with code and the full set of \gls{mcts}-discovered programs for every question studied here, along with further visualizations such as an interactive explorer of the \gls{mcts} trees, at
\ifdeanonymized
\projectpage.
\else
[\emph{removed for blind review}].
\fi

\begin{table}[t]
\centering
\footnotesize
\caption{Every finding in \cref{sec:experiments_and_results}, checked against \citet{li2026polar}'s own study.
\reprodmark{} = reproduced,
\notreprodmark{} = not reproduced (we tested it and got a contradicting result),
\newfindingmark{} = a new finding of ours with no equivalent in the original paper to check against.
\ftag{F1}--\ftag{F4}
are \citeauthor{li2026polar}'s own diagnostic Findings 1--4.
\ftag{F4} holds only on the PoLar split with their search, likely a bias of that search (\cref{sec:appendix_their_split});
for \ftag{F8}, the router gains there only by repairing the output format.}
\label{tab:findings_summary}
\begin{tabular}{clc}
\toprule
Finding & Description & Status \\
\midrule
\hyperref[finding:0]{\ftag{F0}} & Reproducing \citeauthor{li2026polar}'s baseline numbers & \notreprodmark \\
\hyperref[finding:1]{\ftag{F1}} & Skip \& Repeat combined beats either alone & \reprodmark \\
\hyperref[finding:2]{\ftag{F2}} & Occam's razor: most samples have shorter valid programs than a standard forward pass & \reprodmark \\
\hyperref[finding:3]{\ftag{F3}} & Harder inputs need larger programs (test-time scaling via recurrence) & \reprodmark \\
\hyperref[finding:4]{\ftag{F4}} & \begin{tabular}[t]{@{}l@{}}Valid programs concentrate on short, mostly single-recurrence segments (short only on the\\ PoLar split, with their search)\end{tabular} & \notreprodmark \\
\hyperref[finding:5]{\ftag{F5}} & A small menu of programs generalizes across many questions & \newfindingmark \\
\hyperref[finding:6]{\ftag{F6}} & A wrong answer close to the ground truth is rescued ${\sim}1.5\times$ more often than a distant one & \newfindingmark \\
\hyperref[finding:7]{\ftag{F7}} & Programs are fragile: most edits are essential, and fragility depends on model, not difficulty & \newfindingmark \\
\hyperref[finding:8]{\ftag{F8}} & \begin{tabular}[t]{@{}l@{}}Router gives a pass@$1$ gain over \Base\ (on the PoLar split and also OOD only as format repair,\\and a single fixed program does better)\end{tabular} & \notreprodmark\,(\newfindingmark) \\
\hyperref[finding:9]{\ftag{F9}} & Router's pass@$5$ gain comes from a small, fixed menu, mainly input-independent & \newfindingmark \\
\hyperref[finding:10]{\ftag{F10}} & MMLU-Pro programs and router overfit to answer position, not content & \newfindingmark \\
\bottomrule
\end{tabular}
\end{table}

\section{Motivation}
\label{sec:motivation}
\glsresetall

\subsection{Biological Relevance}
Brain neurons inspired the earliest artificial neural networks \citep{McCulloch1943, Rosenblatt1958}, and neuroscience supplied foundational ideas for the architectures that followed \citep{Fukushima1980, Hopfield1982}.
The two fields have since diverged mostly for engineering reasons: hardware rewards architectures suited to dense matrix multiplication, regardless of biological plausibility \citep{hooker2021hardwarelottery}.
Yet artificial networks still repeatedly exhibit representations and computational motifs resembling biological ones \citep{Dabney2020, Guclu2015, Wang2021, Sussillo2015} (see \cref{sec:appendix_brain_ml} for further examples).

One particularly relevant difference is how computation is organized and routed.
Conventional transformers pass every token representation through a fixed sequence of layers in order,
but information clearly does not flow this way biologically.
Early routing is anatomical, fixed by the sensory modality.
Further processing then propagates through progressively higher-order cortical columns, with the thalamus routing information between them.
A cortical column, a vertically organized population of neurons, is often treated as a local processing block.
Its cell types and connectivity are remarkably conserved throughout the cortex, much as transformer layers repeat a single form of local computation.
There is also evidence that cortical areas built from these columns realize attention-layer-like computation directly \citep{babu2026corticosubcortical} (\cref{sec:cortical_analogies} gives the mechanistic detail).
\citet{Larkum2013} argues that a column's architecture and its interaction with the thalamus let the brain integrate context into local computation and distribute the context-sensitive output around the cortex.
This parallels how contextual information modulates attention over token embeddings across transformer blocks.
How the brain selects which information reaches which columns is still under investigation \citep{Munn2026,Zolnik2026}.
But cortical computation clearly executes in parallel across columns, with only a relevant subset active at any time.

\citet{li2026polar}'s \gls{polar} caught our attention for exactly this reason.
Their \textsf{skip} and \textsf{repeat} operations let a \gls{llm} omit or revisit entire layers at inference time without retraining, a complementary form of conditional computation.
Specifically, \gls{polar} stood out because its predictor is meant to play a \emph{thalamus-like} coordinating role (\cref{fig:overview}): it decides which modules engage and how information flows between them, rather than computing locally itself.
We treat this as a hypothesis that \gls{polar} attempts to instantiate, and test in \cref{sec:experiments_and_results} whether the router fulfils it (see \cref{sec:cortical_analogies} for further analogy and limitations).
Recently, \citet{cox2026frontalbottleneck} report that extensive task practice remodels neural circuitry to escape a default, effortful ``frontal bottleneck''.
This is further evidence that a practiced, selectively routed path, not one fixed circuit, is the biological norm.
Parallel to biological computation under a constrained budget \citep{Gilad2018}, \citeauthor{li2026polar} find that harder inputs need increasingly long programs.

\subsection{Preliminary Evidence for Flexible Layer Execution}
\label{sec:motivation_layer_redundancy}

Before finding \citet{li2026polar}, we asked ourselves: do layers of
a pretrained \gls{llm} tolerate rerouting at inference time, or could such rerouting even improve it?
We were prompted by a blog report that \emph{repeating} a contiguous layer block
(re-running it immediately after itself, no retraining) can raise accuracy \citep{ng2026rys}.
Independently,
\citet{sun2024painters} measure cosine similarity between layer activations and find a
related pattern: early and final layers are distinct, while middle layers share a
common representation space, a split echoed again and again \citep{lad2025robustness}, most recently on
vision transformers \citep{jacobs2026block}.

\begin{figure}[h]
  \centering
  \begin{subfigure}[t]{0.48\textwidth}
    \centering
    \includegraphics[width=0.7\textwidth]{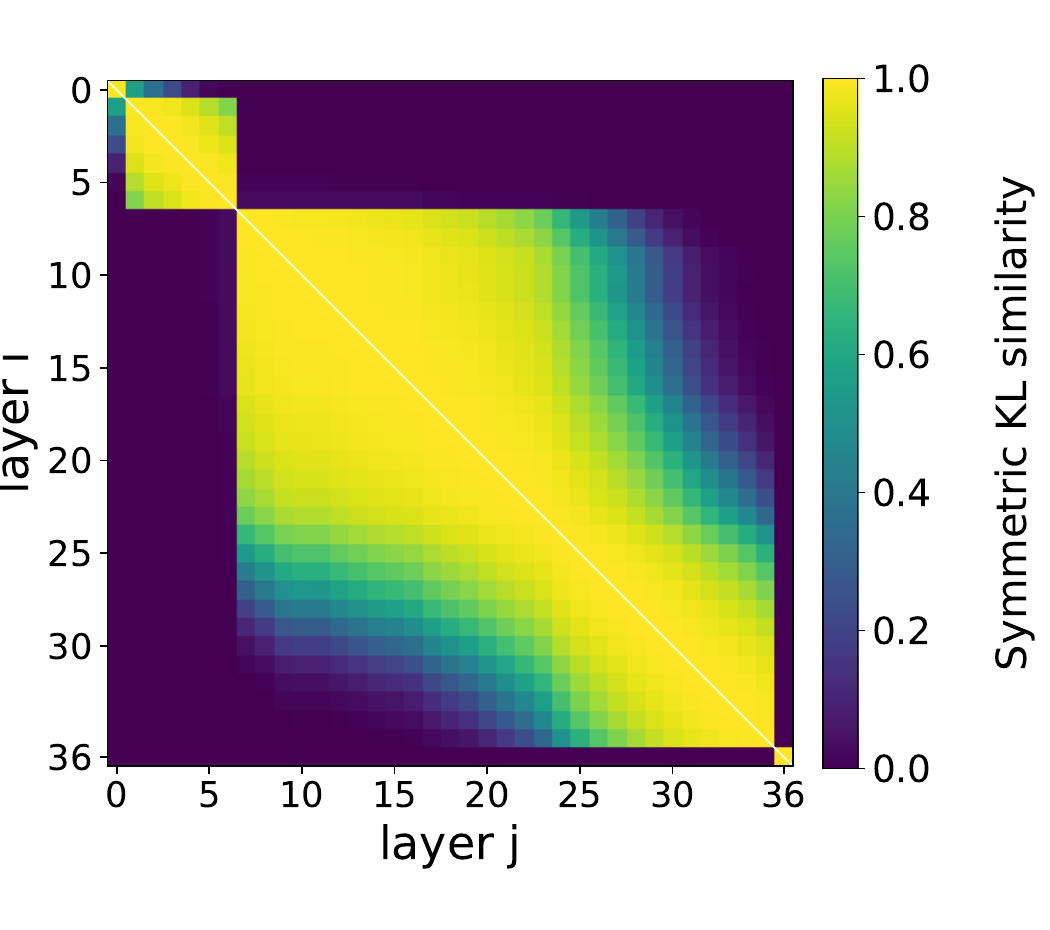}
    \caption{Symmetric \gls{kl} similarity between layer $i$'s and layer $j$'s
    output token distributions.}
    \label{fig:motivation-symkl}
  \end{subfigure}
  \hfill
  \begin{subfigure}[t]{0.48\textwidth}
    \centering
    \includegraphics[width=0.7\textwidth]{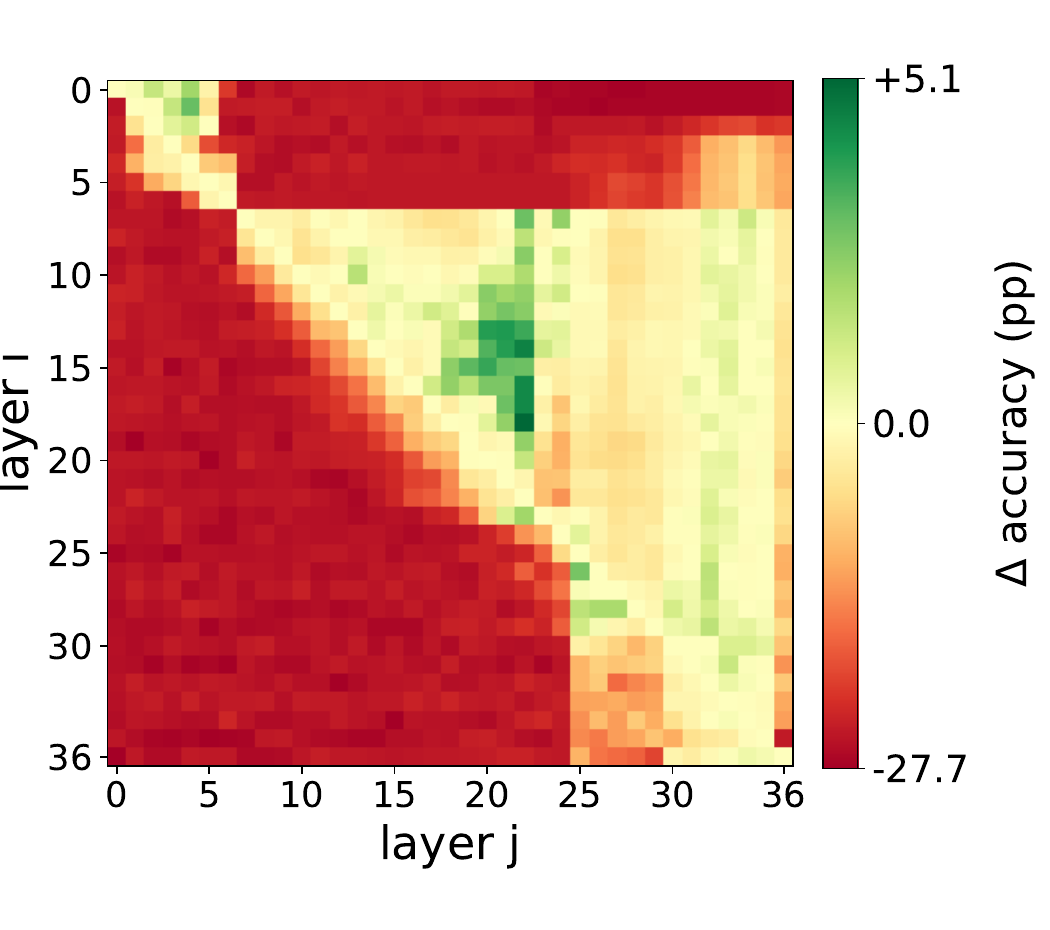}
    \caption{Accuracy $\Delta$ (pp) from re-running block $[i,j)$ a second time
    (repeat, upper triangle) or removing it (skip, lower triangle).}
    \label{fig:motivation-dupskip}
  \end{subfigure}
  \caption{Simply repeating a well-chosen middle block (layers 18--22), no retraining, no other
  intervention, already improves Qwen3-8B's accuracy on a 6-domain subset of MMLU-Pro
  \citep{wang2024mmlupro}.
  The representational pattern in (a) and the behavioral
  pattern in (b) agree on two boundaries: layer 6, where mixing layers from before
  and after collapses accuracy under both repeat and skip, and layer 25, past which
  layers grow mutually redundant, so skipping there turns from harmful to roughly
  neutral.}
  \label{fig:motivation}
\end{figure}

\Cref{fig:motivation-symkl,fig:motivation-dupskip} test this directly on Qwen3-8B, on a 6-domain subset of MMLU-Pro \citep{wang2024mmlupro}.
\Cref{fig:motivation-symkl} measures pairwise similarity across all 37 layer-states (the embedding output, plus the output of each of the 36 decoder layers) via symmetric \gls{kl} divergence.
\Cref{fig:motivation-dupskip} separately sweeps every contiguous layer block $[i,j)$ under \emph{repeat} (re-running a block immediately after itself) and \emph{skip} (removing it).
Layer 6 is the critical boundary in both.
Blocks that mix layers from before and after layer 6 collapse accuracy under both repeat and skip, matching a sharp similarity boundary at the same layer in (a).
The same pattern holds again past layer 25: layers there are mutually similar in (a) and can again be combined into a contiguous segment, so skipping them in (b) is tolerated, with a well-chosen middle block (best: layers 18--22) even net positive under repeat.
Overall, which layers can be combined into a contiguous skip or repeat segment in (b) closely follows which layers share a similar distribution in (a).

This single-model, single-benchmark result is correlational, but it already shows naive repeat/skip
interventions moving accuracy in a model-specific, structured way.
It is an informal, single-operator instance of the question \citeauthor{li2026polar}
formalize and that we reproduce and extend: given a library of layer functions, what
is the best per-input program over them, and can systematic search, or a lightweight
learned predictor in its place, recover the remaining gain?

\section{Methodology}
\label{sec:methodology}
\glsresetall

\subsection{Program-of-Layers Formalism}

A pretrained \gls{llm} computes its output for a sequence of length $T$ through a fixed sequence of $D$ transformer
layers.
Each layer is a function $f_i:\mathbb{R}^{T\times d}\to\mathbb{R}^{T\times d}$ with $i=0,\dots,D-1$, hidden dimension $d$.
A per-input \gls{polar}
$\pi=(i_1,\dots,i_K)$ replaces this fixed order with an
arbitrary finite index sequence, inducing the composition
$F_\pi = f_{i_K}\circ\cdots\circ f_{i_1}$.
We call $\pi$ \emph{valid} for a given input if
$F_\pi$ produces a correct prediction. Multiple distinct valid programs can exist for
the same input.  Simply applying $F_{\text{id}} := f_{D-1}\circ\cdots\circ f_0$ is the
standard forward pass, which we call \Base\ when reporting results.
We call $\pi$ a \emph{rescuing} program for a given input if $F_{\text{id}}$ is not valid but $F_\pi$ is.

An arbitrary index sequence $\pi$ of any length $K$ is intractable to search over directly, so \citet{li2026polar} restrict this space of possible programs to a small, structured subset.
The $D$ layers are
partitioned into contiguous segments by a boundary mask $\mathbf{z}^{\text{seg}}\in\{0,1\}^D$, where a 1 starts a new segment (each segment is at most $K_{\max}{=}4$ layers long).
Every segment is assigned one
operation from $\{\textsf{skip},\textsf{keep},\textsf{repeat}\}$ (label vector
$\mathbf{z}^{\text{op}}\in\{\textsf{skip},\textsf{keep},\textsf{repeat}\}^D$, read only at segment starts;
the other entries are ignored): \textsf{skip} removes the segment, \textsf{keep} runs it once
as usual, and \textsf{repeat} runs it twice.
\Cref{sec:reproduction_mcts_search} describes how \gls{mcts} explores this
operation set.
\Cref{sec:reproduction_router} describes how the router is trained to
predict it.

\subsection{Search Procedure through \gls{mcts}}
\label{sec:reproduction_mcts_search}

We describe \gls{polar}'s search procedure in more detail than either source paper \citep{li2025cola,li2026polar} does.
Both leave it substantially underspecified,
so we reconstructed it from the papers alone, before \citeauthor{li2026polar} released their search code, and specify previously unstated choices precisely below.
Their released search makes several choices that neither paper states, for example multi-layer edits only at the root (\cref{sec:appendix_their_split_search}).
\Cref{sec:appendix_mcts_design} justifies each of our choices and explains why our main results use our search rather than their released one.

\paragraph{Search nodes as edit sets.}
We depart from a literal action-by-action reading of
$\pi=(i_1,\dots,i_K)$ for tractability: a node is a \emph{set of edits} on top of the identity program
$\text{id}=(0,\dots,D-1)$.
An edit $e=(s,\ell,o,\rho)$ places operation $o\in\{\textsf{skip},\textsf{repeat}\}$ (\textsf{keep} is not an explicit edit, since every layer not covered by an edit is kept
by default) on
the contiguous block $[s,s{+}\ell)$, $\ell\le K_{\max}$, where $\rho\ge2$ is the block's total
execution count under \textsf{repeat};
we allow up to $\rho{=}5$, the bound \citet{li2025cola} state (\cref{sec:appendix_mcts_design}), although the router only expresses $\rho{=}2$ (\cref{sec:reproduction_router}).
A node $N$ is a set of pairwise non-overlapping edits, root $N_0=\emptyset$.
Its program
$\mathrm{Prog}(N)$ tiles every gap between edits with \textsf{keep} segments and is therefore always complete and directly executable.
Because the same edit set is reachable via more than one placement order, we canonicalize
nodes by their sorted edit tuple and cache them in a transposition table, which turns the search
graph into a \gls{dag} rather than a tree (see \cref{sec:appendix_mcts_design} for details).

\begin{figure}[h]
  \centering
  \includegraphics[width=0.95\linewidth]{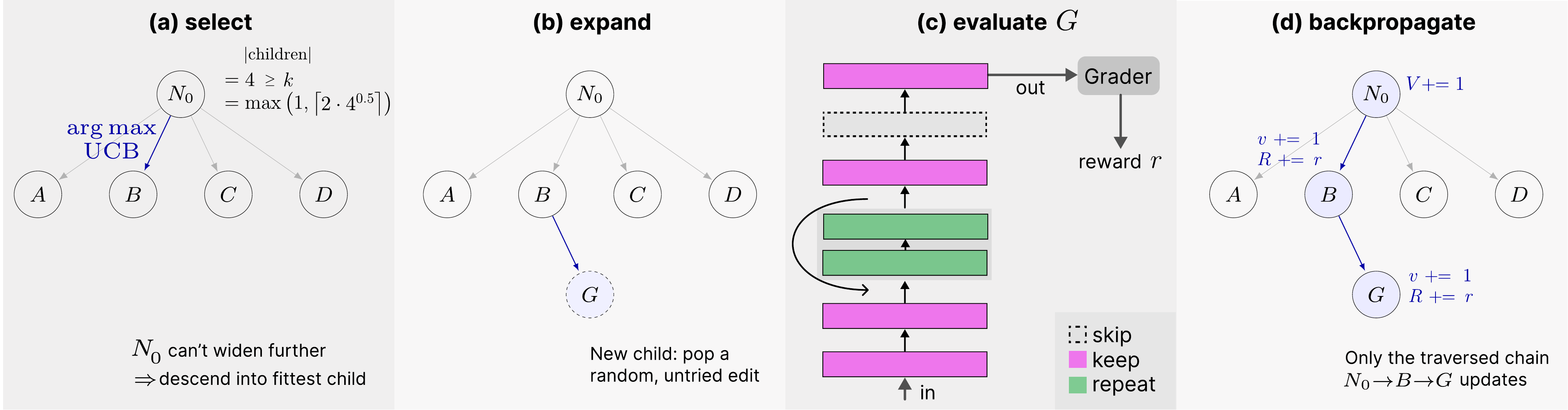}
  \caption{One simulation of \cref{alg:mcts}, traced end to end on a toy tree ($D{=}7$).
  (a) $N_0$ already has $k(4){=}4$ children, so selection descends by $\arg\max$ \gls{ucb} into child
  $B$ rather than expanding $N_0$ further.
  (b) $B$ has no child yet and may have $k(1){=}2$, so \textsc{SelectExpand} expands it directly, popping a
  random untried edit to create $G$. (c) The input $x$ is scored under $G$'s edit set
  ($\{\textsf{skip}[5,6),\,\textsf{repeat}[2,4)\!\times\!2\}$; layers counted from $0$ at the bottom, $\times2$ = executed twice), producing reward $r$.
  (d) $r$ backpropagates along the traversed chain $N_0\!\to\!B\!\to\!G$, and the global
  counter $V$ increments once.}
  \label{fig:mcts_diagram}
\end{figure}

\paragraph{Control flow.}
Each simulation descends from the root, choosing between two moves at every node it visits:
while the node still has room under its widening cap $k(v)$ (\cref{eq:ucb_widening}), search expands it directly with a new child;
once capped, search instead selects the child that currently maximizes its \gls{ucb} score (\cref{eq:ucb_widening}) and descends into it.
Unlike classical \gls{mcts}, there is no separate rollout phase: search scores the node it lands on directly, executing its program with exactly one model call, rather than simulating forward with a default policy until a terminal state.
Within a search round, the model calls of the different samples, each scoring its own program, are batched rather than evaluated one at a time (\cref{sec:appendix_masked_batch_kv}).
Reward is $1$ if the model's output is correct and $0$ otherwise.
That reward backs up along the exact chain of nodes just visited, and a global counter $V$ increments once per simulation, feeding every node's \gls{ucb} explore term equally.
\Cref{sec:appendix_mcts_design} gives the full pseudocode
(\cref{alg:mcts,alg:selectexpand}) and the
formal definitions of the expansion rule behind this summary.
\Cref{fig:mcts_diagram}
traces one simulation end to end on a toy tree.

\paragraph{Selection: \gls{ucb} score.}
We use the same \gls{ucb} equation as \citeauthor{li2026polar} to rank a capped node's children: each child's score trades off how well it has scored so far against how rarely it has been tried, so search neither fixates on one early success nor keeps re-visiting a branch it already understands well, with a further penalty against candidates that make the executed program longer:
\begin{equation}
\mathrm{UCB}(N) \;=\; \underbrace{\frac{R(N)}{v(N)}}_{\text{exploit}} \;+\; \underbrace{c\sqrt{\frac{\ln V}{v(N)}}}_{\text{explore}} \;-\; \underbrace{\lambda\,\frac{|\pi_N|}{D}}_{\text{length penalty}}, \qquad\qquad k(v) = \max\!\big(1,\ \lceil \alpha\, v^{\beta}\rceil\big),
\label{eq:ucb_widening}
\end{equation}
where $R(N)/v(N)$ is the cumulative reward over visits per node, $c$ the exploration factor, $V$ the total simulation count across the whole
search and
$|\pi_N|=D-\sum_{e\in N_{\text{skip}}}\ell_e+\sum_{e\in N_{\text{repeat}}}(\rho_e{-}1)\ell_e$ (with $N_{\text{skip}}, N_{\text{repeat}}\subseteq N$ its skip and repeat edits)
is the executed program length that, together with $\lambda$, penalizes long programs.
$k(v)$ is a progressive-widening cap absent from \citeauthor{li2026polar}: a node
may only be expanded with a new
child while $|\mathrm{children}(N)|<k(v(N))$, which forces the tree to explore programs with more than one edit rather than exhausting single-edit siblings of the root.

\subsection{Router: Architecture and Training-Target Construction}
\label{sec:reproduction_router}

\citet{li2026polar} use the programs found by per-input \gls{mcts} to train a single-shot predictor, which predicts a program for each question without search.
This router outputs a program
representation $(\mathbf{z}^{\text{seg}},\mathbf{z}^{\text{op}})$ given the input.
It first encodes the input with a \emph{frozen} embedding model (Qwen3-Embedding-0.6B
\citep{qwen3embedding}).
Learnable per-layer query embeddings then cross-attend to the resulting token representations,
giving one input-conditioned representation per layer.
These pass through a $2$-layer transformer encoder for global context
across depth.
Two linear heads produce segmentation logits ($\mathbb{R}^D$) and operation logits
($\mathbb{R}^{D\times3}$).
Training loss is segmentation binary cross-entropy plus a
masked operation cross-entropy (operation loss computed only at segment starts),
supervised directly on valid \gls{mcts}-discovered programs.
\Cref{sec:appendix_router_recipe} traces every architecture and training-data choice above to its source in \citeauthor{li2026polar}'s own released code.
At inference, per-layer segmentation probabilities are thresholded into segment
boundaries, and each resulting segment is capped at $K_{\max}$ layers.
A small beam
search over segment operations then ranks the surviving candidate programs before mapping the
winner to an executed program.
This router adds only $\approx\!2.1$M parameters ($0.01$--$0.06\%$ of the base model), and per \citeauthor{li2026polar}'s own reported figure, it costs $\approx\!0.8\%$ extra compute
over a single forward pass.
We reproduce this architecture, loss, and decoding procedure exactly as \citeauthor{li2026polar} describe and implement it (\cref{sec:appendix_router_recipe});
what their paper leaves open is how the search results become training data.

The router's training data comes from the \gls{mcts}.
Every valid program for a question becomes its own training example, capped at $50$ per question.
Segmentation and operation targets are read off the layers that each program $\mathrm{Prog}(N)$ executes (\cref{sec:reproduction_mcts_search}):
neighbouring layers with the same operation form one segment of at most $K_{\max}$ layers, so two adjacent \textsf{skip} edits become one segment, while two adjacent \textsf{repeat} blocks stay separate.
Unlike in the search, \textsf{keep} segments are capped at $K_{\max}$ too, matching \cref{sec:methodology}'s original per-segment bound.
This matches what the router's own decoding
enforces at inference, where a predicted \textsf{repeat} always decodes to exactly two executions, so any \gls{mcts}-discovered program whose repeat count differs from two is dropped from training entirely (\emph{Drop}), matching \citeauthor{li2026polar}'s own behavior.
Each example carries a scalar weight: $1.0$ by default, except a question's own identity program is down-weighted to $0.3$ whenever a strictly shorter valid program also exists for that question.
Every example is further scaled by $1/n_{\text{prog}}$ for its question, so questions with many discovered valid programs do not dominate the loss.
\Cref{sec:appendix_router_recipe} traces each of these choices to its source, most of them not stated in the original paper's text, and compares \emph{Drop} with two alternative treatments.

\section{Experiments and Results}
\label{sec:experiments_and_results}
\label{sec:reproduction}
\label{sec:analysis}
\glsresetall
This section reports our \gls{mcts} and router reproduction of \citet{li2026polar} (\cref{sec:study_setup}), and checks that reproduction against their own diagnostic findings and their main router claim (\cref{sec:analysis_polar_recap,sec:router_collapse_results}).
It then extends the reproduction with program-level analyses of our own that have no equivalent in \citeauthor{li2026polar} (\cref{sec:analysis_beyond_polar,sec:mmlu_pro_track}).
\Cref{tab:findings_summary} lists our findings \ftag{F0}--\ftag{F10}, checked against their study.
Unless stated otherwise, results in this section use our search on our split;
\cref{sec:appendix_their_split} repeats the router and \ftag{F1}--\ftag{F4} with their released search on their split.

\subsection{Experimental Setup}
\label{sec:study_setup}
\textbf{Models.} We evaluate 6 pretrained, instruction-tuned \glspl{llm}: Qwen1.5-MoE-A2.7B-Chat, Qwen2.5-3B-Instruct, Qwen2.5-7B-Instruct, Qwen3-8B, Qwen3-32B, and LLaMA-3.2-3B-Instruct.
All 6 are used fully frozen throughout this paper. \citeauthor{li2026polar}'s own study covers four of those, and Qwen2.5-7B-Instruct and Qwen3-32B are our own additions.

\textbf{Data.} In-distribution: DART-Math \citep{tong2024dartmath}, split into five difficulty levels DM-1 (easiest) through DM-5 (hardest) (\cref{sec:appendix_dart_math} explains the construction), $2{,}000$ queries per level ($1{,}250$ train / $250$ val / $500$ test), $10{,}000$ total.
Out-of-distribution: MMLU-Pro \citep{wang2024mmlupro} ($n{=}12{,}032$, multiple-choice, up to $10$ lettered options per question), used as a zero-shot transfer target for \Base\ and for the routers trained on the PoLar split (\cref{sec:appendix_baseline_full,sec:appendix_their_split_ood}) and, separately, as its own in-domain \gls{mcts}/router track (\cref{sec:mmlu_pro_track}).
We also use two arithmetic transfer targets, ASDiv \citep{miao2020asdiv} ($n{=}2{,}305$) and MAWPS \citep{kadlcik2023mawps} ($n{=}520$).

\textbf{Metrics.} \Base\ ($\tau{=}0$) is the unmodified model under greedy decoding.
\Base\ (sampling) draws $k$ stochastic decodes at temperatures $\tau\in\{0.3,0.7,1.0\}$ and reports the best result across temperatures for each $k$, as \citeauthor{li2026polar} do.
Pass@$k$ is the probability that at least one of $k$ candidates is correct:
for \Base\ (sampling) the candidates are $k$ stochastic decodes, for the router its top-$k$ beam-search programs.
The fixed program is one program per difficulty, applied to every test question: the best, on the validation questions, of the $20$ non-identity programs with the highest (smoothed) success rate in the training questions' \gls{mcts} data.
Every search runs three times with different seeds, and we report the mean and standard deviation over the three runs.

\textbf{Evaluation protocol.}
\label{sec:eval_protocol}
We keep their generation cap of $50$ tokens.
On our split, we use the \texttt{oneshot} prompt instead of their \texttt{polar\_default} prompt, and no chat template, which fixes two failure modes of their setup (\cref{sec:appendix_placeholder_echo,sec:appendix_chat_template}).
ASDiv and MAWPS use the same protocol;
MMLU-Pro is scored as forced-choice over answer-letter logits, with no generation step (\cref{sec:appendix_baseline_full}).
We report hardware and GPU-hours by pipeline stage (\cref{sec:appendix_hardware}).

\finding{0}\ \notreprodmark
\label{sec:baseline_reproduction}
On our reconstruction of their test set, with their prompt and without the chat template, \Base\ matches \citeauthor{li2026polar}'s published numbers for LLaMA-3.2-3B, on some levels for Qwen1.5-MoE-A2.7B and Qwen3-8B, and on none for Qwen2.5-3B (\cref{tab:their_split_base}).
Our reconstruction of their split may not match it exactly (\cref{sec:appendix_their_split_setup}).
Their evaluation code applies a chat template, but with it, \Base\ is below their numbers in every cell for the three Qwen models (\cref{sec:appendix_their_split_base});
we therefore deliberately run their split without it, as our own setup does (\cref{sec:appendix_chat_template}).

\subsection{Reproducing the \gls{mcts} Diagnostic Findings}
\label{sec:analysis_polar_recap}

\citet{li2026polar}'s own diagnostic \gls{mcts} study
already establishes what a frozen model's layers can do beyond the standard forward pass.
What follows checks each of those findings against our own \gls{mcts}
reproduction (6 models, DART-Math Diff 1--5), and reports what we find beyond them. \Cref{sec:appendix_analysis_extra} provides additional results;
\cref{sec:appendix_their_split_findings} repeats \ftag{F1}--\ftag{F4} with their search on their split.

\begin{figure}[b] 
\centering
\begin{subfigure}{0.3440\linewidth}\centering
\includegraphics[width=1.0000\linewidth]{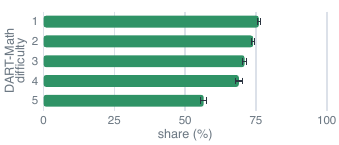}
\caption{Share admitting a shorter program.}
\end{subfigure}%
\hspace{73.7pt}%
\begin{subfigure}{0.3502\linewidth}\centering
\includegraphics[width=1.0000\linewidth]{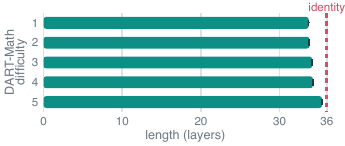}
\caption{Shortest valid program's length.}
\end{subfigure}
\caption{\hyperref[finding:2]{\ftag{F2}}\ On Qwen3-8B, most solved questions at every difficulty have a valid program shorter than the standard forward pass.
By DART-Math difficulty (1 easiest, 5 hardest):
(a) the share of solved questions (those with at least one valid program) with a program shorter than the standard forward pass; (b) the mean length of the
shortest valid program. The dotted line is the standard forward pass length $D$.
Bars are the mean over the three search runs, whiskers are $\pm$ 1 std.
\Cref{fig:appendix_shorteridentity_diff_rate,fig:appendix_shorteridentity_diff_len} in the appendix show all 6 models.}
\label{fig:main_shorter_qwen3_8b}
\end{figure}

\finding{1}\ \reprodmark\ \textsf{skip} and \textsf{repeat} combined beat either alone, on every model and every
DART-Math difficulty level, with \textsf{repeat} the stronger of the two mechanisms on its own.
This is the same ordering \citeauthor{li2026polar} report.
\Cref{tab:analysis_table1} reports \Base $<$ Skip-only $<$ Repeat-only $<$
Skip\&Repeat accuracy for each of our 6 models at every DART-Math difficulty level.
The Skip\&Repeat gain over \Base is positive everywhere.

\finding{2}\ \reprodmark\ (Occam's razor). On 5 of our 6 models, most solved questions have a valid program shorter than the standard
forward pass ($50.5$--$91\%$ per difficulty), confirming \citeauthor{li2026polar}'s finding;
on Qwen1.5-MoE-A2.7B, fewer than half do ($39$--$45\%$).
\Cref{fig:main_shorter_qwen3_8b} shows Qwen3-8B,
\cref{fig:appendix_shorteridentity_diff_rate,fig:appendix_shorteridentity_diff_len} all 6 models.

\begin{figure}[h]
\centering
\footnotesize
\rule[0.4ex]{1.6em}{1.4pt}\,recurrence\qquad\rule[0.4ex]{0.5em}{1.4pt}\,\rule[0.4ex]{0.5em}{1.4pt}\,\rule[0.4ex]{0.5em}{1.4pt}\,skip\\[0.3em]
\normalsize
\begin{subfigure}{0.1996\linewidth}\centering
\includegraphics[width=0.9421\linewidth]{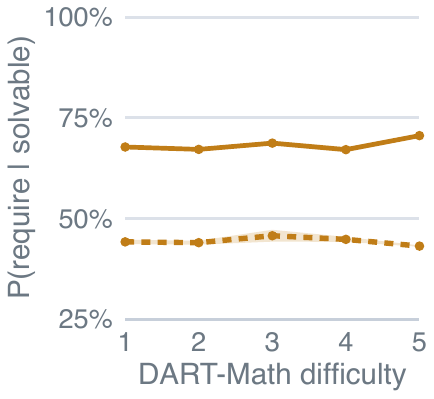}
\caption{\scriptsize\mbox{Qwen1.5-MoE-A2.7B}}\end{subfigure}%
\hspace{4.4pt}%
\begin{subfigure}{0.1384\linewidth}\centering
\includegraphics[width=1.0000\linewidth]{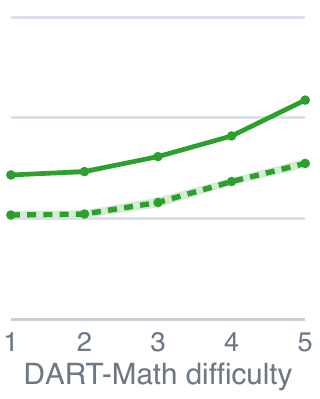}
\caption{\scriptsize\mbox{Qwen2.5-3B}}\end{subfigure}%
\hspace{7.1pt}%
\begin{subfigure}{0.1384\linewidth}\centering
\includegraphics[width=1.0000\linewidth]{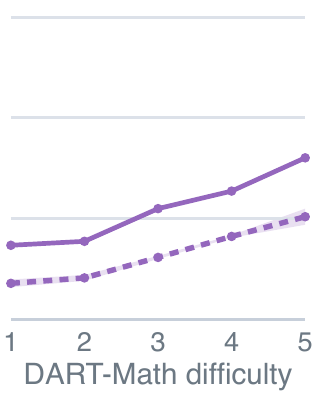}
\caption{\scriptsize\mbox{Qwen2.5-7B}}\end{subfigure}%
\hspace{7.1pt}%
\begin{subfigure}{0.1384\linewidth}\centering
\includegraphics[width=1.0000\linewidth]{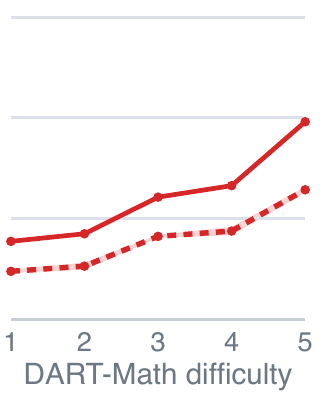}
\caption{\scriptsize\mbox{Qwen3-8B}}\end{subfigure}%
\hspace{7.1pt}%
\begin{subfigure}{0.1384\linewidth}\centering
\includegraphics[width=1.0000\linewidth]{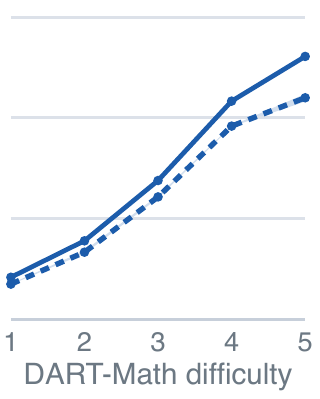}
\caption{\scriptsize\mbox{Qwen3-32B}}\end{subfigure}%
\hspace{3.8pt}%
\begin{subfigure}{0.1526\linewidth}\centering
\includegraphics[width=0.9074\linewidth]{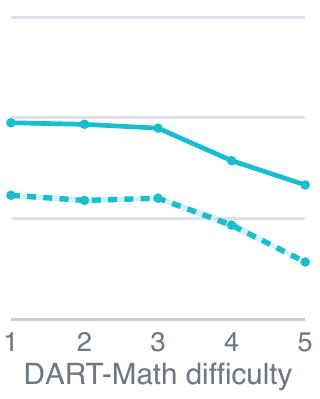}
\caption{\scriptsize\mbox{LLaMA-3.2-3B}}\end{subfigure}
\vspace{-4pt}
\caption{\hyperref[finding:3]{\ftag{F3}}\ On 5 of 6 models, harder questions need recurrence more often, and on 4 also skipping.
Of every question that is solvable at all, the share that needed recurrence
(solid) or needed skipping (dashed) to be solved. Lines are the mean over the three search runs, bands are $\pm$ 1 std.}
\label{fig:appendix_fig5b}
\end{figure}

\begin{figure}[h]
\centering
\begin{subfigure}{0.4910\linewidth}\centering
\includegraphics[width=1.0000\linewidth]{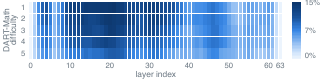}
\caption{Skip.}\end{subfigure}%
\hspace{6.7pt}%
\begin{subfigure}{0.4552\linewidth}\centering
\includegraphics[width=1.0000\linewidth]{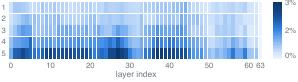}
\caption{Repeat.}\end{subfigure}
\caption{\hyperref[finding:3]{\ftag{F3}}\ On Qwen3-32B, easier difficulties concentrate on skipped layers, while harder ones spread repeats across more of the stack.
Share of shortest-valid programs that \textsf{skip} (a) or \textsf{repeat} (b) each
layer, by layer index and DART-Math difficulty.
Each cell is the mean over the three search runs; each panel has its own color scale. Same measurement as \cref{fig:appendix_layerops_skip,fig:appendix_layerops_repeat}, all 6 models, in the appendix.}
\label{fig:main_layerops_qwen3_32b}
\end{figure}

\finding{3}\ \reprodmark\ Harder inputs rely more on the \textsf{skip}/\textsf{repeat} program space, and allowing more
test-time recurrence steps gives access to more valid programs, confirming \citeauthor{li2026polar}'s
finding.
\Cref{fig:analysis_fig5a,fig:appendix_fig3} confirm the recurrence-budget part directly: on every model, the share
of questions with at least one valid program rises monotonically as more recurrence
steps are allowed.
\Cref{fig:appendix_fig5b} confirms the difficulty part: the share of solvable questions that
specifically need recurrence rises from the easiest to the hardest difficulty on 5 of 6
models, and so does the share that needs skipping on 4 of them;
on Qwen1.5-MoE-A2.7B it stays about flat ($-1.0$pp).
The exception on both is LLaMA-3.2-3B:
on the PoLar split its shares stay about flat, and on ours they fall, mostly because of the question mix (\cref{sec:appendix_llama_levels}).
\Cref{fig:main_layerops_qwen3_32b} makes the same pattern concrete at the layer level for
Qwen3-32B: easier questions solve with skips concentrated in a few positions, while
harder questions increasingly rely on repeats spread across all layers.
Some layer positions are consistently preferred for an edit and others consistently avoided
across all 6 models (\cref{fig:appendix_layerops_skip,fig:appendix_layerops_repeat} in
the appendix).
\Cref{sec:motivation_layer_redundancy} finds the same avoided region
independently for Qwen3-8B (layer 6), and the Qwen3-8B panel of \cref{fig:appendix_layerops_skip} shows it as an early band that edits rarely touch.

\finding{4}\ \notreprodmark\ Valid programs' segments are as short as \citeauthor{li2026polar} find only under their setup.
They report that $57.7\%$ of segments in valid programs touch a single layer and over two-thirds at most two.
With our search on our split, 5 of our 6 models have fewer segments of both kinds ($21$--$38\%$ single-layer, \cref{tab:analysis_f4}, shortest-valid-program segments pooled over all difficulties),
and even the closest model, Qwen1.5-MoE-A2.7B, only partially matches their numbers.
\Cref{fig:f4_seglen} below gives the full segment-length distribution behind
these shares.
The other sub-claim, that segments are repeated at most once, holds on all 6 models for the shortest valid program:
in $74$--$86\%$ of solved questions, even its most-repeated segment is repeated at most once (\cref{fig:f4_maxrepeat}),
and in $45$--$72\%$ no segment is repeated at all.
Among the programs that do repeat, the most-repeated segment is repeated exactly once in only $49$--$61\%$.
The larger Qwen models repeat least (Qwen3-32B: $86\%$ at most once), Qwen1.5-MoE-A2.7B and LLaMA-3.2-3B most ($74$ and $77\%$).
\citeauthor{li2026polar} do not state which valid programs they count.
Over all valid programs, segment lengths barely change ($30$--$39\%$ single-layer), and most edit segments are still repeated at most once ($57$--$67\%$),
but only about half of the repeated segments are ($38$--$54\%$).
With their search on the PoLar split, all $4$ models we ran it on have more short segments ($70$--$81\%$ single-layer), likely because their search makes multi-layer moves only at the root (\cref{sec:appendix_their_split_findings}).

\begin{figure}[h]
\centering
\footnotesize
\legenditem{polQwen15moe}{Qwen1.5-MoE-A2.7B}\quad\legenditem{polQwen253b}{Qwen2.5-3B}\quad
\legenditem{polQwen257b}{Qwen2.5-7B}\quad\legenditem{polQwen38b}{Qwen3-8B}\quad
\legenditem{polQwen332b}{Qwen3-32B}\quad\legenditem{polLlama32}{LLaMA-3.2-3B}\\[0.3em]
\normalsize
\begin{subfigure}[t]{0.46\linewidth}\centering
\includegraphics[width=\linewidth]{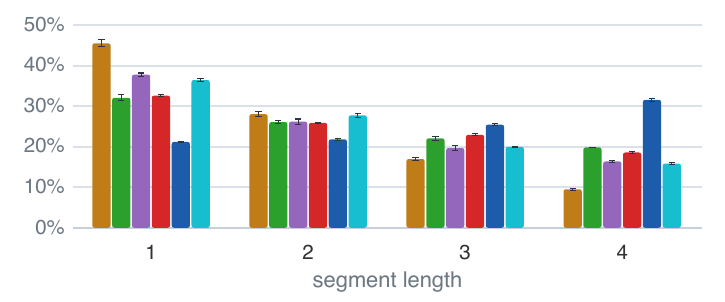}
\caption{Segment length.}
\label{fig:f4_seglen}
\end{subfigure}%
\hfill
\begin{subfigure}[t]{0.46\linewidth}\centering
\includegraphics[width=\linewidth]{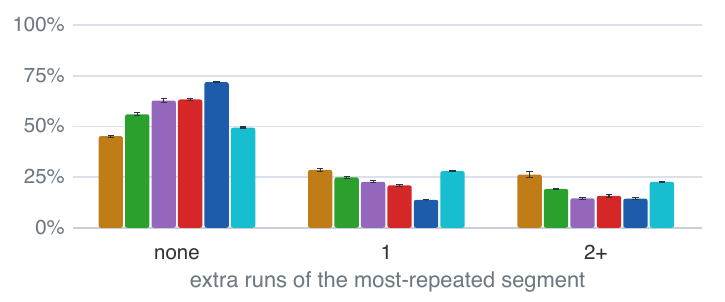}
\caption{Most-repeated segment.}
\label{fig:f4_maxrepeat}
\end{subfigure}
\caption{\hyperref[finding:4]{\ftag{F4}}\ On 5 of 6 models, most edit segments are longer than one layer, but segments are rarely repeated more than once.
(a) Length of the edit segments and (b) how many extra times the most-repeated segment runs (none: no segment is repeated), in the shortest valid program per solved question, pooled over all difficulties.
Bars are the mean over the three search runs, whiskers $\pm 1$ std.}
\label{fig:f4_segments}
\end{figure}

\subsection{Beyond Reproduction: Menu Generalization, Error Analysis, and Program Robustness}
\label{sec:analysis_beyond_polar}

\begin{figure}[h]
\centering
\begin{subfigure}[t]{0.4304\linewidth}\centering
\includegraphics[width=1.0000\linewidth]{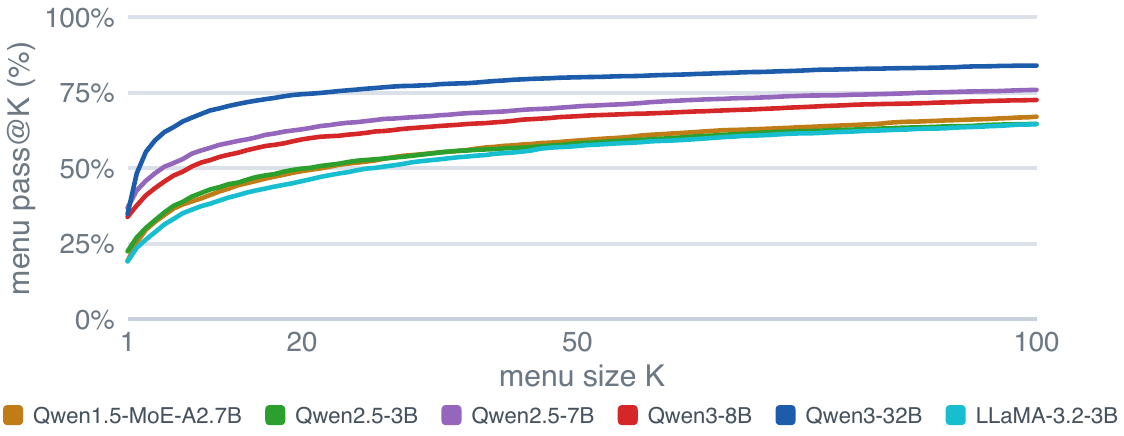}
\caption{Menu pass@$K$.}
\end{subfigure}%
\hspace{26.0pt}%
\begin{subfigure}[t]{0.4057\linewidth}\centering
\includegraphics[width=1.0000\linewidth]{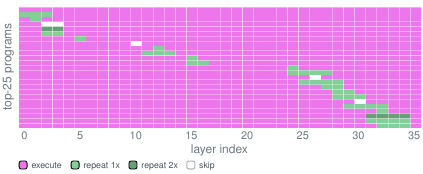}
\caption{Top-25 programs, Qwen3-8B, DM-2.}
\end{subfigure}
\caption{\hyperref[finding:5]{\ftag{F5}}\ A small fixed menu of programs solves far more questions than identity alone ($2.6\times$ at $K{=}100$), and its top programs edit many different layer blocks.
(a) Menu pass@$K$ against menu size $K$, pooled over difficulty, one line per model.
(b) The layers edited by the top-$25$ programs of Qwen3-8B on DM-2, one row per program, sorted by first edited layer, identity first.
One search run (seed 42).}
\label{fig:menu_pooled_top25}
\end{figure}

\finding{5}\ \newfindingmark\ A small, fixed menu of programs solves far more questions than identity alone:
the same handful of programs generalizes across many different questions, rather than
every question needing its own bespoke program.
For each model and difficulty level, we rank every program, identity included, by how many questions it solved during search,
then run the top-$K$ of those programs against every question of that level (where the search already tried a program on a question, we reuse its result).
\Cref{fig:menu_pooled_top25}(a) pools the menu's pass@$K$ (the share of questions that at least one of its $K$ programs solves) across all 6 models as $K$ grows
from $1$ to $100$ (\cref{fig:menu_topk} gives the full per-model/per-difficulty curves).
At $K{=}100$, the menu solves on average $72\%$ of the questions, against $28\%$ for identity (about $2.6\times$).
Ranking the programs on the $1{,}500$ train and validation questions and running the menu only on the $500$ test questions gives almost the same result ($71\%$ against $28\%$), so the menu is not fitted to the questions it was ranked on.
With the same budget as \Base\ (sampling) at $k{=}5$, the held-out menu's five best programs already beat five stochastic decodes on every model and level, on the same test questions ($43\%$ against $37\%$ on average, \cref{tab:main_results}).
Apart from Qwen3-32B, whose answers are often cut off at $50$ tokens (\cref{sec:appendix_q32b_budget}), no single edit solves clearly more questions than identity, but each helps a different slice of questions.
\Cref{fig:menu_pooled_top25}(b) shows the top-25 programs themselves are structurally diverse, editing many different layer bands rather than converging on one recipe: each edits exactly one contiguous layer block, with edits clustering away from an early band around layer 6 and a later one around layer 25, the same two boundaries as in \cref{fig:motivation}.
The fixed program (\cref{tab:main_results}) is a menu of size one, chosen on validation questions.

\finding{6}\ \newfindingmark\ It matters a lot how close a wrong answer already is to the ground truth:
a \emph{related} wrong answer (numerically or structurally close to ground truth) is
rescued by some valid program roughly $1.5\times$ more often than an \emph{unrelated}
one, in every model.
We reach this using \texttt{MiniMax-M2.7-AWQ} as an \gls{llm}-assisted classifier for how close each wrong answer is to the ground truth, applied to every DART-Math question the identity program answers wrong, across all 6 models and 5 difficulties
(\cref{sec:appendix_bht_large} gives the full taxonomy, methodology, and
error-category/closeness-rescue tables).
Together with the format repair behind the router's gain on the PoLar split (\hyperref[finding:8]{\ftag{F8}}), this suggests that programs mostly make a small adjustment to an answer the model has nearly reached, rather than new reasoning.

\begin{figure}[h]
\centering
\begin{minipage}[c]{0.5029\linewidth}\centering
\includegraphics[width=\linewidth]{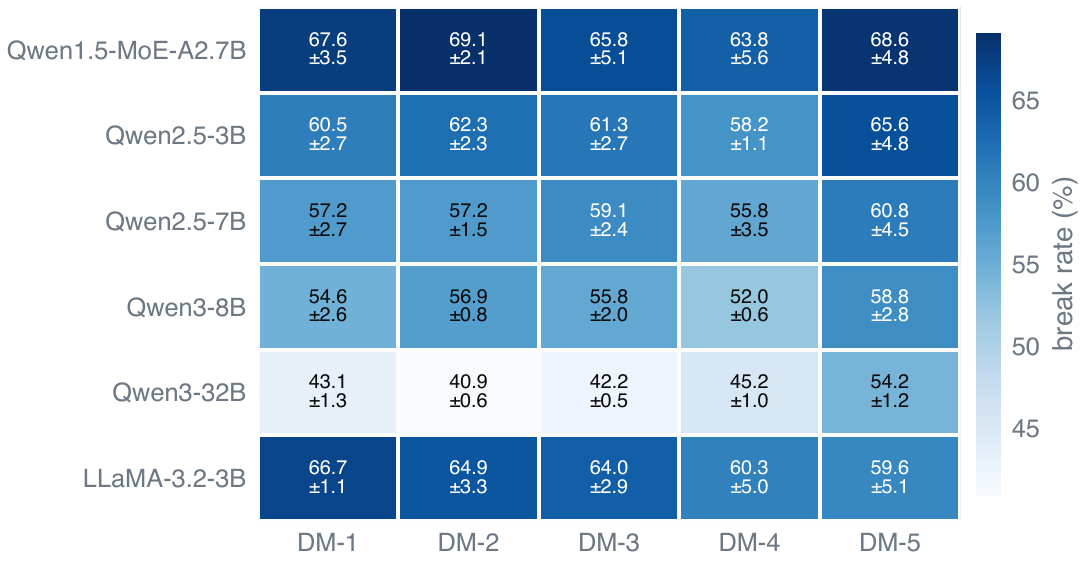}
\captionof{figure}{\hyperref[finding:7]{\ftag{F7}}\ $45$--$67\%$ of the edits in multi-edit rescuing programs are essential: undoing one breaks the program, depending on the model rather than the difficulty.
Each cell is the break rate over all single-segment reversions to \textsf{keep} in rescuing programs (identity wrong, this program right) with at least two edited segments. Mean $\pm$ std over the three search runs.}
\label{fig:op_transition_robustness}
\end{minipage}
\hfill
\begin{minipage}[c]{0.46\linewidth}\centering
\resizebox{\linewidth}{!}{%
\begin{tabular}{lr@{\,}c@{\,}rr@{\,}c@{\,}lr@{\,}c@{\,}l}
\toprule
Model & \multicolumn{3}{c}{Examples} & \multicolumn{3}{c}{Identity \%} & \multicolumn{3}{c}{Keep \%} \\
\midrule
Qwen1.5-MoE-A2.7B & $3{,}401$ & {\scriptsize$\pm$} & {\scriptsize 56} & $7.1$ & {\scriptsize$\pm$} & {\scriptsize 0.2} & $80.2$ & {\scriptsize$\pm$} & {\scriptsize 0.0} \\
Qwen2.5-3B & $6{,}277$ & {\scriptsize$\pm$} & {\scriptsize 31} & $4.2$ & {\scriptsize$\pm$} & {\scriptsize 0.0} & $83.8$ & {\scriptsize$\pm$} & {\scriptsize 0.0} \\
Qwen2.5-7B & $8{,}550$ & {\scriptsize$\pm$} & {\scriptsize 35} & $5.0$ & {\scriptsize$\pm$} & {\scriptsize 0.0} & $80.6$ & {\scriptsize$\pm$} & {\scriptsize 0.0} \\
Qwen3-8B & $8{,}604$ & {\scriptsize$\pm$} & {\scriptsize 27} & $4.4$ & {\scriptsize$\pm$} & {\scriptsize 0.0} & $83.7$ & {\scriptsize$\pm$} & {\scriptsize 0.0} \\
Qwen3-32B & $10{,}624$ & {\scriptsize$\pm$} & {\scriptsize 38} & $2.4$ & {\scriptsize$\pm$} & {\scriptsize 0.0} & $89.5$ & {\scriptsize$\pm$} & {\scriptsize 0.0} \\
LLaMA-3.2-3B & $4{,}614$ & {\scriptsize$\pm$} & {\scriptsize 26} & $5.0$ & {\scriptsize$\pm$} & {\scriptsize 0.1} & $80.8$ & {\scriptsize$\pm$} & {\scriptsize 0.1} \\
\bottomrule
\end{tabular}}
\captionof{table}{\hyperref[finding:8]{\ftag{F8}}\ Identity is a small minority of the router's training programs, while \textsf{keep} is the majority operation.
Router training-data composition, averaged across all $5$ DART-Math difficulty splits (mean $\pm$ sample standard deviation over the three search runs), $1{,}250$-question train slice. \emph{Examples} = program-examples, one per valid \gls{mcts}-discovered program, capped at $50$/question; \emph{Identity \%} = share of the identity program; \emph{Keep \%} = share of \textsf{keep} operations, across every example.}
\label{tab:router-training-data-composition}
\end{minipage}%
\end{figure}

\finding{7}\ \newfindingmark\ Program fragility is model-dependent, not difficulty-dependent:
for every genuine rescuing program (on average $2$ edited segments, $1.78$ to $2.07$ depending on the model), we revert one edited segment at a time back to \textsf{keep} and check whether the reward survives.
$15$--$33\%$ of rescuing programs have a single edit, which is then necessary by construction, since reverting it gives identity;
we test the others, wherever the search itself evaluated the reverted program (about $18{,}000$ programs per search run).
\Cref{fig:op_transition_robustness} plots the resulting break rate: it is roughly flat across DM-1--DM-5 in five of six models (only Qwen3-32B rises at DM-5),
ranging $44.5\%$ (Qwen3-32B) to $66.8\%$ (Qwen1.5-MoE-A2.7B) pooled over difficulty.
On 5 of 6 models, most edits inside a working program are individually essential, and which model it
is predicts fragility far more than how hard the question is.

\subsection{Router Reproduction: Format Repair and Identity Collapse}
\label{sec:router_collapse_results}

\begin{table}[h]
\centering
\caption{\hyperref[finding:8]{\ftag{F8}}\ Accuracy (\%) on our DART-Math split (\cref{sec:appendix_dart_math}; PoLar split: \cref{sec:appendix_their_split}) by difficulty.
\Base, \Router, and the fixed program are measured on the $500$ test questions per difficulty.
\gls{mcts} is the share of the $2{,}000$ questions with at least one valid program.
\gls{mcts}, \Router, and fixed program: mean $\pm$ sample standard deviation over the three search runs.
\Router\ p@1 equals \Base\ $\tau{=}0$ in all $90$ model--difficulty--seed cells and is not listed.
\gls{mcts} is an oracle upper bound, since it uses the ground truth to pick the program.
Bold: the fixed program or \Router\ p@5 when above the corresponding \Base\ ($\tau{=}0$, both one greedy decode, or p@5) by more than one standard deviation.
\Base\ does not depend on the search, and its sampling is a single run, so it has no standard deviation.}
\label{tab:main_results}
\footnotesize 
\renewcommand{\arraystretch}{0.92}
\begin{tabular}{lll *{5}{r@{\,}l}}
\toprule
Model & & & \multicolumn{2}{c}{DM-1} & \multicolumn{2}{c}{DM-2} & \multicolumn{2}{c}{DM-3} & \multicolumn{2}{c}{DM-4} & \multicolumn{2}{c}{DM-5} \\
\midrule
\multirow{6}{*}{Qwen1.5-MoE-A2.7B}
 & \multirow{3}{*}{\Base} & $\tau{=}0$ & 22.6 &  & 24.2 &  & 23.0 &  & 17.4 &  & 13.6 &  \\
 &  & p@1 & 21.6 &  & 24.2 &  & 21.2 &  & 18.2 &  & 11.4 &  \\
 &  & p@5 & 34.4 &  & 36.2 &  & 30.8 &  & 28.6 &  & 21.2 &  \\
 \cmidrule(l){2-13}
 & Fixed program & & 22.0 & {\tiny$\pm$0.5} & 17.6 & {\tiny$\pm$2.3} & 17.9 & {\tiny$\pm$2.1} & 14.4 & {\tiny$\pm$0.9} & 12.1 & {\tiny$\pm$1.1} \\
 \cmidrule(l){2-13}
 & \Router & p@5 & 35.1 & {\tiny$\pm$1.9} & \textbf{38.3} & {\tiny$\pm$0.2} & \textbf{33.7} & {\tiny$\pm$0.1} & \textbf{30.2} & {\tiny$\pm$0.0} & 19.1 & {\tiny$\pm$0.3} \\
 \cmidrule(l){2-13}
 & \gls{mcts} & & 73.3 & {\tiny$\pm$0.4} & 70.5 & {\tiny$\pm$0.8} & 67.0 & {\tiny$\pm$0.8} & 63.3 & {\tiny$\pm$0.9} & 43.6 & {\tiny$\pm$0.3} \\
\specialrule{0.35pt}{3pt}{3pt}
\multirow{6}{*}{Qwen2.5-3B}
 & \multirow{3}{*}{\Base} & $\tau{=}0$ & 32.8 &  & 28.8 &  & 23.8 &  & 16.6 &  & 8.2 &  \\
 &  & p@1 & 33.0 &  & 31.4 &  & 24.8 &  & 16.0 &  & 9.4 &  \\
 &  & p@5 & 41.2 &  & 37.0 &  & 33.6 &  & 24.0 &  & 15.4 &  \\
 \cmidrule(l){2-13}
 & Fixed program & & 31.1 & {\tiny$\pm$0.2} & 28.6 & {\tiny$\pm$0.7} & 24.1 & {\tiny$\pm$1.3} & 16.4 & {\tiny$\pm$1.1} & 7.9 & {\tiny$\pm$0.5} \\
 \cmidrule(l){2-13}
 & \Router & p@5 & \textbf{46.3} & {\tiny$\pm$0.4} & \textbf{45.5} & {\tiny$\pm$0.3} & \textbf{38.1} & {\tiny$\pm$1.4} & \textbf{29.3} & {\tiny$\pm$0.2} & \textbf{16.2} & {\tiny$\pm$0.3} \\
 \cmidrule(l){2-13}
 & \gls{mcts} & & 82.3 & {\tiny$\pm$0.3} & 81.3 & {\tiny$\pm$1.0} & 74.0 & {\tiny$\pm$1.3} & 67.5 & {\tiny$\pm$1.0} & 43.5 & {\tiny$\pm$0.4} \\
\specialrule{0.35pt}{3pt}{3pt}
\multirow{6}{*}{Qwen2.5-7B}
 & \multirow{3}{*}{\Base} & $\tau{=}0$ & 50.2 &  & 50.6 &  & 41.2 &  & 29.2 &  & 18.0 &  \\
 &  & p@1 & 50.2 &  & 50.4 &  & 41.2 &  & 29.4 &  & 19.0 &  \\
 &  & p@5 & 58.8 &  & 57.6 &  & 49.6 &  & 38.4 &  & 25.2 &  \\
 \cmidrule(l){2-13}
 & Fixed program & & 47.6 & {\tiny$\pm$1.7} & 48.3 & {\tiny$\pm$1.7} & 37.6 & {\tiny$\pm$0.0} & 27.5 & {\tiny$\pm$0.5} & 17.3 & {\tiny$\pm$0.6} \\
 \cmidrule(l){2-13}
 & \Router & p@5 & \textbf{63.9} & {\tiny$\pm$0.6} & \textbf{65.5} & {\tiny$\pm$0.2} & \textbf{53.2} & {\tiny$\pm$0.4} & \textbf{42.2} & {\tiny$\pm$0.6} & 25.2 & {\tiny$\pm$1.4} \\
 \cmidrule(l){2-13}
 & \gls{mcts} & & 89.0 & {\tiny$\pm$0.2} & 88.5 & {\tiny$\pm$0.2} & 81.4 & {\tiny$\pm$0.2} & 75.5 & {\tiny$\pm$0.4} & 50.8 & {\tiny$\pm$0.4} \\
\specialrule{0.35pt}{3pt}{3pt}
\multirow{6}{*}{Qwen3-8B}
 & \multirow{3}{*}{\Base} & $\tau{=}0$ & 48.2 &  & 46.0 &  & 35.6 &  & 28.2 &  & 9.4 &  \\
 &  & p@1 & 48.8 &  & 47.6 &  & 34.6 &  & 27.6 &  & 10.0 &  \\
 &  & p@5 & 54.0 &  & 53.8 &  & 42.2 &  & 35.2 &  & 13.0 &  \\
 \cmidrule(l){2-13}
 & Fixed program & & 48.2 & {\tiny$\pm$0.7} & 44.2 & {\tiny$\pm$1.0} & \textbf{37.1} & {\tiny$\pm$0.6} & 25.3 & {\tiny$\pm$1.3} & \textbf{10.1} & {\tiny$\pm$0.1} \\
 \cmidrule(l){2-13}
 & \Router & p@5 & \textbf{60.5} & {\tiny$\pm$0.8} & \textbf{58.5} & {\tiny$\pm$0.4} & \textbf{48.3} & {\tiny$\pm$0.3} & \textbf{36.7} & {\tiny$\pm$0.6} & \textbf{17.3} & {\tiny$\pm$0.6} \\
 \cmidrule(l){2-13}
 & \gls{mcts} & & 88.5 & {\tiny$\pm$0.4} & 86.9 & {\tiny$\pm$0.5} & 78.4 & {\tiny$\pm$0.5} & 72.3 & {\tiny$\pm$0.6} & 46.7 & {\tiny$\pm$0.4} \\
\specialrule{0.35pt}{3pt}{3pt}
\multirow{6}{*}{Qwen3-32B}
 & \multirow{3}{*}{\Base} & $\tau{=}0$ & 63.2 &  & 51.6 &  & 38.2 &  & 16.6 &  & 5.6 &  \\
 &  & p@1 & 60.0 &  & 48.2 &  & 35.6 &  & 17.4 &  & 5.2 &  \\
 &  & p@5 & 74.6 &  & 66.0 &  & 52.8 &  & 34.0 &  & 14.4 &  \\
 \cmidrule(l){2-13}
 & Fixed program & & \textbf{64.9} & {\tiny$\pm$1.3} & \textbf{62.0} & {\tiny$\pm$0.0} & \textbf{47.3} & {\tiny$\pm$2.2} & \textbf{33.5} & {\tiny$\pm$0.9} & \textbf{13.3} & {\tiny$\pm$0.7} \\
 \cmidrule(l){2-13}
 & \Router & p@5 & 76.9 & {\tiny$\pm$2.7} & 65.7 & {\tiny$\pm$2.6} & \textbf{61.0} & {\tiny$\pm$3.0} & \textbf{36.5} & {\tiny$\pm$0.9} & \textbf{16.3} & {\tiny$\pm$0.5} \\
 \cmidrule(l){2-13}
 & \gls{mcts} & & 97.1 & {\tiny$\pm$0.2} & 96.3 & {\tiny$\pm$0.2} & 91.3 & {\tiny$\pm$0.4} & 85.7 & {\tiny$\pm$0.4} & 58.4 & {\tiny$\pm$0.3} \\
\specialrule{0.35pt}{3pt}{3pt}
\multirow{6}{*}{LLaMA-3.2-3B}
 & \multirow{3}{*}{\Base} & $\tau{=}0$ & 18.8 &  & 14.4 &  & 17.0 &  & 19.8 &  & 24.4 &  \\
 &  & p@1 & 19.2 &  & 14.8 &  & 16.2 &  & 19.0 &  & 24.4 &  \\
 &  & p@5 & 27.2 &  & 25.0 &  & 25.2 &  & 28.0 &  & 35.8 &  \\
 \cmidrule(l){2-13}
 & Fixed program & & 17.2 & {\tiny$\pm$0.7} & 14.9 & {\tiny$\pm$2.2} & 14.3 & {\tiny$\pm$1.1} & 15.5 & {\tiny$\pm$0.7} & 21.9 & {\tiny$\pm$2.0} \\
 \cmidrule(l){2-13}
 & \Router & p@5 & 26.5 & {\tiny$\pm$1.0} & \textbf{26.9} & {\tiny$\pm$0.8} & 24.3 & {\tiny$\pm$3.6} & \textbf{29.0} & {\tiny$\pm$0.9} & 34.3 & {\tiny$\pm$0.6} \\
 \cmidrule(l){2-13}
 & \gls{mcts} & & 68.7 & {\tiny$\pm$1.2} & 67.2 & {\tiny$\pm$0.5} & 61.7 & {\tiny$\pm$0.7} & 65.3 & {\tiny$\pm$0.9} & 60.2 & {\tiny$\pm$0.1} \\
\bottomrule
\end{tabular}
\end{table}

\finding{8}\ \notreprodmark\,(\newfindingmark)\ \citet{li2026polar}'s headline router claim is a pass@$1$ gain over \Base\ (sampling) on every model they report, growing with $k$.
On the PoLar split, with \citeauthor{li2026polar}'s search and prompt, the router mostly does not collapse (except for LLaMA-3.2-3B),
but it gains only where \texttt{polar\_default} breaks the output format (Qwen3-8B, Qwen2.5-3B),
and how much depends on the search seed (\cref{tab:their_split_results,sec:appendix_their_split_variance}).
A single fixed program gains more than the router in every case,
and with \texttt{oneshot} neither gains on any model (\cref{sec:appendix_their_split_results}).
Out of distribution, the same router gains on ASDiv and MAWPS only by repairing answers cut off before the box, and does not gain on MMLU-Pro (\cref{sec:appendix_their_split_ood}).
Under their setup, the router's gain is therefore a repair of the output format, not a per-input choice of program.

In our setup (our split, our search, and our prompt, which already fixes the output format; \cref{sec:appendix_placeholder_echo}), the router's top-$1$ decode is the literal identity program on every model, difficulty, and training-data treatment tested,
both for our reimplementation and for \citeauthor{li2026polar}'s released router-training code run on our data (\cref{tab:main_results,tab:router-results-strict-ce}).
A single fixed program gains clearly only on Qwen3-32B, where it makes the model answer before the $50$-token cut-off rather than reason better (\cref{tab:main_results,sec:appendix_q32b_budget});
elsewhere it stays close to \Base.
No anti-collapse lever that we or their code expose gains over \Base\ on Qwen3-8B DM-$1$;
where one moves the router off identity, it loses at least as many questions as it rescues (\cref{sec:appendix_router_recipe}):
segmentation-threshold calibration, class-weighted and focal losses, down-weighting the identity program's training loss to zero, length-preference reweighting, and a \textsf{keep}-probability penalty.
Out of distribution, all routers also predict identity, on every ASDiv, MAWPS and MMLU-Pro question, so there the router equals \Base.

The collapse is a property of the setup, not of an under-tuned reproduction.
Identity is only $2.4$--$7.1\%$ of the training programs per model, but \textsf{keep} is $80.2$--$89.5\%$ of the per-layer labels (\cref{tab:router-training-data-composition}).
The router outputs the decoded average of its training labels:
for our search's spread-out edits that is identity, and for their search a single dominant edit, since it explores long programs below a single first edit and starts every question from the same seed (\cref{sec:appendix_their_split_summary}).
With their search on our split, the Qwen3-8B router does not collapse either, although its training labels are built the same way (every valid program, up to $50$ per question, as in their code), so the search decides whether the router collapses, and the prompt whether it then gains (\cref{sec:appendix_their_split_results}).
With neither search does the router learn a useful per-input choice of program (\cref{tab:their_split_menu}), and out of distribution it predicts one program for nearly every question (\cref{sec:appendix_their_split_ood}).

\finding{9}\ \newfindingmark\ \citeauthor{li2026polar}'s pass@$5$ gain does replicate in our setup (\cref{tab:main_results,tab:router-results-strict-ce}),
and unlike on the PoLar split, it is not a format repair: it comes from beam-search diversity.
The router never places a correct non-identity program at rank $1$, only somewhere in its top-$5$ (\cref{sec:appendix_router_recipe} has the full evidence chain, grading methodology, and candidate-pool diversity breakdown).
Furthermore, \cref{tab:router-topk5-menu-test} shows that the router's top-$5$ predicted programs collapse to a fixed menu of few programs only.

\subsection{MMLU-Pro Track}
\label{sec:mmlu_pro_track}

Beyond the DART-Math track above, we also ran a dedicated \gls{mcts} and router
training directly on MMLU-Pro. The finding below has no equivalent in \citeauthor{li2026polar}'s paper.

\finding{10}\ \newfindingmark\ On Qwen3-8B, with answers scored by the logits of the option letters, reshuffling which option sits behind which letter (content unchanged) shows
\gls{mcts}-discovered rescuing programs substantially overfit to the correct answer's
\emph{position}, not its content.
Reshuffled rescuing accuracy falls \emph{below}
identity's own, while an identity control confirms the base model itself is close to
shuffle-invariant (\cref{sec:appendix_mmlu_reshuffle_original}).
A dedicated, in-domain MMLU-Pro router shows the same
pass@$1$ identity-collapse as DART-Math,
and its pass@$5$ gain is position overfitting too:
it proposes one fixed menu of $5$ programs, three of which answer a near-constant letter,
and identity plus four fixed letters matches its pass@$5$ (\cref{sec:appendix_mmlu_router_random_baseline}).

\section{Related Work}
\label{sec:related_work}
\glsresetall

\textbf{Saving compute.}
One line of work removes layers permanently, the same way for every input:
LayerDrop \citep{fan2019layerdrop} trains a model so arbitrary layer subsets can be dropped at inference,
LaCo \citep{yang2024laco} merges consecutive layers via weight arithmetic,
and ShortGPT \citep{men2025shortgpt} drops layers ranked as redundant by input-output similarity.
\citet{sun2024painters} find the same pattern in language models, probing a similar skip/reorder/parallel-execution space to the one we investigate in \cref{sec:motivation};
\citet{balef2026tabularlayers} find it again outside language models, in tabular foundation models.
A second line keeps every layer but decides, per input, which ones to run:
early-exit classifiers stop once confident \citep{xin2020deebert,zhou2020pabee},
LayerSkip \citep{elhoushi2024layerskip} shares one such classifier across layers with self-speculative decoding,
and FlexiDepth \citep{luo2025flexidepth}, MindSkip \citep{he2024mindskip}, and GateSkip \citep{laitenberger2025gateskip}
use lightweight routers or residual gates to skip interior layers, still aimed at spending less compute, not unlocking more.

\textbf{Reusing compute.}
A third line reuses layers instead of removing them:
Universal Transformers \citep{dehghani2019universal} apply one shared block recurrently with a per-token halting policy,
looped transformers \citep{fan2025looped} repeat a block for a fixed or learned number of iterations to improve length generalization,
Encode-Think-Decode \citep{koishekenov2025encode} lets each token adaptively halt a middle, recurrent ``think'' block,
MoEUT \citep{csordas2024moeut} sparsifies the shared block with mixture-of-experts to scale parameter count without the cost of always running all of it,
and Inner Thinking Transformer \citep{chen2025inner} combines adaptive loops with a residual ``thinking'' stream and per-token routing.
Mixture-of-Depths \citep{raposo2024mod}, Mixture-of-Recursions \citep{bae2025mor},
and DNA \citep{cowsik2025towards} route at the token level in a similar spirit, but still need architectural change and training.
Unlike all of these, \gls{polar} \citep{li2026polar} reuses a frozen, pretrained model's own layers with no architectural change, training a router to predict a whole program upfront, replacing per-input search with a single forward pass.
DR.LLM \citep{heakl2025drllm} also learns routing policies from \gls{mcts} supervision, but locally, per layer, \emph{during} the forward pass, so it cannot represent the multi-layer block repetitions \gls{polar}'s (and our) programs use. As part of our reproduction, which confirms \gls{polar}'s central finding but not all of it (our router reproduction never matches its published numbers, and \cref{sec:analysis} surfaces failure modes it does not report), we release all code and data.
A separate, adjacent line unlocks a \emph{fixed} model's latent capability without retraining
through inference-time interventions on the input or embedding space, rather than the execution path:
random-token embedding injection \citep{devansh2026cheapest}, random soft prompts \citep{kim2026noise2diversity},
meaningless-token activation redistribution \citep{shi2025meaninglesstokens}, filler-token hidden computation \citep{pfau2024dotbydot},
prompt engineering \citep{liu2023promptsurvey}, and lightweight adaptation via \gls{lora} \citep{hu2022lora}.

\textbf{Brain-like compute.}
\label{sec:cortical_analogies}
\citet{babu2026corticosubcortical} show that a cortico-thalamic circuit can realize multihead self-attention,
with one head per cortical area and the thalamus computing and routing queries, keys, and values across them.
\citet{koenig2026neurotransformers} independently propose transformer operations as a framework for laminar cortical computation.
A program-of-layers resembles the coordinating role these accounts assign to the thalamus:
deciding which areas to engage, in what order, and when to loop, with the \gls{polar} \citep{li2026polar} predictor as a functional analogue of the thalamus.
To our knowledge this specific reading is new, and it is the main way this paper connects to neuroscience.
\citet{alkhamissi2025micro} similarly partition transformer layers into modules with specialized cognitive networks,
starting from pretrained layers but fine-tuning their weights into these modules,
unlike our own routing on top of a pretrained model's frozen layers; \cref{sec:appendix_brain_ml} reviews further such brain-inspired \gls{ml} approaches.
Both approaches coordinate computation \emph{globally}, over the full path rather than a single layer's decision,
in the spirit of the shared, bandwidth-limited channel of \citet{goyal2021globalworkspace}.
Mixture-of-experts routing looks similar but stays local, deciding per layer and per token.

\section{Conclusion}
\label{sec:conclusion}
\glsresetall
We revisited the \gls{polar} idea proposed by \citet{li2026polar} (i.e., input-dependent execution of transformer layers) and replicated its core finding that dynamically skipping or repeating layer operations reliably outperformed the standard, fixed-depth forward pass.
However, the main result, the router's gain, did not reproduce, and we found why:
in our setup, across every model and recipe tested, the router's top prediction consistently collapsed back to the default forward pass;
under their own setup, its gains came from repairing the output format, which a single fixed program did better.
Our broader analysis suggests that the collapse lies in the setup rather than in under-tuned training,
since with either search the router learns no useful per-input choice of program: with ours it predicts identity for every input, with theirs it mostly switches between identity and one dominant edit, and its other choices are less accurate than \Base.
Because the error surface over program space is highly fractured, single-shot deterministic prediction is a poor strategy.
This echoes how the brain resolves difficult computations, by engaging multiple pathways concurrently rather than committing to a single route prematurely, and opens a concrete direction for widening the program space itself (\cref{sec:limitations}).

The router thus does not yet play the coordinating role that we compare to the thalamus.
Yet the search finds a correct program for far more questions than the standard pass solves,
so a frozen model's layers can be reused as a library of functions, run in a different order, to improve accuracy,
not only to save compute by skipping, as most prior work does (\cref{sec:related_work}).
Even the collapsed router's top-$5$ programs mostly beat five samples of the unmodified model on our split (\cref{tab:main_results}), so what is missing is mainly the choice among them for each input.
What a thalamus-like router would need is visible in our findings: the useful programs form a small, largely input-independent menu (\hyperref[finding:5]{\ftag{F5}}), and they are fragile, so the choice must be exact (\hyperref[finding:7]{\ftag{F7}}).
A working thalamus analogue, a router that makes this choice for each input, also out of distribution, remains to be found.

\subsection{Limitations and Future Work}
\label{sec:limitations}
\label{sec:future_work}
A primary limitation of our study is that we cannot explain all of \citeauthor{li2026polar}'s router gains:
on Qwen2.5-3B and Qwen3-8B, \texttt{oneshot} alone exceeds their reported router accuracy, but on LLaMA-3.2-3B and Qwen1.5-MoE-A2.7B, which have no format failure, neither our router nor our prompts reach their reported numbers (\cref{sec:appendix_their_split_results}).
A router that weighs several candidate programs at once instead of committing to one, resembling the brain's approach, could address the collapse.
A router that decides per layer during the forward pass, from the hidden states, as DR.LLM does \citep{heakl2025drllm}, is another candidate;
we do not compare against it, since it cannot represent block repetitions (\cref{sec:related_work}).
Secondly, several of our \gls{mcts}-based findings could not reproduce \citet{li2026polar}'s numbers precisely even when matching their setup closely.
Beyond closing these two gaps, several other directions remain.
Splitting attention and \gls{mlp} sub-layers into independent operators would provide a finer-grained action space;
\citet{press2020sublayers} already show that reordering them can help, though when training from scratch.
Such a split could also uncover gains similar to those \citet{chen2026attnres} report from treating attention as a separate residual stream.
Moving from one static program per input to per-token or per-segment routing decisions is a natural extension.
Crucially, our operator set represents only a fraction of biological routing; extending the program space with a parallel-layer execution operator on the same input is a promising direction.
Finally, it is unclear whether individual layers play a similar functional role across different backbone models, an analogue at the layer level to cross-model concept neurons \citep{dravid2023rosetta}.

\section*{Broader Impact Statement}
A gain from a program-of-layers on a benchmark can come from the format or position of the answer rather than from better reasoning:
under \citet{li2026polar}'s setup, the router gains by repairing the output format (\hyperref[finding:8]{\ftag{F8}}),
and on MMLU-Pro, rescuing programs follow the position of the correct option (\hyperref[finding:10]{\ftag{F10}}).
Such gains should be checked with robustness tests, such as a prompt that fixes the answer format or reshuffling the answer options, before they are read as capability gains.
This matters most if programs are deployed, since a program chosen for its benchmark gain can fail on inputs without that artifact.
We release all discovered programs so that such tests can be run on them.

\ifdeanonymized
\subsubsection*{Acknowledgments}
Our work is funded by the Federal Ministry of Research, Technology and Space under grant number 01GQ2505A and by the Deutsche Forschungsgemeinschaft (DFG, German Research Foundation) -- FIP-12 -- Project-ID 528483508.
Responsibility for the content of this publication lies with the author.
\fi

\bibliographystyle{tmlr}
\bibliography{references}

\clearpage

\appendix
\etocdepthtag.toc{appendix}
\etocsettagdepth{mainbody}{none}
\etocsettagdepth{appendix}{section}
\etocsettocstyle{\section*{Appendix Contents}}{}
\tableofcontents

\section{Reproduction Details}
\glsresetall

This appendix documents our main track: our DART-Math split, prompt, search and router.
It gives the details the main text leaves out and, wherever \citet{li2026polar} leave a choice open or we depart from them, what we chose and why.
\Cref{sec:appendix_their_split} repeats the key parts under \citeauthor{li2026polar}'s own setup, and \repolar\ releases the code for all of it.

\subsection{Dataset Construction: Our DART-Math Split}
\label{sec:appendix_dart_math}

\citet{li2026polar} take their difficulty levels from MATH \citep{hendrycks2021math}: DM-$N$ is MATH level $N$ (\cref{sec:appendix_their_split_setup}).
We keep the size their first version\footnote{arXiv v1 of \citet{li2026polar}: \url{https://arxiv.org/abs/2606.06574v1}.} states, $2{,}000$ distinct questions per level,
but MATH alone has only $565$ distinct questions on DM-1, and their later version uses $565$ to $1{,}579$ per level.
We therefore add GSM8K \citep{cobbe2021gsm8k}, which has no levels, and sort both datasets together by the fail rate \citet{tong2024dartmath} report for each question,
which keeps all our levels at the same, larger size.
Our levels thus mix MATH and GSM8K questions by fail rate, and our split is not theirs.
We do not redistribute the resulting split directly: the MATH half of the underlying pool traces back to the same competition-math problems as \texttt{hendrycks/competition\_math}, now delisted from HuggingFace amid a disputed-ownership claim over the source content, so re-hosting a derived copy ourselves would just repackage the same disputed content once more; the GSM8K half has no such issue.
Instead, our released code includes a single script that deterministically rebuilds this exact split from the public DART-Math pool, available at
\ifdeanonymized
\url{https://github.com/DATEXIS/RE-PoLar}.
The released data (\url{https://huggingface.co/datasets/SWAMBIT/re-polar}, about $6$\,GB) contain, for every model, search run and level, every program the search evaluated with its verdict, the model's answer behind each verdict, the trained routers with their test results, the fixed programs and the menu evaluation, but no question text.
\else
[\emph{removed for double-blind review}].
\fi

To build the pool, we use the following four HuggingFace datasets, released by \citet{tong2024dartmath}:
the MATH query pool (\texttt{hkust-nlp/dart-math-pool-math}, giving question, ground-truth answer, and level, with $7{,}473$ distinct queries) and its per-query fail rates (\texttt{hkust-nlp/dart-math-pool-math-query-info}), and the corresponding GSM8K query pool (\texttt{hkust-nlp/dart-math-pool-gsm8k}, $7{,}468$ distinct queries) and fail rates (\texttt{hkust-nlp/dart-math-pool-gsm8k-query-info}).
Fail rate is defined as $1-\text{pass rate}$, where the pass rate is the value \citeauthor{tong2024dartmath} measure from sampling DeepSeekMath-7B-RL \citep{shao2024deepseekmath}.
We join each pool with its query-info table on query id, concatenate the MATH and GSM8K records into a single pool, and sort the result ascending by fail rate, breaking ties by query id for determinism.

We then cut the sorted pool into 5 near-equal quantile bins, from easiest to hardest, and call these bins DM-1 (easiest) through DM-5 (hardest).
Within each bin we apply a seeded shuffle (seed $42$, per level) and then take the first $1{,}250$ queries for train, the next $250$ for val, and the next $500$ for test, giving $2{,}000$ queries per level and $10{,}000$ in total, and capping each level well below the full labeled pool's $14{,}941$ queries across the 5 levels.
Because the pool is sorted by fail rate, the mix of the two datasets changes strongly with the level (\cref{tab:dart_math_mix}): DM-1 is almost all GSM8K, DM-5 almost all MATH.

\begin{table}[!ht]
\centering
\small
\caption{Our easier levels are mostly GSM8K and our harder levels mostly MATH ($92\%$ vs.\ $5\%$ GSM8K at DM-1 and DM-5).
Composition of our DM-1 to DM-5 ($2{,}000$ questions each) by source dataset, and the range of DART-Math fail rates (from DeepSeekMath-7B-RL samples) in each level.}
\label{tab:dart_math_mix}
\begin{tabular}{lrrrl}
\toprule
Level & MATH & GSM8K & GSM8K \% & Fail rate \\
\midrule
DM-1 & $161$ & $1{,}839$ & $92.0$ & $0$--$0.004$ \\
DM-2 & $604$ & $1{,}396$ & $69.8$ & $0.004$--$0.018$ \\
DM-3 & $836$ & $1{,}164$ & $58.2$ & $0.018$--$0.066$ \\
DM-4 & $1{,}508$ & $492$ & $24.6$ & $0.066$--$0.60$ \\
DM-5 & $1{,}903$ & $97$ & $4.8$ & $0.60$--$1$ \\
\midrule
All & $5{,}012$ & $4{,}988$ & $49.9$ & \\
\bottomrule
\end{tabular}
\end{table}

\subsection{Prompt Sensitivity in Evaluation}
\label{sec:appendix_placeholder_echo}
Initially, we adopted \citet{li2026polar}'s own DART-Math evaluation protocol verbatim, which we call \texttt{polar\_default} throughout this paper.
In their Appendix~D.4, models are instructed to answer with ``\texttt{...formatted strictly as \textbackslash boxed\{ANSWER\}},'' under greedy, 50-token-capped generation.
The word inside the braces however is a literal placeholder, not an empty template.
This section documents two failure modes that follow from this and how we fix both. 
This is not unexpected: \gls{llm} evaluation is known to be highly sensitive to prompt-formatting choices that are semantically irrelevant to the underlying task \citep{sclar2024quantifying}.
On the PoLar split, these two failures are where all gains of their router and of a fixed program come from (\cref{sec:appendix_their_split_results}).
We therefore remove them in our setup, so that a router can only gain by choosing better programs, not by repairing the format.

\paragraph{Two failure modes.}
We call the first \errtag{prompt template echo}: the model emits \texttt{\textbackslash boxed\{ANSWER\}} itself,
verbatim, instead of a real answer
(also catching \texttt{\$\textbackslash boxed\{ANSWER\}\$}, \texttt{\textbackslash boxed\{\textbackslash text\{answer\}\}}, and case variants).
The DART-Math extraction code we use, unmodified from \citeauthor{li2026polar}'s own,
always takes the \emph{last} \texttt{\textbackslash boxed\{...\}} span in the response;
an echo is almost always the response's only span, so taking the first span instead changes accuracy by at most $0.2$pp (``A prompt-format sweep'' below).
On Qwen3-8B, $43.8\%$ of \Base's generations under this exact wording are an echo.
A second, mechanistically distinct failure produces no \texttt{\textbackslash boxed\{\}} span at all for a different reason:
the model reasons in free-form prose and exhausts the 50-token budget before ever writing \texttt{\textbackslash boxed\{\}},
rather than echoing the placeholder
(this holds even for the Qwen3 models, where thinking mode is explicitly disabled for every generation in this paper).
This is the same \errtag{unboxed} failure mode \cref{sec:appendix_bht_large} later reports for Qwen3-32B.
It dominates \texttt{polar\_default} on 3 of our 6 models,
each with a near-zero echo rate.
\Cref{tab:echo-sweep} reports \errtag{unboxed} rate and \errtag{prompt template echo} rate as separate columns
because accuracy alone conflates the two failure modes.

\paragraph{A prompt-format sweep.}
A defect this large was worth checking for a fix that does not change the answer format itself,
and for how much it varies across models.
We swept 6 prompt wordings on all 6 models in our study, using our DART-Math split
(\cref{sec:appendix_dart_math}, $10{,}000$ samples),
and report accuracy, \errtag{unboxed} rate, and \errtag{prompt template echo} rate for each
(\cref{tab:echo-sweep}; the exact wordings are in \cref{tab:prompt-templates}).
The sweep evaluates the standard forward pass only (greedy), so it does not depend on the search.
We started from the protocol of DR.LLM \citep{heakl2025drllm},
one of the closest related works to \gls{polar} (\cref{sec:related_work}) that also learns \gls{mcts}-supervised routing policies,
so testing its exact wording is a natural comparison point.
An earlier, smaller pilot sweep showed that two interventions work best, a response prefill and a single worked demonstration.
The final sweep therefore contains \texttt{polar\_default}, DR.LLM's own wording, and DR.LLM's chat-native prefill and one-shot variants, combined and separately.
It also contains \texttt{oneshot}, a variant we designed to stay as close as possible to \citeauthor{li2026polar}'s own wording:
it keeps their exact instruction and format and only adds one trivial worked example (``What is 1+1?'') in the same style.
Because this variant repeats the instruction-and-answer block twice inside the 50-token budget,
the model may hallucinate a third, incomplete \texttt{\textbackslash boxed\{...\}} span that a last-match extraction would grab instead of the real answer.
We therefore truncate before grading at the first complete span, so \texttt{oneshot} is, in effect, graded on the \emph{first} span rather than the last.
Grading \texttt{polar\_default} the same way changes its accuracy by at most $0.2$pp on every model, so \texttt{oneshot}'s gain comes from the demonstration, not from the grading rule.

\begin{table}[tb]
\centering
\footnotesize
\caption{\texttt{oneshot} is the most accurate prompt on 4 of 6 models and never significantly worse than \texttt{polar\_default};
only on Qwen3-32B is a chat-templated variant better ($51.4\%$ vs.\ $34.8\%$).
Accuracy, \errtag{unboxed} and \errtag{prompt template echo} rates for all $36$ model/prompt combinations on our DART-Math split (\cref{sec:appendix_dart_math}).
\textbf{Bold}: best accuracy per model.
\texttt{polar\_default}/\texttt{oneshot} are graded with \citet{li2026polar}'s extraction, the \texttt{drllm} variants with \citeauthor{heakl2025drllm}'s \texttt{mathruler} \citep{heakl2025drllm}.
The \texttt{drllm} variants are also wrapped in the model's chat template (one user turn, no system message), which on its own lowers accuracy with \texttt{oneshot}'s content (\cref{tab:chat-template-ablation}).
Superscripts: significant difference from the model's \texttt{polar\_default} row in either direction (McNemar's test, $^{***}$ $p<0.001$, $^{*}$ $p<0.05$).
\Cref{tab:prompt-templates} gives every variant's wording.}
\label{tab:echo-sweep}
\begin{tabular}{lllccc}
\toprule
Model & Variant & Grader & Accuracy & \errtag{unboxed} & \errtag{prompt template echo} \\
\midrule
\multirow{6}{*}{Qwen1.5-MoE-A2.7B}
 & \texttt{polar\_default} & \texttt{PoLar} & 16.3\% & 0.3\% & 2.70\% \\
 & \texttt{drllm} (faithful) & \texttt{mathruler} & 2.0\%$^{***}$ & 89.2\% & 0.00\% \\
 & \texttt{drllm\_chat\_prefill} & \texttt{mathruler} & 15.6\% & 0.0\% & 0.00\% \\
 & \texttt{oneshot} & \texttt{PoLar} & \textbf{19.3\%}$^{***}$ & 2.0\% & 0.00\% \\
 & \texttt{drllm\_chat\_oneshot} & \texttt{mathruler} & 13.2\%$^{***}$ & 18.3\% & 0.00\% \\
 & \texttt{drllm\_chat\_oneshot\_prefill} & \texttt{mathruler} & 15.3\%$^{*}$ & 0.0\% & 0.00\% \\
\midrule
\multirow{6}{*}{Qwen2.5-3B}
 & \texttt{polar\_default} & \texttt{PoLar} & 0.1\% & 99.5\% & 0.00\% \\
 & \texttt{drllm} (faithful) & \texttt{mathruler} & 22.1\%$^{***}$ & 2.0\% & 0.00\% \\
 & \texttt{drllm\_chat\_prefill} & \texttt{mathruler} & 22.0\%$^{***}$ & 0.0\% & 0.00\% \\
 & \texttt{oneshot} & \texttt{PoLar} & \textbf{22.6\%}$^{***}$ & 1.0\% & 0.00\% \\
 & \texttt{drllm\_chat\_oneshot} & \texttt{mathruler} & 21.6\%$^{***}$ & 0.0\% & 0.00\% \\
 & \texttt{drllm\_chat\_oneshot\_prefill} & \texttt{mathruler} & 21.2\%$^{***}$ & 0.0\% & 0.00\% \\
\midrule
\multirow{6}{*}{Qwen2.5-7B}
 & \texttt{polar\_default} & \texttt{PoLar} & 22.7\% & 42.6\% & 0.00\% \\
 & \texttt{drllm} (faithful) & \texttt{mathruler} & 35.0\%$^{***}$ & 0.0\% & 0.00\% \\
 & \texttt{drllm\_chat\_prefill} & \texttt{mathruler} & 33.7\%$^{***}$ & 0.0\% & 0.00\% \\
 & \texttt{oneshot} & \texttt{PoLar} & \textbf{37.0\%}$^{***}$ & 0.0\% & 0.00\% \\
 & \texttt{drllm\_chat\_oneshot} & \texttt{mathruler} & 35.0\%$^{***}$ & 0.0\% & 0.00\% \\
 & \texttt{drllm\_chat\_oneshot\_prefill} & \texttt{mathruler} & 33.6\%$^{***}$ & 0.0\% & 0.00\% \\
\midrule
\multirow{6}{*}{Qwen3-8B}
 & \texttt{polar\_default} & \texttt{PoLar} & 22.1\% & 0.4\% & 43.84\% \\
 & \texttt{drllm} (faithful) & \texttt{mathruler} & 5.1\%$^{***}$ & 91.2\% & 0.00\% \\
 & \texttt{drllm\_chat\_prefill} & \texttt{mathruler} & 31.6\%$^{***}$ & 0.0\% & 0.00\% \\
 & \texttt{oneshot} & \texttt{PoLar} & \textbf{33.7\%}$^{***}$ & 0.0\% & 0.02\% \\
 & \texttt{drllm\_chat\_oneshot} & \texttt{mathruler} & 30.0\%$^{***}$ & 6.9\% & 0.00\% \\
 & \texttt{drllm\_chat\_oneshot\_prefill} & \texttt{mathruler} & 30.5\%$^{***}$ & 0.0\% & 0.00\% \\
\midrule
\multirow{6}{*}{Qwen3-32B}
 & \texttt{polar\_default} & \texttt{PoLar} & 0.5\% & 98.2\% & 0.17\% \\
 & \texttt{drllm} (faithful) & \texttt{mathruler} & 31.2\%$^{***}$ & 46.4\% & 0.00\% \\
 & \texttt{drllm\_chat\_prefill} & \texttt{mathruler} & 48.8\%$^{***}$ & 0.0\% & 0.00\% \\
 & \texttt{oneshot} & \texttt{PoLar} & 34.8\%$^{***}$ & 38.6\% & 1.01\% \\
 & \texttt{drllm\_chat\_oneshot} & \texttt{mathruler} & \textbf{51.4\%}$^{***}$ & 0.2\% & 0.00\% \\
 & \texttt{drllm\_chat\_oneshot\_prefill} & \texttt{mathruler} & 49.4\%$^{***}$ & 0.0\% & 0.00\% \\
\midrule
\multirow{6}{*}{LLaMA-3.2-3B}
 & \texttt{polar\_default} & \texttt{PoLar} & \textbf{19.4\%} & 2.4\% & 0.08\% \\
 & \texttt{drllm} (faithful) & \texttt{mathruler} & 16.0\%$^{***}$ & 25.7\% & 0.00\% \\
 & \texttt{drllm\_chat\_prefill} & \texttt{mathruler} & 19.3\% & 0.0\% & 0.00\% \\
 & \texttt{oneshot} & \texttt{PoLar} & 19.2\% & 0.0\% & 0.00\% \\
 & \texttt{drllm\_chat\_oneshot} & \texttt{mathruler} & 18.8\% & 1.5\% & 0.00\% \\
 & \texttt{drllm\_chat\_oneshot\_prefill} & \texttt{mathruler} & 18.2\%$^{*}$ & 0.0\% & 0.00\% \\
\bottomrule
\end{tabular}
\end{table}

\begin{table}[tb]
\centering
\scriptsize
\caption{The 6 prompt variants of \cref{tab:echo-sweep}, every message in full (question abbreviated as \texttt{<q>}).
$\dagger$: chat-templated (\cref{sec:appendix_chat_template}); the template's special tokens are omitted.}
\label{tab:prompt-templates}
\begin{tabular}{lp{0.68\linewidth}}
\toprule
Variant & Template \\
\midrule
\texttt{polar\_default} &
``\texttt{Solve the following math problem and output ONLY the final answer directly, formatted strictly as \textbackslash boxed\{ANSWER\}.\newline \#\#\# Problem Start\newline <q>\newline \#\#\# Problem End\newline Answer:}'' \\
\addlinespace
\texttt{drllm} (faithful)$^\dagger$ &
User: ``\texttt{Question: <q>\newline The final answer MUST BE put in \textbackslash boxed\{\} and no explanation.}'' \\
\addlinespace
\texttt{drllm\_chat\_prefill}$^\dagger$ &
User: ``\texttt{Question: <q>\newline The final answer MUST BE put in \textbackslash boxed\{\} and no explanation.}''
\newline Assistant: ``\texttt{\textbackslash boxed\{}'' \\
\addlinespace
\texttt{oneshot} &
``\texttt{Solve the following math problem and output ONLY the final answer directly, formatted strictly as \textbackslash boxed\{ANSWER\}.\newline \#\#\# Problem Start\newline What is 1+1?\newline \#\#\# Problem End\newline Answer: \textbackslash boxed\{2\}\newline\newline Solve the following math problem and output ONLY the final answer directly, formatted strictly as \textbackslash boxed\{ANSWER\}.\newline \#\#\# Problem Start\newline <q>\newline \#\#\# Problem End\newline Answer:}'' \\
\addlinespace
\texttt{drllm\_chat\_oneshot}$^\dagger$ &
User: ``\texttt{Question: What is 1+1?\newline The final answer MUST BE put in \textbackslash boxed\{\} and no explanation.}''
\newline Assistant: ``\texttt{\textbackslash boxed\{2\}}''
\newline User: ``\texttt{Question: <q>\newline The final answer MUST BE put in \textbackslash boxed\{\} and no explanation.}'' \\
\addlinespace
\texttt{drllm\_chat\_oneshot\_prefill}$^\dagger$ &
User: ``\texttt{Question: What is 1+1?\newline The final answer MUST BE put in \textbackslash boxed\{\} and no explanation.}''
\newline Assistant: ``\texttt{\textbackslash boxed\{2\}}''
\newline User: ``\texttt{Question: <q>\newline The final answer MUST BE put in \textbackslash boxed\{\} and no explanation.}''
\newline Assistant: ``\texttt{\textbackslash boxed\{}'' \\
\bottomrule
\end{tabular}
\end{table}

\paragraph{Both fixes work almost everywhere.}
Response prefill and the single trivial demonstration both remove almost all format failures (\errtag{unboxed} answers and \errtag{prompt template echo}), at no extra token cost beyond our existing 50-token budget,
except that Qwen3-32B still runs out of tokens under \texttt{oneshot} (\cref{sec:appendix_q32b_budget}).
Both interventions are a significant improvement over \texttt{polar\_default} (McNemar's test, $p<0.001$) on 4 of 6 models: Qwen2.5-3B, Qwen2.5-7B, Qwen3-8B, Qwen3-32B.
On Qwen1.5-MoE-A2.7B only \texttt{oneshot} is significant ($p<10^{-14}$); \texttt{drllm\_chat\_prefill} is not ($p=0.06$, numerically slightly \emph{worse} than \texttt{polar\_default}).
Two models are exceptions, for different reasons.
On LLaMA-3.2-3B, neither intervention significantly improves on \texttt{polar\_default}.
On Qwen3-32B, \texttt{oneshot} is a large, significant gain ($+34.3$pp, $p<0.001$), but only a partial fix, as its \errtag{unboxed} rate falls from $98.2\%$ only to $38.6\%$.
Overall, \texttt{oneshot} stays close to \citeauthor{li2026polar}'s original wording (appropriate for a reproduction) while fixing the majority of the format-compliance failures: the \errtag{unboxed} rate falls substantially for Qwen3-32B and the Qwen2.5 models, and the \errtag{prompt template echo} rate falls from $43.8\%$ on Qwen3-8B to effectively $0\%$ across the board.
Both failure modes are fixed, not eliminated, under this tradeoff.
\Cref{sec:appendix_bht_large} tags what remains in more depth, including the \errtag{unboxed}/\errtag{prompt template echo} split used there.
On the PoLar split, the same comparison gives the same picture (\cref{tab:their_split_base}).
\texttt{oneshot} is more accurate than \texttt{polar\_default} for all three Qwen models on every level, and within $4$pp for LLaMA-3.2-3B, so our choice of prompt does not depend on our split.

\paragraph{Consequence for our program analysis.}
Left unaddressed, \texttt{polar\_default} would make a large share of ``rescuing programs'' on Qwen3-8B specifically just formatting fixes for the echo defect.
Better still, \texttt{oneshot} improves accuracy relative to \texttt{polar\_default} on every model except a tie on LLaMA-3.2-3B,
so we use it as the generation prompt throughout our program-level analysis, on every model.
Still, Qwen3-32B carries a residual risk the others do not: even under \texttt{oneshot}, $38.6\%$ of its generations still contain no \texttt{\textbackslash boxed\{\}} span, so format-compliance failures may still contaminate its rescuing-program counts, reflected in the error-category breakdown in \cref{sec:appendix_bht_large}.
\Cref{sec:appendix_q32b_budget} shows that these answers are cut off at $50$ tokens, and that this cut-off explains the fixed program's gain on Qwen3-32B.

\subsection{Decision against Chat Templates}
\label{sec:appendix_chat_template}

Beyond prompt \emph{content} (as discussed before in \cref{sec:appendix_placeholder_echo}), a separate, largely orthogonal axis is prompt \emph{mechanism}: whether the model's native chat template is applied at all.
We ablate this directly, holding content fixed at \texttt{oneshot} on both sides, on a $200$-question pooled sample (seed $42$, DART-Math) per model (\cref{tab:chat-template-ablation}). Applying the chat template lowers accuracy on all $6$ models, typically by a substantial amount. We therefore use no chat template in our setup.
On the PoLar split, the template lowers accuracy in most cells as well (\cref{sec:appendix_their_split_base}), so we leave it out there too, unless stated otherwise.

\begin{table}[tb]
\centering
\small
\caption{Applying the model's chat template lowers \Base($\tau{=}0$) accuracy on all 6 models.
Prompt content fixed at \texttt{oneshot}; $200$-question pooled sample (seed $42$) per model.
``Chat template'' is the model's native template, with the system message that \citet{li2026polar}'s \texttt{polar/eval.py} adds for the Qwen2.5 models and Qwen1.5-MoE-A2.7B, and without one for Qwen3-8B, as in their code.
For LLaMA-3.2-3B and Qwen3-32B, it is the model's own template with one user turn; \citeauthor{li2026polar} apply no template to LLaMA-3.2-3B and do not study Qwen3-32B.}
\label{tab:chat-template-ablation}
\begin{tabular}{lccc}
\toprule
Model & no chat template & chat template & $\Delta$ (pp) \\
\midrule
Qwen1.5-MoE-A2.7B & $22.5\%$ & $7.0\%$ & \textcolor{polred}{$-15.5$} \\
Qwen2.5-3B & $22.0\%$ & $20.0\%$ & \textcolor{polred}{$-2.0$} \\
Qwen2.5-7B & $44.0\%$ & $36.5\%$ & \textcolor{polred}{$-7.5$} \\
Qwen3-8B & $34.0\%$ & $33.5\%$ & \textcolor{polred}{$-0.5$} \\
Qwen3-32B & $40.0\%$ & $1.0\%$ & \textcolor{polred}{$-39.0$} \\
LLaMA-3.2-3B & $20.5\%$ & $2.5\%$ & \textcolor{polred}{$-18.0$} \\
\bottomrule
\end{tabular}
\end{table}

For LLaMA-3.2-3B, no chat template is also what \citet{li2026polar}'s code does.
For their other models (Qwen3-8B, Qwen2.5-3B, Qwen1.5-MoE-A2.7B), their code applies one, but leaving it out gives higher accuracy on our split (\cref{tab:chat-template-ablation}) and in most cells of the PoLar split (\cref{sec:appendix_their_split_base}).

\subsection{Full Baseline Reproduction Breakdown}
\label{sec:appendix_baseline_full}

\Cref{tab:baseline-dart-math} extends \cref{tab:main_results} with the p@2--p@4 it omits,
all under \texttt{oneshot} with no chat template (\cref{sec:appendix_placeholder_echo,sec:appendix_chat_template} argue for this choice).

\begin{table}[tb]
\centering
\scriptsize
\caption{Full \Base($\tau{=}0$) and \Base(sampling) p@1--p@5 breakdown (\%) on our DART-Math split (\cref{sec:appendix_dart_math}), $500$ questions/difficulty, all 6 models, \texttt{oneshot}, no chat template (\cref{sec:appendix_chat_template}).
One sampling run per model; \Base\ does not depend on the search, so these are the same numbers as in \cref{tab:main_results}.}
\label{tab:baseline-dart-math}
\begin{tabular}{ll *{5}{r}}
\toprule
Model & Metric & DM-1 & DM-2 & DM-3 & DM-4 & DM-5 \\
\midrule
\multirow{6}{*}{Qwen1.5-MoE-A2.7B}
 & $\tau{=}0$ & 22.6 & 24.2 & 23.0 & 17.4 & 13.6 \\
 & p@1 & 21.6 & 24.2 & 21.2 & 18.2 & 11.4 \\
 & p@2 & 25.8 & 29.0 & 24.8 & 21.2 & 14.6 \\
 & p@3 & 30.2 & 31.6 & 27.6 & 24.2 & 17.4 \\
 & p@4 & 32.4 & 35.0 & 29.0 & 27.4 & 19.8 \\
 & p@5 & 34.4 & 36.2 & 30.8 & 28.6 & 21.2 \\
\midrule
\multirow{6}{*}{Qwen2.5-3B}
 & $\tau{=}0$ & 32.8 & 28.8 & 23.8 & 16.6 & 8.2 \\
 & p@1 & 33.0 & 31.4 & 24.8 & 16.0 & 9.4 \\
 & p@2 & 37.2 & 34.8 & 29.4 & 19.0 & 11.2 \\
 & p@3 & 39.2 & 35.4 & 31.6 & 21.0 & 13.0 \\
 & p@4 & 40.2 & 36.2 & 32.8 & 22.4 & 14.2 \\
 & p@5 & 41.2 & 37.0 & 33.6 & 24.0 & 15.4 \\
\midrule
\multirow{6}{*}{Qwen2.5-7B}
 & $\tau{=}0$ & 50.2 & 50.6 & 41.2 & 29.2 & 18.0 \\
 & p@1 & 50.2 & 50.4 & 41.2 & 29.4 & 19.0 \\
 & p@2 & 52.4 & 53.0 & 44.8 & 33.6 & 20.8 \\
 & p@3 & 56.2 & 55.4 & 46.8 & 35.4 & 22.2 \\
 & p@4 & 57.8 & 56.8 & 48.2 & 37.2 & 24.0 \\
 & p@5 & 58.8 & 57.6 & 49.6 & 38.4 & 25.2 \\
\midrule
\multirow{6}{*}{Qwen3-8B}
 & $\tau{=}0$ & 48.2 & 46.0 & 35.6 & 28.2 & 9.4 \\
 & p@1 & 48.8 & 47.6 & 34.6 & 27.6 & 10.0 \\
 & p@2 & 50.4 & 50.4 & 38.6 & 31.4 & 11.0 \\
 & p@3 & 52.0 & 51.6 & 40.2 & 32.4 & 11.6 \\
 & p@4 & 53.0 & 53.0 & 41.4 & 33.8 & 12.4 \\
 & p@5 & 54.0 & 53.8 & 42.2 & 35.2 & 13.0 \\
\midrule
\multirow{6}{*}{Qwen3-32B}
 & $\tau{=}0$ & 63.2 & 51.6 & 38.2 & 16.6 & 5.6 \\
 & p@1 & 60.0 & 48.2 & 35.6 & 17.4 & 5.2 \\
 & p@2 & 65.6 & 53.8 & 40.8 & 20.6 & 8.0 \\
 & p@3 & 70.8 & 59.2 & 45.2 & 25.8 & 11.2 \\
 & p@4 & 73.0 & 62.8 & 48.6 & 30.8 & 12.6 \\
 & p@5 & 74.6 & 66.0 & 52.8 & 34.0 & 14.4 \\
\midrule
\multirow{6}{*}{LLaMA-3.2-3B}
 & $\tau{=}0$ & 18.8 & 14.4 & 17.0 & 19.8 & 24.4 \\
 & p@1 & 19.2 & 14.8 & 16.2 & 19.0 & 24.4 \\
 & p@2 & 22.6 & 18.0 & 20.2 & 21.2 & 28.2 \\
 & p@3 & 24.6 & 21.2 & 21.8 & 25.0 & 32.0 \\
 & p@4 & 25.2 & 23.4 & 23.4 & 26.4 & 34.6 \\
 & p@5 & 27.2 & 25.0 & 25.2 & 28.0 & 35.8 \\
\bottomrule
\end{tabular}
\end{table}

\Cref{tab:baseline-mmlu-ood} reports \Base($\tau{=}0$) accuracy on MMLU-Pro (all 14 official categories, $n{=}12{,}032$, scored as forced-choice: a single forward pass's argmax over answer-letter logits, no generation step) and the two other \gls{ood} transfer targets,
ASDiv \citep{miao2020asdiv} and MAWPS \citep{kadlcik2023mawps}, both scored under the identical setup as \cref{tab:baseline-dart-math}.
Where \citet{li2026polar} report matching ASDiv/MAWPS numbers, $\Delta$ (in parentheses) gives ours minus theirs; we do not report a matching MMLU-Pro $\Delta$, since \citeauthor{li2026polar} publish only a per-subject breakdown (13 of the 14 categories we evaluate, excluding computer science), never a single aggregate accuracy.

\begin{table}[tb]
\centering
\small
\caption{Our \Base\ is above \citet{li2026polar}'s reported numbers on ASDiv and MAWPS except for LLaMA-3.2-3B (both) and Qwen1.5-MoE-A2.7B (ASDiv).
\Base($\tau{=}0$) accuracy (\%): MMLU-Pro (overall, all 14 categories, forced-choice logit scoring) and the two other \gls{ood} transfer targets, ASDiv ($n{=}2{,}305$) and MAWPS ($n{=}520$, generation-based, same prompt/mechanism as \cref{tab:baseline-dart-math}), all 6 models. Colored $\Delta =$ ours $-$ \citeauthor{li2026polar}'s matching table, where one exists.}
\label{tab:baseline-mmlu-ood}
\begin{tabular}{lc r@{\hspace{1pt}}l@{\hspace{6pt}} r@{\hspace{1pt}}l}
\toprule
Model & MMLU-Pro (overall) & \multicolumn{2}{c}{ASDiv} & \multicolumn{2}{c}{MAWPS} \\
\midrule
Qwen1.5-MoE-A2.7B & 24.4 & 52.7 & \dr{-6.4} & 62.3 & \dg{+20.6} \\
Qwen2.5-3B & 35.2 & 68.1 & \dg{+18.6} & 62.9 & \dg{+26.7} \\
Qwen2.5-7B & 42.3 & 75.3 & & 67.9 & \\
Qwen3-8B & 42.1 & 78.7 & \dg{+11.6} & 71.5 & \dg{+19.4} \\
Qwen3-32B & 48.7 & 77.0 & & 77.7 & \\
LLaMA-3.2-3B & 31.4 & 64.4 & \dr{-14.0} & 70.6 & \dr{-0.9} \\
\bottomrule
\end{tabular}
\end{table}

\subsection{Why the Fixed Program Gains on Qwen3-32B}
\label{sec:appendix_q32b_budget}
On Qwen3-32B, the fixed program (one program per difficulty, chosen on the validation questions and applied to every test question, \cref{sec:study_setup}) gains $+6.6$ to $+17.8$pp over identity on DM-2 to DM-5 for every seed (\cref{tab:main_results}); for the other models it stays between $-9.0$ and $+2.2$ (Qwen3-32B DM-1: $+0.2$ to $+2.8$).
The cause is the output format, not better reasoning.
Our \texttt{oneshot} prompt asks for the boxed answer only.
Qwen3-8B does so on every question, but the identity program of Qwen3-32B often starts to reason instead and is cut off after $50$ tokens before any box ($8\%$ of the answers on DM-1, $75\%$ on DM-5; \cref{tab:q32b_budget}).
Such a question counts as wrong.
The fixed program puts the model back into answering first, starting with a box on $83$ to $99\%$ of the questions (seed $42$).
Given more tokens, identity does not need this, as it is above the fixed program on every level at $512$ and at $1{,}024$ tokens (on DM-5, $173$ of $500$ answers still have no box at $1{,}024$ tokens).
So the gain of the fixed program on Qwen3-32B repairs answers that are cut off at $50$ tokens, but does not make the model solve more.
On DM-5, it leaves fewer answers without a box than identity with $1{,}024$ tokens ($16.2$ vs.\ $34.6\%$), yet identity is far more accurate there ($37.0$ vs.\ $12.6\%$), since it reasons before it answers.
The router predicts identity on every test question (\cref{sec:router_collapse_results}), so it does not find this repair.

\begin{table}[tb]
\centering
\small
\caption{At $50$ tokens, the fixed program gains on Qwen3-32B because it makes the model box an answer before the cut-off; identity with more tokens gains far more.
Accuracy (\%) on the $500$ test questions per difficulty of identity with a generation cap of $50$, $512$ and $1{,}024$ tokens, and of the fixed program (search run seed $42$) with $50$ tokens.
In brackets: share (\%) of answers without a box.}
\label{tab:q32b_budget}
\begin{tabular}{l *{5}{r@{\,}r}}
\toprule
 & \multicolumn{2}{c}{DM-1} & \multicolumn{2}{c}{DM-2} & \multicolumn{2}{c}{DM-3} & \multicolumn{2}{c}{DM-4} & \multicolumn{2}{c}{DM-5} \\
\midrule
Identity, $50$ tokens & 63.2 & {\scriptsize(8.0)} & 51.6 & {\scriptsize(16.6)} & 38.2 & {\scriptsize(31.4)} & 16.6 & {\scriptsize(58.6)} & 5.6 & {\scriptsize(75.2)} \\
Identity, $512$ tokens & 71.4 & {\scriptsize(0.4)} & 67.8 & {\scriptsize(0.2)} & 66.8 & {\scriptsize(2.4)} & 57.6 & {\scriptsize(16.0)} & 15.4 & {\scriptsize(61.6)} \\
Identity, $1{,}024$ tokens & 71.6 & {\scriptsize(0.0)} & 68.0 & {\scriptsize(0.0)} & 68.6 & {\scriptsize(0.2)} & 70.6 & {\scriptsize(2.6)} & 37.0 & {\scriptsize(34.6)} \\
Fixed program, $50$ tokens & 63.4 & {\scriptsize(1.0)} & 62.0 & {\scriptsize(0.8)} & 48.6 & {\scriptsize(5.2)} & 33.4 & {\scriptsize(9.2)} & 12.6 & {\scriptsize(16.2)} \\
\bottomrule
\end{tabular}
\end{table}

\subsection{LLaMA-3.2-3B across Difficulty Levels}
\label{sec:appendix_llama_levels}
On our split, LLaMA-3.2-3B's \Base($\tau{=}0$) accuracy does not fall with difficulty: it is lowest at DM-2 and highest at DM-5 (\cref{tab:baseline-dart-math}).
The cause is the question mix of our levels, not the model.
We sort MATH and GSM8K questions together by fail rate, so DM-1 is $92\%$ GSM8K and DM-5 only $5\%$ (\cref{sec:appendix_dart_math}).
LLaMA-3.2-3B is far less accurate on the GSM8K questions than on the MATH questions:
With \texttt{oneshot}, its \Base accuracy on the GSM8K questions of DM-1 to DM-5 is $15.6 / 10.0 / 7.2 / 6.9 / 4.1\%$, against $31.1 / 32.6 / 27.5 / 27.4 / 25.4\%$ on the MATH questions (all $2{,}000$ questions per level).
The same mix affects the diagnostics: for LLaMA-3.2-3B, the share of solvable questions that need recurrence or skipping falls with the level (\cref{fig:appendix_fig5b}), unlike for the other models.
GSM8K questions need recurrence more often than MATH questions (e.g.\ $83.4\%$ vs.\ $61.0\%$ at DM-3, seed $42$), and their share falls from $92\%$ to $5\%$.
The mix is only part of the explanation: within the MATH questions, the shares still fall (recurrence $60.6\%$ to $57.0\%$, skipping $51.5\%$ to $37.7\%$ from DM-1 to DM-5).
On the MATH-only PoLar split (\cref{sec:appendix_their_split}), neither effect appears: LLaMA-3.2-3B's \Base\ is within $2.2$pp of the published numbers on every level, and the share that needs recurrence stays about flat.

\subsection{\gls{mcts} Design Choices}
\label{sec:appendix_mcts_design}

\Cref{sec:reproduction_mcts_search} specifies our search reconstruction precisely; this subsection justifies each place it departs from a literal reading of \citet{li2026polar} (ICML version and the authors' own earlier prestudy \citep{li2025cola}, published under the pre-rename name \emph{CoLa}), with the evidence behind each choice.
We checked each choice against their released search code and say where it agrees and where it differs (\cref{sec:appendix_their_split_search}).

\paragraph{Their released search.}
We designed our search from the papers alone, before \citeauthor{li2026polar} released their \gls{mcts} code (without the search results themselves);
\cref{sec:appendix_their_split_search} compares the two.
We keep ours for the main results for two reasons.
First, it searches each question on its own.
Their code starts every question's search from the same random seed, so all questions begin with the same edits, and questions on which no program succeeds all get the same tree.
Second, it tries more alternatives per question.
On Qwen3-8B DM-1, it evaluates about $29$ different single-edit programs per question; theirs evaluates one.
The router thus sees several valid programs per question to learn from (\cref{sec:reproduction_router}).
With their search, the router's gain depends on the run, because the shared seed shapes the training labels of a whole level;
with several runs and selection on validation questions, it can still gain over \Base, by repairing the output format (\cref{sec:appendix_their_split_variance,sec:appendix_their_split_results}).

\Cref{alg:mcts,alg:selectexpand} give the full pseudocode, built around \gls{polar}'s own \gls{ucb} exploration term (\cref{eq:ucb_widening}).

\begin{algorithm}[t]
\caption{$\textsc{ProgramMCTS}(x,y,N_{\mathrm{sim}})$}
\label{alg:mcts}
\begin{algorithmic}[1]
\Require input $x$, ground truth $y$, layer count $D$, simulation budget $N_{\mathrm{sim}}$
\State $N_0 \gets \textsc{Node}(\emptyset)$ \Comment{root = identity program $\text{id}=(0,\dots,D-1)$}
\State $r(N_0) \gets r_{\text{id}} \gets \mathbbm{1}\{F_{\text{id}}(x)=y\}$, $V \gets 0$ \Comment{identity (\Base) is scored once, outside the budget; it counts as a program found, not as a simulation; every node starts with $v{=}R{=}0$}
\For{$n = 1$ \textbf{to} $N_{\mathrm{sim}}$}
    \State $N,\, \mathrm{chain} \gets \Call{SelectExpand}{N_0}$ \Comment{descend/widen by progressive widening, \cref{alg:selectexpand}}
    \State $\pi \gets \mathrm{Prog}(N)$ \Comment{tile $N$'s gaps with \textsf{keep}}
    \State $r \gets 0$ if $\pi$ skips every layer, else $\mathbbm{1}\{F_\pi(x) = y\}$ \Comment{the only model call this iteration}
    \State $r(N) \gets r$ \Comment{the program's own reward; $R$ and $v$ below only feed \gls{ucb}}
    \For{$M \in \mathrm{chain}$} \Comment{every node on the path just taken, root to $N$}
        \State $v(M) \gets v(M) + 1$
        \State $R(M) \gets R(M) + r$
    \EndFor
    \State $V \gets V + 1$ \Comment{global count, read by every node's UCB explore term}
\EndFor
\State \Return $\{(\mathrm{Prog}(N),\, r(N))\}$ over every evaluated $N$
\end{algorithmic}
\end{algorithm}

\begin{algorithm}[t]
\caption{$\textsc{SelectExpand}(N_0)$}
\label{alg:selectexpand}
\begin{algorithmic}[1]
\State $N \gets N_0$
\State $\mathrm{chain} \gets [N_0]$
\While{$\lnot\Call{CanExpand}{N}$ \textbf{and} $N$ has children} \Comment{capped or exhausted, and there's somewhere to go}
    \State $N \gets \arg\max_{C\,\in\,\mathrm{children}(N)} \mathrm{UCB}(C)$ \Comment{descend deeper, looking for room to expand}
    \State append $N$ to $\mathrm{chain}$
\EndWhile
\If{\Call{CanExpand}{N}} \Comment{room under $k(v(N))$, \cref{eq:ucb_widening}, true immediately at the root while $v$ is small}
    \State $N \gets \Call{Expand}{N,\ \textsc{RandomChoice}(\mathrm{untried}(N))}$
    \State append $N$ to $\mathrm{chain}$
\EndIf
\State \Return $(N, \mathrm{chain})$
\end{algorithmic}
\end{algorithm}

\paragraph{Reading the pseudocode.}
\Cref{alg:mcts,alg:selectexpand} use two helper definitions:
$$\textsc{CanExpand}(N) := \mathrm{untried}(N)\neq\emptyset \wedge |\mathrm{children}(N)|<k(v(N)),$$
and $\textsc{Expand}(N,a)$ returns the transposition-deduplicated child for edit set $N\cup\{a\}$, creating it only if no earlier placement order already has.
If that child already exists, \textsc{Expand} only adds the edge, and the simulation re-scores the existing program from the reward cache;
the same happens when selection stops at a node with no untried edit and no children.
Both count towards $N_{\mathrm{sim}}$ but are rare: at most $0.03\%$ of the simulations of any model's searches.
In both cases the edit leaves $\mathrm{untried}(N)$ and the chain is backed up as usual;
ties in \gls{ucb} go to the earliest-created child, whose order is random, since each node's untried edits are shuffled with the question's seed.
\Cref{alg:selectexpand} descends past a node only once it is already at its widening cap or has exhausted its actions; a node with room to grow expands immediately instead of being compared against its children, so budget is spent breadth-first near the root and only pushed deeper once shallow nodes saturate.
The listing omits our optional $\epsilon$-random-exploration mechanism (off by default, unused throughout this paper's results, covered below).

\paragraph{Progressive widening ($k(v)$ in \cref{eq:ucb_widening}).}
Given our action space's size, the root node needs some mechanism that stops it from spending the entire search budget expanding at depth 1.
\citeauthor{li2026polar}'s own $\epsilon$-random exploration mechanism (below) is their search's only source of breadth: their code applies it with probability $0.1$ at every level of the descent (\cref{sec:appendix_their_split_search}); we instead default to progressive widening \citep{couetoux2011continuous}, a standard technique for large or continuous action spaces \citep{browne2012survey}.
At the root, with $D{=}36$ (Qwen3-8B) and $K_{\max}{=}4$, there are already $276$ single-edit actions to try before a \emph{second} edit is even considered, against our own simulation budget of $N_{\mathrm{sim}}{=}200$ per question (matched to the preprint's stated value, see ``Other parameters matched to the papers' stated values'' below). Without $k(v)$ capping a node's children, the root's untried-action list would never empty within budget: every simulation would add one more single-edit sibling under the root, selection would never descend, and the tree would stay stuck at depth $1$, unable to discover the multi-edit programs (e.g.\ \textsf{skip} \emph{and} \textsf{repeat} together) that motivate \hyperref[finding:1]{\ftag{F1}}
(skip-and-repeat jointly beats either alone). $k(v)=\max(1,\lceil\alpha v^\beta\rceil)$ is the standard \emph{progressive widening} technique, capping branching factor as a slowly growing function of visit count so early search stays shallow while still allowing unbounded eventual depth. We keep $\beta{=}0.5$, matching that technique's usual $O(\sqrt{N})$ growth rate.
$\alpha{=}2.0$ has no canonical value in the literature either (domain-tuned everywhere we checked) and is not separately ablated here.

\paragraph{$\epsilon$-random exploration.}
\Cref{alg:selectexpand} omits this mechanism to avoid implying it is active in the results this paper reports.
When enabled ($\epsilon{>}0$), it sits inside the \textbf{while} loop, right after the loop condition (line~3): whenever $N$ has an untried edit and a uniform draw falls below $\epsilon$, $N$ is immediately expanded with a random untried edit and returned instead of continuing to the \gls{ucb}-argmax descent step (line~4).
It substitutes for descending further, and can only trigger at a node that failed \textsc{CanExpand} because of the widening cap, not because it ran out of actions.
The preprint \citep{li2025cola} states this mechanism as: ``the algorithm selects a random unexplored child node with probability $0.1$ instead of the one with the highest UCB score \ldots.''
For comparison, \citet{heakl2025drllm}'s own released \gls{mcts} implements a related mechanism, substituting a uniformly random \emph{existing} child for the \gls{ucb} argmax during descent.
Our default is $\epsilon{=}0$.

\paragraph{Any-order edit placement.}
We allow a node's edits to be placed in any order, rather than requiring each new edit to start after the previous one ends.
That requirement biases the search independently of the model: under a constant reward and $D{=}36$ (Qwen3-8B), $53\%$ of placed edits land in the late third of the layer stack against a $30.4\%$ uniform baseline, and a specific early-then-early two-edit program is $7\times$ less likely to be evaluated than a late-then-late one.
Equal edit sets reached in different orders are merged by the transposition table below.
We deliberately did not also allow overlapping/nested edits: the router's output space cannot represent them, so keeping the search space representable by its downstream consumer took priority, although their released code allows them ($0.3\%$ of its evaluated programs).

\paragraph{Transposition table (\gls{dag} instead of a tree) and the reward cache.}
\label{sec:appendix_reward_noise}
A direct consequence of the previous choice: once edits can be placed in any order, the same edit set is reachable via more than one path, so we canonicalize nodes by their sorted edit tuple and cache them rather than letting different placement orders become different nodes. Without this, one program would accumulate visit/reward statistics under several separate identities, splitting the statistics \gls{ucb} depends on and wasting budget re-discovering programs already evaluated under a different name.
A separate reward cache, keyed by the (input, program) pair rather than by node identity, additionally avoids a repeated model call whenever the same executed program recurs for the same input under a \emph{different} edit set (e.g.\ two edit sets that tile to the same program via \textsf{keep}-filled gaps), which the node-level transposition table above does not catch on its own.
This reward cache also matters for a second reason, beyond efficiency. Reward grades a greedy-decoded generation against ground truth via exact match on the \texttt{\textbackslash boxed\{...\}} span (\cref{sec:appendix_placeholder_echo}), and generation runs in batches; since bf16 addition is not associative,
the batched-matmul/attention kernel a row executes depends on the shape of the batch it happens to share, and greedy decoding argmaxes at every generated token, so this can occasionally flip which token is emitted, and the flip changes every later step's conditioning and can cascade into a different final answer, a full 0/1 reward-label flip caused purely by noise.
For the same reason, each search run is pinned to one GPU architecture (\cref{sec:appendix_hardware}).
Because the cache reuses a program's first-computed generation and reward label instead of regenerating it, the same program is never re-batched and re-flipped once it has been evaluated once.

\paragraph{Backpropagation and \gls{ucb} under multiple parents.}
Once transposition dedup is in place, a node can legitimately be reached from more than one immediate parent (edit-set $\{A,B\}$ from both $\{A\}$ and $\{B\}$), raising two questions.
First, \gls{ucb}: under our default global-$V$ reading (justified below), $N$'s explore term uses the shared global counter, not the specific parent, so $N$ has exactly one \gls{ucb} value regardless of which parent asks, so it stays well-defined under the multi-parent structure.
Even under the non-default alternative reading of $V$ as the immediate parent's own visit count rather than one shared global counter (the next paragraph justifies keeping the global default), no decision is corrupted, since $\arg\max_{C\in\mathrm{children}(P)}\mathrm{UCB}(C)$ only ever compares $N$ against its true siblings under one shared parent $P$.
Second, backpropagation: \cref{alg:mcts,alg:selectexpand} sidestep choosing which parent to propagate through. The $\mathrm{chain}$ variable is simply the list of nodes this simulation's selection loop just walked, and backpropagation replays exactly that list.
Each chain that reaches $N$ updates $v(N)/R(N)$ independently and additively, like any other node's statistics. A node's back-pointer field is consequently left unset (used only by an alternative, non-default selection mode where every node has a single parent by construction).

\paragraph{Global $V$ versus the parent's own visit count.}
Both \citet{li2025cola,li2026polar}
state the \gls{ucb} explore term's $V$ as ``the total number of simulations'', which we read literally as one counter shared by the whole tree, incremented by every completed simulation, not, as an easy misreading, the specific parent's own visit count at the point of comparison. The two readings coincide in a plain, single-path tree, but diverge once the \gls{dag} structure above lets a node have more than one parent. Their released code uses the root's visit count plus one, which counts the same simulations, so our reading matches it. The textbook-local variant remains available as a configuration flag.

\paragraph{Hierarchical descent versus a flat, whole-tree selection reading.}
The Selection line in \citeauthor{li2026polar}'s
appendix reads ``traverse tree using UCB to reach a \emph{leaf} program'' (the preprint \citep{li2025cola}
omits ``leaf''), which could describe either our hierarchical, per-node \gls{ucb} descent (\cref{alg:selectexpand}) or a flatter rule comparing every still-expandable node in the whole tree directly and expanding whichever wins, regardless of depth.
We implemented and ran both: the flat variant produced substantially deeper, near-constant-depth trees on complete production data (median edit-count $10$, max $26$), sitting awkwardly against \hyperref[finding:4]{\ftag{F4}} (valid programs predominantly short).
We also checked \citeauthor{heakl2025drllm}'s own \gls{mcts} implementation (\texttt{data\_generation.py}) as a possible tie-breaker.
Its \texttt{MCTSNode.is\_leaf()} is \texttt{len(self.children) == 0}, so a node can never be selected again once it gets its first child, and a from-scratch reproduction of their exact selection loop (synthetic reward, $20$ runs) confirms every node in their tree gets at most one child, a strict linear chain, kept from visibly degenerating only by their search's own early-exit stopping rule. This rules out DR.LLM's code as evidence for the flat reading, though the comparison is still informative: our hierarchical mode's progressive-widening-gated notion of ``not yet fully expanded'' is the standard \gls{uct} definition of \emph{leaf}, which neither DR.LLM's code nor our flat-selection reading implements. Their released code also descends level by level by \gls{ucb} (\cref{sec:appendix_their_split_search}).
We therefore default to hierarchical descent throughout this paper.

\paragraph{DR.LLM is not directly comparable.}
There are two further differences from \citeauthor{heakl2025drllm}'s setup.
First, their search terminates a question as soon as it finds a program both shorter than the current best and correct, rather than exhausting the full simulation budget the way \cref{alg:mcts} always does, consistent with the strict, at-most-one-child-per-node chain noted above.
Second, their reward for DART-Math questions is a continuous, partial-credit symbolic grader, not the binary correct/incorrect signal \citeauthor{li2026polar} and we both use (valid iff reward $1$, \cref{sec:reproduction_mcts_search}).
Both differences matter beyond implementation detail.
An early-exit search structurally under-counts how many equally good \emph{alternative} programs exist per question (exactly the diversity property \cref{sec:reproduction_router}'s multi-program labelling depends on), and a continuous reward changes what ``valid'' means, so DR.LLM's own reported search statistics are not directly comparable to ours or \citeauthor{li2026polar}'s without accounting for both.

\paragraph{Every tree gets its own random seed.} \citeauthor{li2026polar}'s released code shares one seed across all questions' search trees (\texttt{random.seed(42)} at the start of each).
We instead derive a distinct, deterministic seed per question (a hash of a base seed and the question's id), so no two inputs' trees ever make the same sequence of random expansion/tie-breaking choices.
This matters, because an earlier shared-seed version measurably collapsed search diversity across \emph{different questions}.
Per-question seeding is on by default throughout.

\paragraph{Other parameters matched to the papers' stated values, not tuned.} $\lambda{=}5.0$ (length-penalty weight),
the repeat-count cap $\rho{-}1\le4$ (an edit's total execution count $\rho\le5$, \cref{sec:reproduction_mcts_search}), and the simulation budget $N_{\mathrm{sim}}{=}200$ per question are all stated explicitly in the preprint's Appendix~B.
We match them because the paper states them, not because we tuned them.
Their released code uses other values for three of them (100 simulations, at most two executions of a layer, and programs of $24$ to $44$ layers for Qwen3-8B, \cref{sec:appendix_their_split_search}).
Our main track keeps the papers' values.
The \gls{ucb} exploration constant $c{=}\sqrt{2}$ is stated in neither paper, but it is the standard \gls{uct} default \citep{kocsis2006bandit}, and their code uses it too.

\paragraph{Taken together.} Every choice above, whether it fills a gap in their paper or departs from their code, is backed by an ablation, a structural argument or their released code, and every non-default alternative remains available as a configuration flag in our released code.

\subsection{Accelerating \gls{mcts} Reward Computation}
\label{sec:appendix_masked_batch_kv}

Evaluating a candidate program's reward during \gls{mcts} means generating and grading an answer for every program encountered in a round of tree expansion, each of which may execute a different subset, order, or repetition of layers.
A naive implementation evaluates one program at a time: apply the program to the model (physically reordering, skipping, or repeating its layer modules), then run an ordinary cached \texttt{generate()} call over that round's questions. Once a search round has diversified past a handful of trees, distinct programs approach one per tree and per-program question batches collapse to size $1$.
Generation then becomes launch-overhead-bound rather than compute-bound and severely under-utilizes the GPU.
\citeauthor{li2026polar}'s released search works this way, generating the answer for one question and one program per call;
they report that their full search ran for about a day on $32$ A100 GPUs\footnote{\url{https://github.com/tianyi-lab/PoLar/issues/4}}.

\paragraph{Masked-batch decoding.} This mechanism removes the bottleneck by sharing forward passes across an entire round's rows regardless of which distinct program each row uses. Rather than mutating the model's layer list per program, every row's own layer-execution path is looked up directly.
At each decode step and for each transformer layer, all rows whose next required layer matches are gathered into one batched forward call, executed once, and scattered back into a shared per-row residual-stream state, so a layer's weights are read once per group of co-scheduled rows, not once per row. Grouping uses a greedy largest-group-first schedule: each step buckets rows by their next required layer and fires only the largest bucket before regrouping, trading a little wait time for a few early-ready rows against fewer, larger forward calls.
A configurable group-size cap bounds peak memory when many rows converge on the same layer, and each round is split into length-sorted buckets so one unusually long prompt only forces padding on its own bucket.

\paragraph{A \gls{kv}-cached extension.}
We originally ran masked-batch decoding without caching, recomputing the full prefix at every decode step; we added a real per-row, per-path-position key/value cache, running the identical algorithm as single-program cached generation (prefill once, then decode one token per step through the cache) while still batching across distinct programs, purely to speed up search further.
Because a \textsf{repeat} operation revisits the same physical layer at two different positions in a program's path, each (row, path-position) pair gets its own cache slot rather than one slot per layer, mirroring how our serial execution engine treats each path position as an independent shallow copy of the layer. We validate bit-identical output to single-program cached generation whenever no other row shares a forward call with it (row count $1$, including under production left-padding), on different model families and multiple GPU architectures.
Its residual disagreement with the serial path once real batching is active is indistinguishable from the batch-shape noise already documented in \cref{sec:appendix_reward_noise}.

\paragraph{Validated at production scale.}
On Qwen3-8B's complete DART-Math difficulty-$1$ split ($2{,}000$ questions), \gls{kv}-cached and uncached masked-batch solved counts differ by exactly $1$ question ($1{,}776$ vs.\ $1{,}777$) and shorter-than-identity rates by $0.0012$, inside the established noise floor, while the \gls{kv}-cached version is $1.10\times$ faster wall-clock ($8$h$14$m vs.\ $9$h$04$m).
This was enough evidence to run every split and every model in this paper's \gls{mcts} with both masked-batch decoding and its \gls{kv}-cached extension turned on.
Our port of their search (\cref{sec:appendix_their_split_search}) runs with both mechanisms as well.
On Qwen3-8B DM-1 of the PoLar split ($565$ questions, one A100), it needs about $5\times$ fewer GPU-hours than their code and gives the same results up to the floating-point noise of batched decoding (\cref{sec:appendix_reward_noise}).

\subsection{Router Training Recipe and Identity Collapse}
\label{sec:appendix_router_recipe}

\Cref{sec:reproduction_router} briefly describes how we turn \gls{mcts} output into router training examples, and this subsection traces each architecture and training-data choice to its source;
the exact loss terms and beam search are in our released code.
It also documents every anti-collapse lever we tried, and closes with our best mechanistic account of why none of them gains over \Base, even where they break the collapse.
Every item below is verified directly against \cite{li2026polar}'s own released code at \url{https://github.com/tianyi-lab/PoLar}.
We sweep several recipes (data-treatment/loss combinations; ``The full recipe sweep'' paragraph below defines the full grid) and name each by concatenating the two, e.g.\ Drop-CE.
``Selecting one recipe'' explains why we use Drop-CE everywhere else in this paper, including \cref{tab:router-results-strict-ce}.
Unless a paragraph states a model explicitly, every number in the identity-collapse investigation below, up to ``Selecting one recipe across models and difficulties,'' is Qwen3-8B on DART-Math DM-$1$; that final paragraph is where the investigation's conclusions are checked against all six models and all difficulties.

\begin{table}[tb]
\centering
\small
\caption{The router's pass@$1$ equals \Base($\tau{=}0$) exactly, since it always predicts identity; only its lower-ranked programs gain (\cref{sec:router_collapse_results}).
\Router\ pass@$1$--$5$ with online execution, Drop-CE, averaged over the $5$ difficulties and the three search runs; \cref{tab:main_results} gives pass@$5$ per difficulty.}
\label{tab:router-results-strict-ce}
\begin{tabular}{lccccc}
\toprule
Model & pass@$1$ & pass@$2$ & pass@$3$ & pass@$4$ & pass@$5$ \\
\midrule
Qwen1.5-MoE-A2.7B & $20.2\%$ & $21.7\%$ & $24.2\%$ & $28.4\%$ & $31.3\%$ \\
Qwen2.5-3B & $22.0\%$ & $25.5\%$ & $28.7\%$ & $32.1\%$ & $35.1\%$ \\
Qwen2.5-7B & $37.8\%$ & $40.8\%$ & $44.3\%$ & $47.4\%$ & $50.0\%$ \\
Qwen3-8B & $33.5\%$ & $35.8\%$ & $38.5\%$ & $41.4\%$ & $44.3\%$ \\
Qwen3-32B & $35.0\%$ & $41.2\%$ & $45.6\%$ & $48.7\%$ & $51.3\%$ \\
LLaMA-3.2-3B & $18.9\%$ & $20.6\%$ & $22.7\%$ & $25.5\%$ & $28.2\%$ \\
\bottomrule
\end{tabular}
\end{table}

\paragraph{Evaluation protocol: online execution, not cache-only.} \gls{polar}'s evaluation script scores a router's top-$k$ candidates correct if any is either (a) present in the offline \gls{mcts} cache of certified-valid programs for that question, or (b) actually executed online against the target model and found correct, a fallback for candidates the offline search never encountered.
Their released online-execution fallback raises an uncaught error whenever reached, so we initially evaluated using only the cache-only path (a); this understates the router, since the offline cache only contains programs a finite-budget search happened to explore and certify.
Every result in this section therefore uses online execution, never the cache-only proxy.

\paragraph{Architecture hyperparameters: identical to \gls{polar}'s own defaults, not swept.}
\Cref{sec:reproduction_router} describes the router's architecture qualitatively (frozen embedding backbone, per-layer query cross-attention, a small transformer over the layer dimension, two linear heads).
Our implementation uses working dimension $d_{\text{model}}{=}256$, $4$ cross-attention/encoder heads, a $2$-layer cross-layer transformer encoder (feedforward width $4d_{\text{model}}{=}1024$), dropout $0.1$, and a $512$-token question-truncation limit.
All five match \citeauthor{li2026polar}'s own \texttt{polar/model.py}/\texttt{polar/train.py} exactly.
The frozen encoder is \texttt{Qwen/Qwen3-Embedding-0.6B} (hidden size $1024$), projected into $d_{\text{model}}$ by a single learned linear layer.
Counting only trainable parameters matches \citeauthor{li2026polar}'s own reported $\approx 2.1$M.
That same section also describes segmentation probabilities being thresholded into segment boundaries at inference, without giving a number: a layer starts a new segment once its predicted segmentation probability reaches $0.5$, the code's own fixed default, and the only value on this list we later vary (\emph{Segmentation-threshold calibration} below, one of the anti-collapse levers).
The router's separate training-time hyperparameters (learning rate, batch size, epoch count), unlike the architecture above, are swept rather than fixed to a single default.
\emph{The full recipe sweep} and \emph{Selecting one recipe across models and difficulties} below give the grid and the hyperparameter selection criterion, validation loss, without tabulating every selected value individually.

\paragraph{Multi-program labelling.}
By default, every valid \gls{mcts}-discovered program for a question becomes its own training example, capped at $50$ programs per question, matching \citeauthor{li2026polar}'s own code default exactly; no router-training run in this paper overrides it.

\paragraph{Router training-data handling for \textsf{repeat} segments with multiplicity greater than two.}
\label{sec:appendix_repeat_handling}
\gls{polar}'s router represents a \textsf{repeat} operation as one class among three (skip/keep/repeat) with no encoded repetition count.
Its decoder always realizes a predicted \textsf{repeat} as exactly two executions of the segment, regardless of how many times the original \gls{mcts}-discovered program executed it.
Their released label-construction code only ever matches a \textsf{repeat} segment against exactly two consecutive repetitions.
Any \gls{mcts}-discovered program containing a segment executed three or more times fails to parse under their grammar and is silently dropped from their training set in its entirety, not merely the offending segment.

Our own \gls{mcts} allows repetition counts up to $\rho{-}1=4$ beyond the first execution ($\rho\le5$ total executions, again matching \citeauthor{li2026polar}'s stated search-space bound).
Under Drop this has a cost: across models, Drop removes $53$--$69\%$ of the non-identity valid programs, and on $10$--$22\%$ of the questions that have such a program, all of them.
The $50$-program cap does not make up for this, since after Drop at most $1.3\%$ of training questions still have more than $50$ programs, so about half of the training examples are lost.
A segment executed more than twice is common in what our search actually finds:
a majority of discovered \textsf{repeat}-containing programs ($59$--$76\%$ per model) have at least one segment that is repeated more than once.
This seems to conflict with \citeauthor{li2026polar}'s finding that valid execution programs ``typically require at most a single recurrence per segment,''
but it depends on which programs are counted, which they do not state:
for the shortest valid program per question, our search agrees with them (\hyperref[finding:4]{\ftag{F4}}), whereas over all valid programs, as here, about half of the repeated segments are repeated more than once.

Given a training program contains such a segment, three treatments are possible, and we compare all three under a single, otherwise-identical training and evaluation protocol:
\begin{itemize}
    \item \textbf{Drop.} Discard the entire program from training whenever any \textsf{repeat} segment's count is not exactly two, exactly reproducing \gls{polar}'s own released behavior.
    \item \textbf{Crop.} Retain the program and supervise the segment as an ordinary \textsf{repeat} label, ignoring the true count, equivalent to truncating it to two executions without checking whether that shorter program is still correct. This keeps roughly twice as many training programs.
    \item \textbf{Crop and Re-verify.} Truncate the segment to exactly two executions, \emph{re-execute this shortened program online against the real question and ground truth}, and keep it as a training label only if it is still correct, discard it otherwise. Considerably more expensive than the other two treatments, which need no extra model calls.
\end{itemize}
On the DM-1 questions of our DART-Math split for Qwen3-8B, re-verification of all $18{,}519$ training programs with a \textsf{repeat} count other than two finds that $15{,}582$ ($84.1\%$) remain correct once cropped to two executions and re-run online against the real question and ground truth, while $2{,}937$ ($15.9\%$) do not and are discarded entirely, meaning \emph{Crop} trains on about one label in six that does not hold under the router's own executable semantics.

\paragraph{The full recipe sweep.} Alongside this data-treatment axis (Drop/Crop/Crop-and-Re-verify), we also sweep the loss function: standard cross-entropy (CE), \citeauthor{li2026polar}'s own default, against a class-balanced focal loss, motivated by \citeauthor{heakl2025drllm}'s own use of focal loss to handle the same kind of \textsf{skip}/\textsf{keep}/\textsf{repeat} class imbalance in their router.
Every data-treatment/loss combination is itself swept over learning rate $\in\{1\mathrm{e}{-4}, 3\mathrm{e}{-4}, 5\mathrm{e}{-4}, 8\mathrm{e}{-4}, 1\mathrm{e}{-3}, 3\mathrm{e}{-3}\}$, batch size $\in\{32, 128, 256\}$, and epoch count $\in\{3, 10\}$ with the AdamW optimizer, exactly matching \citeauthor{li2026polar}'s own stated grid ($6\times3\times2=36$ combinations, the ``$36$-combination hyperparameter grid'' referenced throughout this section).

\paragraph{Training-data composition, full breakdown.} \Cref{tab:router-training-data-composition} reports per-model averages: \emph{Identity \%} is the share of training program-examples that are the identity (full-depth) program, and \emph{Keep \%} is the share of per-segment operation labels that are \textsf{keep} rather than \textsf{skip} or \textsf{repeat}.
The $50$-program cap above applies \emph{per question}, not to a category's total count across the split: identity, for instance, is just one label among a question's own capped set, so how often it shows up in total depends on how many different questions it happens to solve, not on the cap itself.
\Cref{tab:router-training-data-composition-full} gives the complete per-difficulty numbers behind it, all $6$ models $\times$ all $5$ DART-Math difficulties, computed on the $1{,}250$-question train splits; values are mean $\pm$ sample standard deviation over the three search runs.
\emph{No-id \%} (no valid identity) is the share of solvable questions whose identity program is not even a valid execution (the trigger condition for the anti-collapse \textsf{keep}-probability penalty below).
It follows each model's own \Base\ accuracy: the lower the accuracy of a level, the higher the share ($36$--$76\%$ at DM-$1$, $60$--$92\%$ at DM-$5$).
Identity-example and \textsf{keep}-label share are comparatively stable within a model across difficulties, but vary meaningfully \emph{across} models ($1.5$--$7.9\%$ identity, $80.0$--$89.7\%$ \textsf{keep}), consistent with the per-model recipe variance documented below (Qwen3-32B's own preference for Crop over Drop), not a single universal constant.

\begin{table}[tb]
\centering
\footnotesize
\caption{Identity is only $1.5$--$7.9\%$ of training examples in every cell, but \textsf{keep} makes up $80$--$90\%$ of segment labels; the lower a level's accuracy, the more solvable questions have no valid identity program.
Full per-difficulty version of \cref{tab:router-training-data-composition} (Drop treatment, $1{,}250$-question train slice; the composition depends only on the data treatment, not the loss function). \emph{Solvable} = questions with $\geq\!1$ \gls{mcts}-verified-correct program; \emph{No-id \%} = share of those with no valid identity program at all; \emph{examples} = the resulting number of training program-examples; \emph{id \%}/\emph{Keep \%} as \cref{tab:router-training-data-composition}.}
\label{tab:router-training-data-composition-full}
\begin{tabular}{llr@{\,}c@{\,}rr@{\,}c@{\,}lr@{\,}c@{\,}rr@{\,}c@{\,}lr@{\,}c@{\,}l}
\toprule
Model & DM & \multicolumn{3}{c}{Solvable} & \multicolumn{3}{c}{No-id \%} & \multicolumn{3}{c}{Examples} & \multicolumn{3}{c}{Id \%} & \multicolumn{3}{c}{Keep \%} \\
\midrule
\multirow{5}{*}{Qwen1.5-MoE-A2.7B}
 & 1 & $929$ & {\scriptsize$\pm$} & {\scriptsize 6} & $68.2$ & {\scriptsize$\pm$} & {\scriptsize 0.1} & $4{,}175$ & {\scriptsize$\pm$} & {\scriptsize 45} & $7.0$ & {\scriptsize$\pm$} & {\scriptsize 0.1} & $80.1$ & {\scriptsize$\pm$} & {\scriptsize 0.1} \\
 & 2 & $873$ & {\scriptsize$\pm$} & {\scriptsize 11} & $68.2$ & {\scriptsize$\pm$} & {\scriptsize 0.2} & $4{,}112$ & {\scriptsize$\pm$} & {\scriptsize 77} & $6.7$ & {\scriptsize$\pm$} & {\scriptsize 0.2} & $80.0$ & {\scriptsize$\pm$} & {\scriptsize 0.1} \\
 & 3 & $819$ & {\scriptsize$\pm$} & {\scriptsize 13} & $73.3$ & {\scriptsize$\pm$} & {\scriptsize 0.9} & $3{,}323$ & {\scriptsize$\pm$} & {\scriptsize 117} & $6.6$ & {\scriptsize$\pm$} & {\scriptsize 0.3} & $80.1$ & {\scriptsize$\pm$} & {\scriptsize 0.2} \\
 & 4 & $803$ & {\scriptsize$\pm$} & {\scriptsize 9} & $67.9$ & {\scriptsize$\pm$} & {\scriptsize 0.3} & $3{,}543$ & {\scriptsize$\pm$} & {\scriptsize 67} & $7.3$ & {\scriptsize$\pm$} & {\scriptsize 0.2} & $80.2$ & {\scriptsize$\pm$} & {\scriptsize 0.0} \\
 & 5 & $544$ & {\scriptsize$\pm$} & {\scriptsize 8} & $73.2$ & {\scriptsize$\pm$} & {\scriptsize 0.4} & $1{,}855$ & {\scriptsize$\pm$} & {\scriptsize 86} & $7.9$ & {\scriptsize$\pm$} & {\scriptsize 0.4} & $80.5$ & {\scriptsize$\pm$} & {\scriptsize 0.1} \\
\midrule
\multirow{5}{*}{Qwen2.5-3B}
 & 1 & $1{,}032$ & {\scriptsize$\pm$} & {\scriptsize 1} & $61.2$ & {\scriptsize$\pm$} & {\scriptsize 0.4} & $8{,}562$ & {\scriptsize$\pm$} & {\scriptsize 71} & $4.4$ & {\scriptsize$\pm$} & {\scriptsize 0.0} & $84.0$ & {\scriptsize$\pm$} & {\scriptsize 0.1} \\
 & 2 & $1{,}015$ & {\scriptsize$\pm$} & {\scriptsize 19} & $62.8$ & {\scriptsize$\pm$} & {\scriptsize 0.5} & $8{,}028$ & {\scriptsize$\pm$} & {\scriptsize 135} & $4.3$ & {\scriptsize$\pm$} & {\scriptsize 0.0} & $83.9$ & {\scriptsize$\pm$} & {\scriptsize 0.1} \\
 & 3 & $914$ & {\scriptsize$\pm$} & {\scriptsize 23} & $66.4$ & {\scriptsize$\pm$} & {\scriptsize 0.8} & $6{,}622$ & {\scriptsize$\pm$} & {\scriptsize 28} & $4.4$ & {\scriptsize$\pm$} & {\scriptsize 0.1} & $83.9$ & {\scriptsize$\pm$} & {\scriptsize 0.1} \\
 & 4 & $863$ & {\scriptsize$\pm$} & {\scriptsize 11} & $72.9$ & {\scriptsize$\pm$} & {\scriptsize 0.2} & $5{,}684$ & {\scriptsize$\pm$} & {\scriptsize 193} & $3.8$ & {\scriptsize$\pm$} & {\scriptsize 0.1} & $83.6$ & {\scriptsize$\pm$} & {\scriptsize 0.1} \\
 & 5 & $541$ & {\scriptsize$\pm$} & {\scriptsize 3} & $81.5$ & {\scriptsize$\pm$} & {\scriptsize 0.1} & $2{,}489$ & {\scriptsize$\pm$} & {\scriptsize 24} & $3.8$ & {\scriptsize$\pm$} & {\scriptsize 0.1} & $83.5$ & {\scriptsize$\pm$} & {\scriptsize 0.1} \\
\midrule
\multirow{5}{*}{Qwen2.5-7B}
 & 1 & $1{,}119$ & {\scriptsize$\pm$} & {\scriptsize 5} & $44.2$ & {\scriptsize$\pm$} & {\scriptsize 0.2} & $11{,}204$ & {\scriptsize$\pm$} & {\scriptsize 218} & $5.1$ & {\scriptsize$\pm$} & {\scriptsize 0.1} & $80.8$ & {\scriptsize$\pm$} & {\scriptsize 0.1} \\
 & 2 & $1{,}097$ & {\scriptsize$\pm$} & {\scriptsize 1} & $44.9$ & {\scriptsize$\pm$} & {\scriptsize 0.3} & $11{,}023$ & {\scriptsize$\pm$} & {\scriptsize 89} & $5.1$ & {\scriptsize$\pm$} & {\scriptsize 0.1} & $80.7$ & {\scriptsize$\pm$} & {\scriptsize 0.0} \\
 & 3 & $1{,}006$ & {\scriptsize$\pm$} & {\scriptsize 2} & $54.7$ & {\scriptsize$\pm$} & {\scriptsize 0.2} & $8{,}770$ & {\scriptsize$\pm$} & {\scriptsize 104} & $4.9$ & {\scriptsize$\pm$} & {\scriptsize 0.1} & $80.6$ & {\scriptsize$\pm$} & {\scriptsize 0.1} \\
 & 4 & $958$ & {\scriptsize$\pm$} & {\scriptsize 2} & $58.3$ & {\scriptsize$\pm$} & {\scriptsize 0.1} & $8{,}038$ & {\scriptsize$\pm$} & {\scriptsize 86} & $4.7$ & {\scriptsize$\pm$} & {\scriptsize 0.1} & $80.4$ & {\scriptsize$\pm$} & {\scriptsize 0.1} \\
 & 5 & $639$ & {\scriptsize$\pm$} & {\scriptsize 10} & $69.0$ & {\scriptsize$\pm$} & {\scriptsize 0.5} & $3{,}716$ & {\scriptsize$\pm$} & {\scriptsize 52} & $5.2$ & {\scriptsize$\pm$} & {\scriptsize 0.1} & $80.6$ & {\scriptsize$\pm$} & {\scriptsize 0.1} \\
\midrule
\multirow{5}{*}{Qwen3-8B}
 & 1 & $1{,}112$ & {\scriptsize$\pm$} & {\scriptsize 4} & $45.1$ & {\scriptsize$\pm$} & {\scriptsize 0.2} & $12{,}257$ & {\scriptsize$\pm$} & {\scriptsize 164} & $4.4$ & {\scriptsize$\pm$} & {\scriptsize 0.1} & $83.9$ & {\scriptsize$\pm$} & {\scriptsize 0.0} \\
 & 2 & $1{,}084$ & {\scriptsize$\pm$} & {\scriptsize 9} & $47.4$ & {\scriptsize$\pm$} & {\scriptsize 0.4} & $11{,}453$ & {\scriptsize$\pm$} & {\scriptsize 127} & $4.4$ & {\scriptsize$\pm$} & {\scriptsize 0.0} & $83.9$ & {\scriptsize$\pm$} & {\scriptsize 0.1} \\
 & 3 & $970$ & {\scriptsize$\pm$} & {\scriptsize 12} & $58.0$ & {\scriptsize$\pm$} & {\scriptsize 0.5} & $8{,}558$ & {\scriptsize$\pm$} & {\scriptsize 126} & $4.4$ & {\scriptsize$\pm$} & {\scriptsize 0.0} & $83.8$ & {\scriptsize$\pm$} & {\scriptsize 0.1} \\
 & 4 & $920$ & {\scriptsize$\pm$} & {\scriptsize 6} & $58.8$ & {\scriptsize$\pm$} & {\scriptsize 0.3} & $7{,}686$ & {\scriptsize$\pm$} & {\scriptsize 71} & $4.5$ & {\scriptsize$\pm$} & {\scriptsize 0.1} & $83.7$ & {\scriptsize$\pm$} & {\scriptsize 0.0} \\
 & 5 & $592$ & {\scriptsize$\pm$} & {\scriptsize 7} & $75.7$ & {\scriptsize$\pm$} & {\scriptsize 0.3} & $3{,}066$ & {\scriptsize$\pm$} & {\scriptsize 32} & $4.4$ & {\scriptsize$\pm$} & {\scriptsize 0.1} & $83.5$ & {\scriptsize$\pm$} & {\scriptsize 0.1} \\
\midrule
\multirow{5}{*}{Qwen3-32B}
 & 1 & $1{,}213$ & {\scriptsize$\pm$} & {\scriptsize 4} & $35.8$ & {\scriptsize$\pm$} & {\scriptsize 0.2} & $15{,}428$ & {\scriptsize$\pm$} & {\scriptsize 271} & $2.9$ & {\scriptsize$\pm$} & {\scriptsize 0.1} & $89.7$ & {\scriptsize$\pm$} & {\scriptsize 0.0} \\
 & 2 & $1{,}202$ & {\scriptsize$\pm$} & {\scriptsize 6} & $45.3$ & {\scriptsize$\pm$} & {\scriptsize 0.3} & $14{,}142$ & {\scriptsize$\pm$} & {\scriptsize 109} & $3.0$ & {\scriptsize$\pm$} & {\scriptsize 0.1} & $89.6$ & {\scriptsize$\pm$} & {\scriptsize 0.0} \\
 & 3 & $1{,}143$ & {\scriptsize$\pm$} & {\scriptsize 8} & $60.0$ & {\scriptsize$\pm$} & {\scriptsize 0.3} & $11{,}552$ & {\scriptsize$\pm$} & {\scriptsize 72} & $2.8$ & {\scriptsize$\pm$} & {\scriptsize 0.1} & $89.5$ & {\scriptsize$\pm$} & {\scriptsize 0.0} \\
 & 4 & $1{,}081$ & {\scriptsize$\pm$} & {\scriptsize 9} & $79.5$ & {\scriptsize$\pm$} & {\scriptsize 0.2} & $8{,}556$ & {\scriptsize$\pm$} & {\scriptsize 50} & $2.0$ & {\scriptsize$\pm$} & {\scriptsize 0.0} & $89.3$ & {\scriptsize$\pm$} & {\scriptsize 0.0} \\
 & 5 & $730$ & {\scriptsize$\pm$} & {\scriptsize 5} & $92.1$ & {\scriptsize$\pm$} & {\scriptsize 0.1} & $3{,}444$ & {\scriptsize$\pm$} & {\scriptsize 32} & $1.5$ & {\scriptsize$\pm$} & {\scriptsize 0.1} & $89.2$ & {\scriptsize$\pm$} & {\scriptsize 0.0} \\
\midrule
\multirow{5}{*}{LLaMA-3.2-3B}
 & 1 & $871$ & {\scriptsize$\pm$} & {\scriptsize 23} & $76.2$ & {\scriptsize$\pm$} & {\scriptsize 0.6} & $4{,}506$ & {\scriptsize$\pm$} & {\scriptsize 185} & $4.5$ & {\scriptsize$\pm$} & {\scriptsize 0.2} & $80.4$ & {\scriptsize$\pm$} & {\scriptsize 0.1} \\
 & 2 & $828$ & {\scriptsize$\pm$} & {\scriptsize 6} & $73.5$ & {\scriptsize$\pm$} & {\scriptsize 0.2} & $4{,}718$ & {\scriptsize$\pm$} & {\scriptsize 49} & $4.5$ & {\scriptsize$\pm$} & {\scriptsize 0.0} & $80.6$ & {\scriptsize$\pm$} & {\scriptsize 0.1} \\
 & 3 & $770$ & {\scriptsize$\pm$} & {\scriptsize 3} & $76.4$ & {\scriptsize$\pm$} & {\scriptsize 0.1} & $3{,}886$ & {\scriptsize$\pm$} & {\scriptsize 57} & $4.5$ & {\scriptsize$\pm$} & {\scriptsize 0.1} & $80.6$ & {\scriptsize$\pm$} & {\scriptsize 0.2} \\
 & 4 & $831$ & {\scriptsize$\pm$} & {\scriptsize 12} & $65.8$ & {\scriptsize$\pm$} & {\scriptsize 0.5} & $5{,}605$ & {\scriptsize$\pm$} & {\scriptsize 120} & $4.9$ & {\scriptsize$\pm$} & {\scriptsize 0.1} & $80.6$ & {\scriptsize$\pm$} & {\scriptsize 0.1} \\
 & 5 & $745$ & {\scriptsize$\pm$} & {\scriptsize 4} & $59.7$ & {\scriptsize$\pm$} & {\scriptsize 0.2} & $4{,}355$ & {\scriptsize$\pm$} & {\scriptsize 58} & $6.8$ & {\scriptsize$\pm$} & {\scriptsize 0.1} & $81.8$ & {\scriptsize$\pm$} & {\scriptsize 0.0} \\
\bottomrule
\end{tabular}
\end{table}

The levers below are Drop-CE unless a paragraph says otherwise.

\paragraph{Segmentation-threshold calibration: rejected, and it isolates where the collapse actually sits.}
Sweeping the segmentation threshold introduced above (\emph{Architecture hyperparameters}, $0.5$ by default) down to $0.05$ finds the segmentation head itself healthy, since at the default threshold every test question already has a segment boundary fire (per-question maximum boundary probability: median $0.863$).
The collapse instead sits in the \emph{operation} head: its argmax operation is \textsf{keep} for $98.3\%$ of layers, \textsf{repeat} for $1.4\%$ and \textsf{skip} for $0.3\%$.
Thresholds from $0.10$ to $0.50$ therefore decode to identity on every test question.
Only $0.05$ yields non-identity programs ($32.4\%$ of test questions), without a gain:
pass@$1$ is $47.6\%$ vs.\ $48.2\%$ for \Base, with $10$ questions rescued and $13$ lost, and pass@$5$ falls from $61.4\%$ to $57.6\%$.
This directly motivates the loss-side levers below: if the operation head's own class predictions are the bottleneck, the fix has to act on that head's training signal, not on the segmentation decode rule.

\paragraph{Down-weighting identity.}
When a question has both its identity (full-depth) program and at least one strictly shorter valid alternative among its labelled examples, we down-weight the identity program's own loss contribution to $0.30\times$ (every other program, including long non-identity ones, stays at $1.0\times$).
This directly implements the paper's own stated justification.
While the paper never gives a number for it, $0.30$ is not our own tuning choice: it is the exact value in \citeauthor{li2026polar}'s own README's worked training command.
Lowering this weight further does not help: at $0.2$, $0.1$, $0.05$ and $0$, every test question still gets identity (pass@$5$ $60.0$--$61.8\%$).

\paragraph{Per-sample weight normalization.} Each program-example is additionally scaled by $1/n_{\text{prog}}$ for its question, where $n_{\text{prog}}$ is that question's own program count after the $50$-program cap above, so a question with many valid \gls{mcts}-discovered programs does not dominate the loss over one with only a few.
Same provenance pattern as the down-weight: absent from the paper text, off by default in \citeauthor{li2026polar}'s own code, and present only in their README's worked example, which we follow.

\paragraph{\textsf{keep} segments are chunked to $K_{\max}$ too.} Converting a raw \gls{mcts} program into router training targets (segmentation + per-segment operation labels) caps every segment type (including \textsf{keep}) at $K_{\max}{=}4$ layers by default, matching what the router's decoding procedure enforces at inference. This is a representation choice at the program$\to$target conversion step, not something the search itself imposes: \gls{mcts}'s action space (\cref{sec:methodology}) bounds \textsf{skip}/\textsf{repeat} block length only.
\textsf{keep} is never an explicit search action (it is the identity baseline any edit is placed on top of, \cref{sec:reproduction_mcts_search}), so a discovered program can contain an arbitrarily long uninterrupted \textsf{keep} run before conversion.

\paragraph{Length-preference reweighting.}
\citeauthor{li2026polar}'s code additionally supports a length-preference reweighting policy mode ($\exp(-\beta \cdot |\pi|)$ per example, $\beta\in\{0,0.02,0.05,0.1,0.2\}$), which exists in our implementation too but which even their own README never demonstrates using. A full sweep of this axis on top of our usual learning rate/batch size/epoch grid, across every recipe above, does not break the identity collapse.
Validation loss selects $\beta{=}0$ for all $6$ recipes, and for every $\beta$ the best configuration still predicts identity on all $250$ validation questions;
the only configurations that do not are $44$ of the $1{,}080$ whose training diverged.

\paragraph{Class-weighted cross-entropy.}
Weighting each operation by (1/frequency)$^{p}$, $p\in\{0.5, 0.6, 0.75, 0.9, 1.0\}$, each with its own best configuration of the usual grid, moves the router off identity only at high $p$:
$0.4\%$, $5.0\%$ and $62.2\%$ of test questions at $p{=}0.75$, $0.9$ and $1.0$, with pass@$1$ $48.2\%$, $45.8\%$ and $28.6\%$ against \Base's $48.2\%$.
At $p{=}1.0$ it rescues $4$ questions and loses $102$.

\paragraph{The anti-collapse \textsf{keep}-probability penalty: real diversity, always at a real accuracy cost.}
A second anti-collapse lever in \citeauthor{li2026polar}'s code, independent of the identity-program down-weight above, is an auxiliary loss term that penalizes the router's mean predicted \textsf{keep}-probability across layers.
It only activates when its weight $\lambda_{\text{keep}}$ is $>0$, and only over questions whose identity (full-depth) program is not even a valid execution.
This targets different questions than the down-weight above:
of our $1{,}116$ DM-$1$ training questions with a valid program, $45.3\%$ ($506$) have no valid identity program, while the down-weight applies to $50.4\%$ ($563$).
Their CLI default is $0.0$ (off), and their README's worked command, which supplies the $0.30$ down-weight value above, never sets it either.

A $216$-configuration sweep ($6$ values of $\lambda_{\text{keep}}\in\{0, 0.1, 0.3, 0.5, 1.0, 2.0\}$ $\times$ the usual $36$-combination hyperparameter grid)
also breaks the collapse under real, online-execution grading.
Up to $\lambda_{\text{keep}}{=}0.5$, every test question still gets identity.
At $\lambda_{\text{keep}}{=}1.0$, $3.8\%$ of test questions receive a non-identity top-$1$ prediction (pass@$1{=}47.4\%$, $-0.8$pp vs.\ \Base, pass@$5$ $59.2\%$ vs.\ $61.4\%$ without the penalty).
At $\lambda_{\text{keep}}{=}2.0$ the penalty is clearly too aggressive:
the non-identity rate reaches $36.2\%$, at $-7.4$pp (pass@$5$ $52.0\%$).
\textbf{This is a real, generation-graded break in the collapse, but always at a real accuracy cost, never a net gain}: at every $\lambda_{\text{keep}}$ tested, router pass@$1$ is at or below \Base.
Rescue@$1$ (the fraction of \Base-wrong questions the router's top-$1$ pick actually rescues) stays $0$ at $\lambda_{\text{keep}}{=}1.0$, and at $\lambda_{\text{keep}}{=}2.0$ the router rescues $6$ questions while losing $43$ that \Base\ gets right.
The router becomes willing to deviate from identity without becoming more likely to deviate \emph{correctly}. We return to why next.

\paragraph{Real per-input signal exists in the router's candidate diversity, but rank-$1$ never captures it.} Forcing identity out of the candidate pool entirely (re-decoding top-$6$, discarding identity if present, keeping the remaining $5$) turns pass@$k$ into ``best of the top-$k$ non-identity candidates.''
On DM-$1$'s $500$ test questions, pass@$1$ drops sharply this way, from $48.2\%$ to $34.6\%$ for Drop-CE (range across recipes $17.8$--$45.8\%$).
At pass@$5$ it does not drop ($62.2\%$ vs.\ $61.4\%$).
The drop at pass@$1$ is surprising, since a non-identity, \gls{mcts}-verified-correct alternative exists for essentially every question where identity itself is correct ($97.9$--$100.0\%$ across DM-$1$--$5$, matching \hyperref[finding:2]{\ftag{F2}}).
The resolution is that, on the $259$ questions identity gets wrong, the router's identity-excluded candidates rarely land inside the question's own \gls{mcts}-verified valid-program set (Drop-CE: $2.3\%$ cumulative through rank-$5$, since that recorded set reflects only a finite-budget search, not an exhaustive one), but real online-execution-graded rescue is far higher ($25.5\%$ for the same checkpoint), a $\sim\!10$--$20\times$ gap, holding at every difficulty level.
Most real rescues come from candidates \gls{mcts} never sampled but which genuinely execute correctly, so the router's candidate diversity is real and carries genuine executable signal.
What it cannot do is \emph{rank} that signal: rescue@$1$ is exactly $0$ for all six recipes at every difficulty, while rescue@$5$ is substantial.
If the router's confidence score correlated at all with correctness, some rescuable questions should land specifically at rank-$1$, but none do.
This is not a train/test artifact, since the same \gls{mcts}-membership check on a same-size slice of \emph{training} data gives $1.2$--$2.1\%$ cumulative membership, no higher than test's $2.3$--$3.5\%$, and real rescue@$1$ on that same train slice is likewise exactly $0$ for Drop-CE ($0/240$).
\textbf{Even on data it was directly trained on, Drop-CE's top-$1$ pick, when identity fails, is never the correct rescuing program.}
This is consistent with (though not conclusively proven by) positive-only supervision: training data is built exclusively from \gls{mcts}'s own recorded valid programs, with no negative signal from the programs it recorded as invalid, or from visit-count/\gls{ucb} preference.
The router's rank ordering plausibly reflects proximity to the collapsed, \textsf{keep}-heavy training marginal rather than per-candidate correctness, since nothing in its training signal penalizes a wrong candidate for scoring above a right one.

\paragraph{Selecting one recipe across models and difficulties.} \Cref{tab:winning-recipe-crossmodel} extends the treatment$\times$loss comparison to all six models across all $5$ difficulty splits, each recipe with its own best hyperparameter configuration by validation loss (\citeauthor{li2026polar}'s own criterion), evaluated with real online-execution pass@$1$--$5$.
The two loss functions are not comparable by raw validation-loss magnitude, since focal loss down-weights easy examples by construction, so it sits on a structurally lower scale than plain cross-entropy's.
Validation loss is not comparable across data treatments either:
each treatment builds its validation examples the same way as its training examples, so Drop is scored without the programs it removes.
We therefore compare all four recipes on pass@$5$, graded online on the same validation and test questions.
Pass@$5$ barely separates CE from focal: averaged over Drop and Crop, all six models and $5$ difficulties, they differ by $0.02$pp ($n{=}60$ configurations each).
Qwen3-8B's larger $6$-recipe grid shows a similarly small gap ($0.19$pp), and its six recipes, including Crop-and-Re-verify, stay within $0.5$pp averaged over the $5$ difficulties.
Crop-CE is ahead of Drop-CE in $18$ of the $30$ model$\times$difficulty combinations, tied in $4$ and behind in $8$.
We use Drop-CE, uniform across every model and difficulty split, because it reproduces \citeauthor{li2026polar}'s released code exactly (Drop) and their plain cross-entropy loss, not because it scores best:
Crop is slightly ahead on pass@$5$ (by $1.0$pp on average over all six models, mostly from Qwen3-32B), while the router's top-$1$ prediction is identity in all $120$ checkpoints, whichever recipe we use.
The gap is largest on Qwen3-32B (DM-$1$: Crop-CE $79.2\%$ vs.\ Drop-CE $74.0\%$; DM-$2$: $73.4\%$ vs.\ $65.8\%$); without it, Crop-CE leads Drop-CE by only $0.25$pp.

\begin{table}[tb]
\centering
\small
\caption{The four recipes reach pass@$5$ within $1.4$pp of each other on every model except Qwen3-32B, where Crop is $5$--$7$pp ahead of Drop.
Test pass@$5$ of each recipe's best hyperparameter configuration by validation loss, averaged over the $5$ difficulties;
search run seed $42$ only, so Drop-CE's pass@$5$ differs slightly from the mean over three runs in \cref{tab:router-results-strict-ce}.
Drop-CE is the recipe used throughout this paper.}
\label{tab:winning-recipe-crossmodel}
\begin{tabular}{lcccc}
\toprule
Model & Drop-CE & Crop-CE & Drop-focal & Crop-focal \\
\midrule
Qwen1.5-MoE-A2.7B & $31.4\%$ & $31.6\%$ & $31.5\%$ & $31.6\%$ \\
Qwen2.5-3B & $35.2\%$ & $35.1\%$ & $35.5\%$ & $35.6\%$ \\
Qwen2.5-7B & $50.0\%$ & $50.5\%$ & $49.4\%$ & $50.8\%$ \\
Qwen3-8B & $44.6\%$ & $44.5\%$ & $44.5\%$ & $44.1\%$ \\
Qwen3-32B & $51.5\%$ & $56.4\%$ & $50.6\%$ & $58.0\%$ \\
LLaMA-3.2-3B & $29.0\%$ & $29.8\%$ & $29.2\%$ & $29.2\%$ \\
\bottomrule
\end{tabular}
\end{table}

\paragraph{Menu size: consistently small across every split.}
\Cref{tab:router-topk5-menu-test} reports how many \emph{distinct} programs appear anywhere across all questions' top-$5$ decoded candidate sets, per model, DART-Math difficulty level, and data split, out of up to $500$ possible (one per question), including identity: the menu stays small, at most $26$ programs in every cell of the table,
and the same small menu appears on the training, validation, and test splits alike for the same model$\times$difficulty cell (\hyperref[finding:9]{\ftag{F9}}), so this is a property of the trained router itself, not a test-time or generalization artifact.

\begin{table}[tb]
\centering
\small
\caption{The \hyperref[finding:9]{\ftag{F9}} small-fixed-menu property holds on every split: the number of \emph{distinct} programs appearing anywhere across all questions' top-$5$ decoded candidate sets, Drop-CE recipe, per model and DART-Math difficulty level, on training ($n{=}1{,}250$/difficulty), validation ($n{=}250$/difficulty), and test ($n{=}500$/difficulty) data.
\Cref{tab:their_split_menu} gives the same count for the PoLar split.}
\label{tab:router-topk5-menu-test} 
\begin{tabular}{llccccc}
\toprule
Model & Split & DM-1 & DM-2 & DM-3 & DM-4 & DM-5 \\
\midrule
\multirow{3}{*}{Qwen1.5-MoE-A2.7B}
 & train & $14$ & $10$ & $9$ & $7$ & $17$ \\
 & val & $11$ & $6$ & $8$ & $6$ & $9$ \\
 & test & $11$ & $6$ & $8$ & $6$ & $13$ \\
\midrule
\multirow{3}{*}{Qwen2.5-3B}
 & train & $7$ & $15$ & $9$ & $12$ & $13$ \\
 & val & $8$ & $15$ & $6$ & $10$ & $12$ \\
 & test & $7$ & $15$ & $6$ & $11$ & $12$ \\
\midrule
\multirow{3}{*}{Qwen2.5-7B}
 & train & $10$ & $13$ & $14$ & $13$ & $11$ \\
 & val & $9$ & $10$ & $9$ & $10$ & $10$ \\
 & test & $9$ & $10$ & $11$ & $12$ & $12$ \\
\midrule
\multirow{3}{*}{Qwen3-8B}
 & train & $8$ & $13$ & $9$ & $10$ & $10$ \\
 & val & $8$ & $13$ & $8$ & $11$ & $8$ \\
 & test & $8$ & $12$ & $8$ & $11$ & $8$ \\
\midrule
\multirow{3}{*}{Qwen3-32B}
 & train & $22$ & $16$ & $26$ & $23$ & $15$ \\
 & val & $19$ & $16$ & $21$ & $18$ & $14$ \\
 & test & $21$ & $17$ & $24$ & $21$ & $15$ \\
\midrule
\multirow{3}{*}{LLaMA-3.2-3B}
 & train & $11$ & $9$ & $10$ & $11$ & $13$ \\
 & val & $9$ & $7$ & $9$ & $10$ & $11$ \\
 & test & $10$ & $8$ & $9$ & $11$ & $12$ \\
\bottomrule
\end{tabular}
\end{table}

\section{Running \textsc{PoLar}'s Setup on the PoLar Split}
\label{sec:appendix_their_split}
\glsresetall

The main results use our DART-Math split, the \texttt{oneshot} prompt, and our search (\cref{sec:appendix_dart_math,sec:appendix_mcts_design,sec:appendix_placeholder_echo}). 
Here we run \citeauthor{li2026polar}'s released search\footnote{\url{https://github.com/tianyi-lab/PoLar}} on our reconstruction of their data split, which we call the \emph{PoLar split}, to test their router under their own setup, with their prompt and, for comparison, ours.

\subsection{Setup}
\label{sec:appendix_their_split_setup}

\textbf{Split.}
Their questions come from MATH \citep{hendrycks2021math} only.
DM-$N$ is the difficulty level (1 to 5) that the MATH dataset assigns to each problem, with $565 / 1{,}349 / 1{,}579 / 1{,}537 / 1{,}577$ questions per level after deduplication.
In contrast, our split mixes MATH and GSM8K questions (\cref{sec:appendix_dart_math}).
Their code defines these question pools.
Their sizes differ from the $2{,}000$ questions per level of their first version and from our split (\cref{sec:appendix_dart_math}),
but the share of train, validation and test is the same in all three: about $62.5 / 12.5 / 25\%$.
We cut the pools at the reported sizes, in their sample order, with contiguous train, validation and test parts; the test sets have $141 / 337 / 395 / 384 / 394$ questions ($1{,}651$ in total).
With these test sets, LLaMA-3.2-3B's \Base, evaluated with their code, is within $4.1$pp of their reported numbers, summed over the five levels ($6.2$ for our own run in \cref{tab:their_split_base}).
To see how likely such a match is by chance, we drew $2{,}000$ random test sets of the same sizes from the same pools: none comes this close (median gap $27$pp).
We therefore take these test sets to be theirs.

\textbf{Models and evaluation.}
We run the four models for which \citeauthor{li2026polar} publish a table: LLaMA-3.2-3B, Qwen1.5-MoE-A2.7B, Qwen2.5-3B and Qwen3-8B.
We use \texttt{polar\_default}, the prompt of their code, and, for comparison, \texttt{oneshot} (\cref{sec:appendix_placeholder_echo}), with a generation cap of $50$ tokens, $\tau{=}0$, and no chat template (see \cref{sec:appendix_their_split_base}).
\citeauthor{li2026polar} evaluate on one test set per level, which is small ($141$ questions on DM-1, $\pm4$pp).
We therefore pool the router and fixed-program results (\cref{tab:their_split_results}) over 4 folds, so every question of a level is a test question once,
and use their exact test sets only to compare \Base\ with their published numbers (\cref{tab:their_split_base}).

\subsection{Search, Router, and Fixed Program}
\label{sec:appendix_their_split_search}

\begin{figure}[tb]
\centering
\begin{subfigure}{0.4798\linewidth}\centering
\includegraphics[width=1.0000\linewidth]{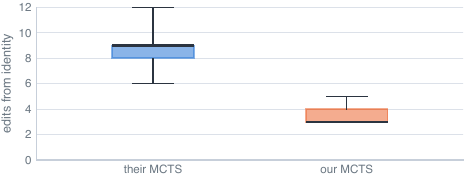}
\caption{Deepest program per question.}\end{subfigure}%
\hspace{15.0pt}%
\begin{subfigure}{0.4798\linewidth}\centering
\includegraphics[width=1.0000\linewidth]{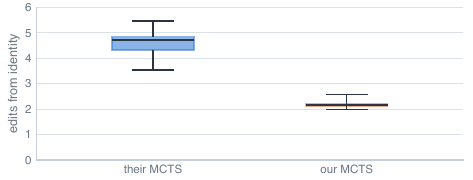}
\caption{Mean over its programs.}\end{subfigure}
\caption{Their search reaches much deeper programs than ours (median deepest program: $9$ vs.\ $3$ edits).
Edit depth (skip and repeat edits from the identity), seed $42$, pooled over models and levels. Their MCTS: PoLar search and split, \texttt{polar\_default} (\cref{sec:appendix_their_split_search}); our MCTS: widening search and DART-Math split, \texttt{oneshot} (\cref{sec:reproduction_mcts_search}). With \texttt{oneshot}, their search gives the same median ($9$ edits).
The search, not the split, sets the depth: on our split, their search still reaches a median of $8$ (Qwen3-8B and LLaMA-3.2-3B, train and validation questions), and ours reaches $3$ on theirs (Qwen3-8B). Each question gives its deepest program (a) and the mean over its programs (b). Box: quartiles; line: median; whiskers: minimum and maximum. Interactive trees of the searches: \ifdeanonymized\projectpage\else[\emph{removed for blind review}]\fi.}
\label{fig:their_split_depth}
\end{figure}

\textbf{Search.}
\Cref{tab:their_split_search} lists how their \gls{mcts} differs from ours.
On its way down the tree, it follows the child with the best \gls{ucb} score, but at each level, with probability $0.1$, it picks a random child instead.
It creates all children of a node but evaluates only one of them, picked at random, and an unevaluated child has no score, so only these random picks lead into new children.
As a result, it spends almost its whole budget below a single first edit, and it reaches much deeper programs than our implementation of the \gls{mcts} (\cref{fig:their_split_depth}): over all models and levels, the deepest program per question has a median of $9$ edits (at most $12$), against $3$ for our search on our DART-Math split.
On Qwen3-8B DM-1, it evaluates one distinct single-edit program per question (a randomly picked root child is expanded, not scored), against about $29$ for ours.
Their code sets \texttt{random.seed(42)} at the start of every question's search, so the random choices of all questions start out the same;
with one seed per question, the most frequent valid program covers $0.9\%$ of questions instead of $60.7\%$, while accuracy, depth and segment lengths barely change (Qwen3-8B, \texttt{oneshot}, DM-1).
We use our port of their search, which we checked program for program against their code and which runs about $5\times$ faster (\cref{sec:appendix_masked_batch_kv}), with all their parameters (\cref{tab:their_split_search}).

\textbf{Runs.}
We run each search three times (seeds $42$, $1$ and $7$, called runs 1 to 3 below), and run 1 uses seed $42$ as their code does.

\begin{table}[tb]
\centering
\small
\caption{Their search evaluates one random child per expansion and starts every question from the same seed; ours widens one evaluated child at a time and seeds each question separately.
Their released \gls{mcts} (\citeauthor{li2026polar}'s code, run here through our port) against our search (\cref{sec:reproduction_mcts_search} and~\labelcref{sec:appendix_mcts_design}).}
\label{tab:their_split_search}
\begin{tabular}{p{0.24\linewidth}p{0.36\linewidth}p{0.3\linewidth}}
\toprule
 & Their search & Our search \\
\midrule
Expansion & all children at once, one evaluated, at random & one child at a time, each evaluated \\
Unevaluated children & never entered by the \gls{ucb} descent & none \\
Breadth & random pick, probability $0.1$ at each level & progressive widening \\
Random seed & \texttt{random.seed(42)} for every question & one seed per question \\
Multi-layer moves & at the root only & at every node \\
Nested edits & allowed & not allowed \\
Simulations per question & $100$ & $200$ \\
Executions of a layer & at most $2$ & at most $5$ \\
Program length & $24$ to $44$ layers (Qwen3-8B; per model) & not bounded \\
Grading timeout & $60$\,s & $5$\,s \\
\bottomrule
\end{tabular}
\end{table}

\textbf{Router and fixed program.}
\Router\ is their router, trained as in their code (Drop-CE, \cref{sec:appendix_router_recipe}), keeping the checkpoint with the lowest validation loss.
The fixed program is not part of \citeauthor{li2026polar}'s method.
We add it to separate what the router adds from what a single good program gives.
For each fold, it is chosen from the $20$ non-identity programs with the highest success rate on the training questions' search data (smoothed for rarely tried programs): the one with the best accuracy on the validation questions.
A program that the search evaluated on $n$ training questions and found valid on $s$ of them gets the rate $(s+10\mu)/(n+10)$, where $\mu$ is the success rate of all non-identity programs in the training data; only programs valid on at least $3$ training questions are candidates.
It is then applied unchanged to every test question (as in the main track, \cref{sec:study_setup}).

\subsection{\Base\ on the PoLar Test Set}
\label{sec:appendix_their_split_base}

\Cref{tab:their_split_base} gives \Base($\tau{=}0$) on the PoLar test set, with and without the chat template that their evaluation code applies (\cref{sec:appendix_chat_template}).
Without the template and with \texttt{polar\_default}, LLaMA-3.2-3B is close to the published numbers, Qwen1.5-MoE-A2.7B and Qwen3-8B on some levels, and Qwen2.5-3B on none.
The chat template lowers accuracy in most cells; for Qwen3-8B with \texttt{polar\_default}, it falls from $39.7$ to $9.9$ on DM-1.
As in the rest of the paper, we therefore use no chat template unless stated otherwise.

\begin{table}[tb]
\centering
\small
\caption{Without the chat template, \Base\ is close to the published numbers for LLaMA-3.2-3B and, on some levels, for Qwen1.5-MoE-A2.7B and Qwen3-8B;
with their protocol ($\dagger$), all three Qwen models fall below them in every cell.
\Base($\tau{=}0$) accuracy (\%) on the PoLar test set ($141 / 337 / 395 / 384 / 394$ questions), generation cap of $50$ tokens. \citeauthor{li2026polar} (reported): the numbers in their paper. Underlined: our run closest to the reported number, per model and level. $\dagger$: their protocol, \texttt{polar\_default} with the chat template their evaluation code applies (none for LLaMA-3.2-3B); these are our runs and would match the reported numbers if our setup were identical. All other rows are our runs with other prompts or without the template.}
\label{tab:their_split_base}
\begin{tabular}{ll *{5}{r}}
\toprule
Model & Setting & DM-1 & DM-2 & DM-3 & DM-4 & DM-5 \\
\midrule
\multirow{5}{*}{Qwen1.5-MoE-A2.7B}
 & \citeauthor{li2026polar} (reported) & 48.2 & 25.5 & 24.1 & 16.7 & 10.9 \\
 & \texttt{polar\_default} & 39.7 & \underline{24.9} & \underline{22.5} & \underline{16.4} & 13.7 \\
 & \texttt{polar\_default}, chat template$^\dagger$ & 41.1 & 19.9 & 14.7 & 13.3 & \underline{9.1} \\
 & \texttt{oneshot} & \underline{50.3} & 35.0 & 27.6 & 19.8 & 17.8 \\
 & \texttt{oneshot}, chat template & 41.1 & 24.0 & 18.7 & 11.7 & 7.4 \\
\midrule
\multirow{5}{*}{Qwen2.5-3B}
 & \citeauthor{li2026polar} (reported) & 17.7 & 7.1 & 3.8 & 3.4 & 1.5 \\
 & \texttt{polar\_default} & 0.0 & 1.2 & 0.5 & 0.3 & \underline{0.5} \\
 & \texttt{polar\_default}, chat template$^\dagger$ & \underline{1.4} & \underline{1.8} & \underline{0.8} & \underline{0.5} & 0.2 \\
 & \texttt{oneshot} & 58.2 & 32.3 & 24.3 & 19.0 & 15.2 \\
 & \texttt{oneshot}, chat template & 50.3 & 20.2 & 17.5 & 13.8 & 8.4 \\
\midrule
\multirow{5}{*}{Qwen3-8B}
 & \citeauthor{li2026polar} (reported) & 57.4 & 30.6 & 17.5 & 16.7 & 11.7 \\
 & \texttt{polar\_default} & 39.7 & \underline{29.7} & \underline{22.5} & \underline{14.8} & \underline{13.7} \\
 & \texttt{polar\_default}, chat template$^\dagger$ & 9.9 & 2.1 & 0.5 & 0.8 & 0.0 \\
 & \texttt{oneshot} & 73.8 & 45.7 & 32.9 & 28.1 & 21.8 \\
 & \texttt{oneshot}, chat template & \underline{71.6} & 46.9 & 34.4 & 25.8 & 17.3 \\
\midrule
\multirow{3}{*}{LLaMA-3.2-3B}
 & \citeauthor{li2026polar} (reported) & 51.1 & 30.6 & 29.9 & 27.1 & 26.1 \\
 & \texttt{polar\_default}$^\dagger$ & 48.9 & \underline{29.4} & \underline{29.4} & \underline{27.6} & \underline{27.9} \\
 & \texttt{oneshot} & \underline{51.8} & 25.8 & 29.1 & 26.0 & 24.1 \\
\bottomrule
\end{tabular}
\end{table}

\subsection{Findings 1--4 under Their Setup}
\label{sec:appendix_their_split_findings}

With their search on the PoLar split, we repeat the measurements behind \hyperref[finding:1]{\ftag{F1}}--\hyperref[finding:4]{\ftag{F4}}, with the same definitions as in \cref{sec:analysis_polar_recap}, on the four models and with both prompts, over the three search seeds (\cref{tab:their_split_findings}).
\hyperref[finding:1]{\ftag{F1}} holds:
\textsf{skip} and \textsf{repeat} combined beat either alone on every model, prompt and level.
\textsf{repeat} is the stronger mechanism on its own on the mean over search seeds, but a single search seed can reverse this (seed $7$, on all three Qwen models).
For \hyperref[finding:2]{\ftag{F2}}, most solved questions have a shorter valid program on Qwen2.5-3B, Qwen3-8B and LLaMA-3.2-3B with either setup, and on Qwen1.5-MoE-A2.7B with neither.
For \hyperref[finding:3]{\ftag{F3}}, the share of solvable questions that need recurrence rises with the level on Qwen3-8B and Qwen1.5-MoE-A2.7B, and on Qwen2.5-3B with \texttt{oneshot}.
With \texttt{polar\_default}, Qwen2.5-3B's standard forward pass fails on almost every question (\cref{tab:their_split_base}), so nearly every solvable question needs an edit at every level.
On LLaMA-3.2-3B, the share stays about flat, unlike on our split, where it falls because our levels mix MATH and GSM8K questions (\cref{sec:appendix_llama_levels}).
Finally, \hyperref[finding:4]{\ftag{F4}} holds with their search:
$70$--$81\%$ of segments touch a single layer and $80$--$93\%$ at most two, above both thresholds of \citet{li2026polar} on every model and with both prompts,
whereas with our search on our split, 5 of our 6 models fall below both (\cref{tab:analysis_f4}).
This likely follows from the searches: theirs makes multi-layer moves only at the root, ours at every node (\cref{tab:their_split_search}), so their segments are mostly built from single-layer moves.

\begin{table}[tb]
\centering
\footnotesize
\caption{With their search, segments are short enough for \hyperref[finding:4]{\ftag{F4}} on every model and prompt, while with ours they mostly are not; \hyperref[finding:2]{\ftag{F2}} and \hyperref[finding:3]{\ftag{F3}} vary less systematically with the search.
\hyperref[finding:2]{\ftag{F2}}--\hyperref[finding:4]{\ftag{F4}} with their search on the PoLar split (Theirs) and with our search on our split (Ours, the main track), over all searched questions ($6{,}607$ and $10{,}000$), mean{\scriptsize$\pm$}std over the three search seeds.
\hyperref[finding:2]{\ftag{F2}}: share of solved questions whose shortest valid program is shorter than the standard forward pass, pooled over levels.
\hyperref[finding:3]{\ftag{F3}}: change in the share of solvable questions that need recurrence from DM-1 to DM-5, in pp.
\hyperref[finding:4]{\ftag{F4}}: share of segments in the shortest valid program per solved question that touch a single layer, and at most two layers, pooled over levels (\citeauthor{li2026polar} report $57.7\%$ and over two-thirds; our-split rows as in \cref{tab:analysis_f4}).
\textbf{Bold}: at or above \citeauthor{li2026polar}'s threshold.}
\label{tab:their_split_findings}
\begin{tabular}{llcccc}
\toprule
& & \hyperref[finding:2]{\ftag{F2}} & \hyperref[finding:3]{\ftag{F3}} & \multicolumn{2}{c}{\hyperref[finding:4]{\ftag{F4}} segments (\%)} \\
\cmidrule(lr){5-6}
Model & Setup & shorter (\%) & recurrence, DM-1$\to$5 & single-layer & $\le 2$ layers \\
\midrule
\multirow{3}{*}{Qwen1.5-MoE-A2.7B} & Theirs, \texttt{polar\_default} & $35.0${\scriptsize$\pm$13.5} & $+14.2${\scriptsize$\pm$3.0} & $\mathbf{77.2}${\scriptsize$\pm$12.9} & $\mathbf{93.0}${\scriptsize$\pm$4.8} \\
 & Theirs, \texttt{oneshot} & $38.0${\scriptsize$\pm$9.4} & $+15.7${\scriptsize$\pm$2.4} & $\mathbf{77.7}${\scriptsize$\pm$9.8} & $\mathbf{91.1}${\scriptsize$\pm$3.1} \\
 & Ours, \texttt{oneshot} & $43.3${\scriptsize$\pm$0.6} & $+2.8${\scriptsize$\pm$0.2} & $45.5${\scriptsize$\pm$0.8} & $\mathbf{73.5}${\scriptsize$\pm$0.2} \\
\midrule
\multirow{3}{*}{Qwen2.5-3B} & Theirs, \texttt{polar\_default} & $70.6${\scriptsize$\pm$13.2} & $-1.5${\scriptsize$\pm$0.6} & $\mathbf{80.5}${\scriptsize$\pm$5.4} & $\mathbf{90.1}${\scriptsize$\pm$4.5} \\
 & Theirs, \texttt{oneshot} & $75.4${\scriptsize$\pm$9.3} & $+35.4${\scriptsize$\pm$0.7} & $\mathbf{78.1}${\scriptsize$\pm$4.6} & $\mathbf{87.3}${\scriptsize$\pm$4.4} \\
 & Ours, \texttt{oneshot} & $65.9${\scriptsize$\pm$0.8} & $+18.6${\scriptsize$\pm$0.7} & $32.1${\scriptsize$\pm$0.7} & $58.1${\scriptsize$\pm$0.3} \\
\midrule
\multirow{3}{*}{Qwen3-8B} & Theirs, \texttt{polar\_default} & $78.8${\scriptsize$\pm$10.5} & $+24.7${\scriptsize$\pm$2.0} & $\mathbf{79.5}${\scriptsize$\pm$3.1} & $\mathbf{89.7}${\scriptsize$\pm$2.3} \\
 & Theirs, \texttt{oneshot} & $82.2${\scriptsize$\pm$6.9} & $+33.4${\scriptsize$\pm$1.4} & $\mathbf{77.7}${\scriptsize$\pm$4.8} & $\mathbf{89.2}${\scriptsize$\pm$3.0} \\
 & Ours, \texttt{oneshot} & $70.7${\scriptsize$\pm$0.3} & $+29.7${\scriptsize$\pm$0.3} & $32.6${\scriptsize$\pm$0.3} & $58.5${\scriptsize$\pm$0.1} \\
\midrule
\multirow{3}{*}{LLaMA-3.2-3B} & Theirs, \texttt{polar\_default} & $56.1${\scriptsize$\pm$3.8} & $-1.8${\scriptsize$\pm$2.6} & $\mathbf{70.5}${\scriptsize$\pm$0.6} & $\mathbf{81.8}${\scriptsize$\pm$3.3} \\
 & Theirs, \texttt{oneshot} & $60.0${\scriptsize$\pm$3.7} & $+4.8${\scriptsize$\pm$1.2} & $\mathbf{70.0}${\scriptsize$\pm$1.2} & $\mathbf{80.5}${\scriptsize$\pm$2.9} \\
 & Ours, \texttt{oneshot} & $54.4${\scriptsize$\pm$0.7} & $-15.4${\scriptsize$\pm$0.7} & $36.5${\scriptsize$\pm$0.4} & $64.2${\scriptsize$\pm$0.3} \\
\bottomrule
\end{tabular}
\end{table}

\subsection{\Router\ and Fixed Program under Their Setup}
\label{sec:appendix_their_split_results}

\Cref{tab:their_split_results} gives the change in accuracy of the router and of the fixed program for each search run.

\begin{table}[tb]
\centering
\scriptsize
\caption{With \texttt{polar\_default}, the router gains only on Qwen2.5-3B and Qwen3-8B, where the format breaks, and wherever it gains, the fixed program gains more, in every run and level;
with \texttt{oneshot}, neither gains on any model.
Results (\%) on the PoLar split, pooled over 4 folds, for the three search runs. $\Delta$ is the accuracy of the \Router's top-$1$ program (p@1) or of the fixed program, minus \Base($\tau{=}0$) on the same questions, in pp. Cells are colored from red (negative) to green (positive); the color saturates at $\pm15$.}
\label{tab:their_split_results}
\setlength{\tabcolsep}{3pt}
\begin{tabular}{lll r rrr rrr}
\toprule
 & & & & \multicolumn{3}{c}{\Router\ $\Delta$, run} & \multicolumn{3}{c}{Fixed program $\Delta$, run} \\
\cmidrule(lr){5-7}\cmidrule(lr){8-10}
Model & Prompt & & \Base($\tau{=}0$) & $1$ & $2$ & $3$ & $1$ & $2$ & $3$ \\
\midrule
\multirow{10}{*}{Qwen1.5-MoE-A2.7B} & \multirow{5}{*}{\texttt{polar\_default}} & DM-1 & 38.2 & \cellcolor{red!22!white}$-$5.5 & \cellcolor{red!60!white}$-$18.9 & \cellcolor{red!57!white}$-$14.2 & \cellcolor{red!6!white}$-$1.6 & \cellcolor{green!2!white}+0.4 & \cellcolor{red!6!white}$-$1.4 \\
 &  & DM-2 & 28.1 & \cellcolor{red!21!white}$-$5.2 & \cellcolor{red!51!white}$-$12.8 & \cellcolor{red!22!white}$-$5.6 & \cellcolor{red!19!white}$-$4.8 & \cellcolor{red!17!white}$-$4.2 & \cellcolor{red!4!white}$-$1.0 \\
 &  & DM-3 & 21.0 & \cellcolor{red!25!white}$-$6.3 & \cellcolor{red!50!white}$-$12.4 & \cellcolor{red!26!white}$-$6.4 & \cellcolor{red!16!white}$-$3.9 & \cellcolor{red!18!white}$-$4.4 & \cellcolor{red!6!white}$-$1.6 \\
 &  & DM-4 & 15.2 & \cellcolor{red!16!white}$-$4.1 & \cellcolor{red!28!white}$-$7.0 & \cellcolor{red!18!white}$-$4.4 & \cellcolor{red!16!white}$-$4.0 & \cellcolor{red!11!white}$-$2.7 & \cellcolor{green!2!white}+0.4 \\
 &  & DM-5 & 9.6 & \cellcolor{red!12!white}$-$3.0 & \cellcolor{red!14!white}$-$3.4 & \cellcolor{red!8!white}$-$2.0 & \cellcolor{red!9!white}$-$2.2 & \cellcolor{red!7!white}$-$1.8 & \cellcolor{red!2!white}$-$0.6 \\
\cmidrule{2-10}
 & \multirow{5}{*}{\texttt{oneshot}} & DM-1 & 43.5 & \cellcolor{red!12!white}$-$3.0 & \cellcolor{red!60!white}$-$31.9 & \cellcolor{red!38!white}$-$9.4 & \cellcolor{red!26!white}$-$6.5 & \cellcolor{red!13!white}$-$3.2 & \cellcolor{red!20!white}$-$5.1 \\
 &  & DM-2 & 33.2 & \cellcolor{red!12!white}$-$3.0 & \cellcolor{red!60!white}$-$18.1 & \cellcolor{red!26!white}$-$6.4 & \cellcolor{red!32!white}$-$7.9 & \cellcolor{red!19!white}$-$4.8 & \cellcolor{red!12!white}$-$3.0 \\
 &  & DM-3 & 25.6 & \cellcolor{red!13!white}$-$3.2 & \cellcolor{red!60!white}$-$16.3 & \cellcolor{red!10!white}$-$2.5 & \cellcolor{red!27!white}$-$6.8 & \cellcolor{red!25!white}$-$6.3 & \cellcolor{red!13!white}$-$3.2 \\
 &  & DM-4 & 17.4 & \cellcolor{red!17!white}$-$4.2 & \cellcolor{red!46!white}$-$11.6 & \cellcolor{red!4!white}$-$1.0 & \cellcolor{red!10!white}$-$2.5 & \cellcolor{red!11!white}$-$2.8 & \cellcolor{red!7!white}$-$1.8 \\
 &  & DM-5 & 13.0 & \cellcolor{red!9!white}$-$2.2 & \cellcolor{red!38!white}$-$9.6 & \cellcolor{red!18!white}$-$4.6 & \cellcolor{red!17!white}$-$4.2 & \cellcolor{red!15!white}$-$3.8 & \cellcolor{red!14!white}$-$3.6 \\
\midrule
\multirow{10}{*}{Qwen2.5-3B} & \multirow{5}{*}{\texttt{polar\_default}} & DM-1 & 0.9 & \cellcolor{green!26!white}+6.4 & \cellcolor{green!30!white}+7.4 & \cellcolor{green!4!white}+1.1 & \cellcolor{green!60!white}+34.3 & \cellcolor{green!60!white}+43.0 & \cellcolor{green!60!white}+38.4 \\
 &  & DM-2 & 0.7 & \cellcolor{green!29!white}+7.2 & \cellcolor{green!54!white}+13.5 & \cellcolor{green!3!white}+0.7 & \cellcolor{green!60!white}+23.1 & \cellcolor{green!60!white}+29.9 & \cellcolor{green!60!white}+27.6 \\
 &  & DM-3 & 0.2 & \cellcolor{green!31!white}+7.7 & \cellcolor{green!23!white}+5.8 & \cellcolor{green!2!white}+0.6 & \cellcolor{green!60!white}+15.3 & \cellcolor{green!60!white}+20.5 & \cellcolor{green!60!white}+19.9 \\
 &  & DM-4 & 0.1 & \cellcolor{green!18!white}+4.5 & \cellcolor{green!10!white}+2.6 & \cellcolor{green!2!white}+0.5 & \cellcolor{green!41!white}+10.3 & \cellcolor{green!58!white}+14.6 & \cellcolor{green!36!white}+9.1 \\
 &  & DM-5 & 0.3 & \cellcolor{green!6!white}+1.5 & \cellcolor{green!8!white}+2.1 & \cellcolor{green!4!white}+0.9 & \cellcolor{green!23!white}+5.7 & \cellcolor{green!20!white}+5.1 & \cellcolor{green!26!white}+6.6 \\
\cmidrule{2-10}
 & \multirow{5}{*}{\texttt{oneshot}} & DM-1 & 54.2 & \cellcolor{red!8!white}$-$2.1 & \cellcolor{red!14!white}$-$3.4 & \cellcolor{red!16!white}$-$3.9 & \cellcolor{red!15!white}$-$3.7 & \cellcolor{red!9!white}$-$2.3 & \cellcolor{red!14!white}$-$3.5 \\
 &  & DM-2 & 36.9 & \cellcolor{red!15!white}$-$3.7 & \cellcolor{red!15!white}$-$3.7 & \cellcolor{red!13!white}$-$3.2 & \cellcolor{red!13!white}$-$3.2 & \cellcolor{red!18!white}$-$4.4 & \cellcolor{red!14!white}$-$3.4 \\
 &  & DM-3 & 25.6 & \cellcolor{red!8!white}$-$2.0 & \cellcolor{red!2!white}$-$0.6 & \cellcolor{red!12!white}$-$3.0 & \cellcolor{red!6!white}$-$1.6 & \cellcolor{red!14!white}$-$3.5 & \cellcolor{red!16!white}$-$3.9 \\
 &  & DM-4 & 16.5 & \cellcolor{red!4!white}$-$1.1 & \cellcolor{red!7!white}$-$1.8 & \cellcolor{red!0!white}$-$0.1 & \cellcolor{red!6!white}$-$1.4 & \cellcolor{red!13!white}$-$3.3 & \cellcolor{red!16!white}$-$4.0 \\
 &  & DM-5 & 9.8 & \cellcolor{red!3!white}$-$0.8 & \cellcolor{red!4!white}$-$0.9 & \cellcolor{red!6!white}$-$1.5 & \cellcolor{red!10!white}$-$2.4 & \cellcolor{red!8!white}$-$1.9 & \cellcolor{red!7!white}$-$1.7 \\
\midrule
\multirow{10}{*}{Qwen3-8B} & \multirow{5}{*}{\texttt{polar\_default}} & DM-1 & 42.8 & \cellcolor{green!6!white}+1.4 & \cellcolor{green!6!white}+1.6 & \cellcolor{red!28!white}$-$6.9 & \cellcolor{green!38!white}+9.4 & \cellcolor{green!27!white}+6.7 & \cellcolor{green!35!white}+8.7 \\
 &  & DM-2 & 31.1 & \cellcolor{green!5!white}+1.3 & \cellcolor{red!2!white}$-$0.4 & \cellcolor{red!24!white}$-$5.9 & \cellcolor{green!27!white}+6.7 & \cellcolor{green!27!white}+6.8 & \cellcolor{green!26!white}+6.6 \\
 &  & DM-3 & 21.2 & \cellcolor{green!2!white}+0.4 & \cellcolor{red!4!white}$-$1.0 & \cellcolor{red!19!white}$-$4.8 & \cellcolor{green!22!white}+5.5 & \cellcolor{green!15!white}+3.7 & \cellcolor{green!34!white}+8.5 \\
 &  & DM-4 & 11.7 & \cellcolor{green!8!white}+1.9 & \cellcolor{green!6!white}+1.6 & \cellcolor{red!8!white}$-$2.0 & \cellcolor{green!26!white}+6.4 & \cellcolor{green!24!white}+6.1 & \cellcolor{green!30!white}+7.6 \\
 &  & DM-5 & 8.6 & \cellcolor{red!1!white}$-$0.3 & \cellcolor{red!5!white}$-$1.2 & \cellcolor{red!8!white}$-$1.9 & \cellcolor{green!9!white}+2.2 & \cellcolor{green!8!white}+2.0 & \cellcolor{green!0!white}+0.1 \\
\cmidrule{2-10}
 & \multirow{5}{*}{\texttt{oneshot}} & DM-1 & 65.8 & \cellcolor{red!4!white}$-$1.1 & \cellcolor{red!8!white}$-$1.9 & \cellcolor{red!60!white}$-$15.9 & \cellcolor{red!18!white}$-$4.4 & \cellcolor{red!18!white}$-$4.6 & \cellcolor{red!6!white}$-$1.6 \\
 &  & DM-2 & 50.7 & \cellcolor{red!10!white}$-$2.5 & \cellcolor{red!12!white}$-$3.0 & \cellcolor{red!48!white}$-$12.1 & \cellcolor{red!16!white}$-$3.9 & \cellcolor{red!16!white}$-$4.1 & \cellcolor{red!23!white}$-$5.7 \\
 &  & DM-3 & 36.2 & \cellcolor{red!17!white}$-$4.3 & \cellcolor{red!12!white}$-$2.9 & \cellcolor{red!25!white}$-$6.2 & \cellcolor{red!2!white}$-$0.6 & \cellcolor{red!20!white}$-$4.9 & \cellcolor{red!9!white}$-$2.2 \\
 &  & DM-4 & 25.8 & \cellcolor{red!14!white}$-$3.4 & \cellcolor{red!3!white}$-$0.7 & \cellcolor{red!28!white}$-$6.9 & \cellcolor{red!5!white}$-$1.2 & \cellcolor{red!18!white}$-$4.6 & \cellcolor{red!12!white}$-$2.9 \\
 &  & DM-5 & 15.2 & \cellcolor{red!2!white}$-$0.5 & \cellcolor{red!5!white}$-$1.2 & \cellcolor{red!4!white}$-$1.0 & \cellcolor{red!7!white}$-$1.8 & \cellcolor{red!12!white}$-$3.0 & \cellcolor{red!9!white}$-$2.3 \\
\midrule
\multirow{10}{*}{LLaMA-3.2-3B} & \multirow{5}{*}{\texttt{polar\_default}} & DM-1 & 38.8 & \cellcolor{red!2!white}$-$0.5 & \cellcolor{red!11!white}$-$2.7 & \cellcolor{red!4!white}$-$1.1 & \cellcolor{red!26!white}$-$6.5 & \cellcolor{red!37!white}$-$9.2 & \cellcolor{red!27!white}$-$6.7 \\
 &  & DM-2 & 29.8 & \cellcolor{red!1!white}$-$0.3 & \cellcolor{red!19!white}$-$4.7 & \cellcolor{red!2!white}$-$0.4 & \cellcolor{red!13!white}$-$3.3 & \cellcolor{red!24!white}$-$5.9 & \cellcolor{red!22!white}$-$5.4 \\
 &  & DM-3 & 24.0 & \cellcolor{red!0!white}$-$0.1 & \cellcolor{red!5!white}$-$1.3 & \cellcolor{red!2!white}$-$0.4 & \cellcolor{red!29!white}$-$7.2 & \cellcolor{red!22!white}$-$5.6 & \cellcolor{red!27!white}$-$6.8 \\
 &  & DM-4 & 23.3 & +0.0 & \cellcolor{green!0!white}+0.1 & \cellcolor{red!1!white}$-$0.2 & \cellcolor{red!42!white}$-$10.5 & \cellcolor{red!17!white}$-$4.2 & \cellcolor{red!30!white}$-$7.4 \\
 &  & DM-5 & 26.1 & +0.0 & \cellcolor{red!8!white}$-$2.0 & \cellcolor{red!1!white}$-$0.2 & \cellcolor{red!49!white}$-$12.3 & \cellcolor{red!16!white}$-$3.9 & \cellcolor{red!34!white}$-$8.4 \\
\cmidrule{2-10}
 & \multirow{5}{*}{\texttt{oneshot}} & DM-1 & 41.6 & \cellcolor{red!4!white}$-$0.9 & \cellcolor{red!18!white}$-$4.6 & \cellcolor{red!4!white}$-$0.9 & \cellcolor{red!35!white}$-$8.8 & \cellcolor{red!23!white}$-$5.7 & \cellcolor{red!40!white}$-$10.1 \\
 &  & DM-2 & 29.6 & \cellcolor{red!1!white}$-$0.2 & \cellcolor{red!5!white}$-$1.3 & \cellcolor{red!2!white}$-$0.4 & \cellcolor{red!27!white}$-$6.8 & \cellcolor{red!5!white}$-$1.3 & \cellcolor{red!35!white}$-$8.8 \\
 &  & DM-3 & 25.5 & \cellcolor{green!0!white}+0.1 & \cellcolor{red!6!white}$-$1.6 & \cellcolor{green!0!white}+0.1 & \cellcolor{red!34!white}$-$8.5 & \cellcolor{red!24!white}$-$5.9 & \cellcolor{red!41!white}$-$10.3 \\
 &  & DM-4 & 23.8 & \cellcolor{red!1!white}$-$0.3 & \cellcolor{red!4!white}$-$1.0 & \cellcolor{red!0!white}$-$0.1 & \cellcolor{red!31!white}$-$7.8 & \cellcolor{red!26!white}$-$6.4 & \cellcolor{red!36!white}$-$9.1 \\
 &  & DM-5 & 24.9 & +0.0 & \cellcolor{red!8!white}$-$1.9 & \cellcolor{red!0!white}$-$0.1 & \cellcolor{red!39!white}$-$9.8 & \cellcolor{red!19!white}$-$4.7 & \cellcolor{red!18!white}$-$4.4 \\
\bottomrule
\end{tabular}
\end{table}

\textbf{Their prompt.}
With \texttt{polar\_default}, Qwen3-8B often answers with the placeholder \texttt{\textbackslash boxed\{ANSWER\}} instead of an answer, and Qwen2.5-3B usually reasons until the $50$ tokens run out, without writing a box.
The fixed program fixes both, and gains on every run and level (\cref{tab:their_split_results}).
It differs between runs and levels; for example, it repeats layers $15$ and $34$ once for Qwen3-8B, run 1, DM-3 to DM-5.
The router gains less than the fixed program, and on Qwen3-8B DM-2 to DM-5 it does not reach the gains that \citeauthor{li2026polar} report (DM-2: at most $+1.3$ against their $+12.1$).
All of these gains come from the questions where identity's answer is unusable, the placeholder for Qwen3-8B and no box for Qwen2.5-3B:
on the questions where identity gives a real answer, and on LLaMA-3.2-3B and Qwen1.5-MoE-A2.7B, which have no such failure, neither the fixed program nor the router gains, on any run or level.
So the gain is a repair of the output format, not better reasoning.

\textbf{Our prompt.}
With \texttt{oneshot}, Qwen3-8B no longer answers with the placeholder, and Qwen2.5-3B's \Base\ on DM-1 rises from $0.9$ to $54.2\%$.
With the format already correct, there is nothing left for the router or the fixed program to repair, and no model gains.
The one-shot example is thus a cheaper fix of the same failure than a fixed program.

\textbf{Search, not split or prompt.}
Whether the router collapses depends on the search.
With their search on our split, the Qwen3-8B router does not collapse either, with either prompt, and with \texttt{polar\_default} it again gains only by repairing the placeholder answers;
with our search on the questions of the PoLar split, it collapses to identity, as on our split.
With their search, only the LLaMA-3.2-3B router collapses, to identity, while the other routers mostly switch between identity and one dominant edit (\cref{tab:their_split_menu}).
They also do not propose a small fixed menu: one router's top-$5$ sets hold on average $26$ to $101$ distinct programs on its own fold, against $6$ to $24$ on our split (\cref{tab:router-topk5-menu-test}).

\begin{table}[tb]
\centering
\footnotesize
\newcommand{\mn}[2]{\makebox[#1][r]{$#2$}}
\newcommand{\ml}[2]{\makebox[#1][l]{$#2$}}
\caption{Except for LLaMA-3.2-3B, the PoLar-split routers switch between identity and one dominant edit and propose far more distinct programs than on our split.
Each fold router on its own test questions, PoLar split, their search, all levels, three search runs and 4 folds ($60$ routers per row).
Top-$5$ menu: distinct programs in the router's top-$5$ sets, range of the per-level means over runs (our split: $6$ to $24$, \cref{tab:router-topk5-menu-test}).
Most common top-$1$: how often a router's most frequent top-$1$ program is its top-$1$, in \% of its test questions; mean over routers (lowest and highest value in brackets).
Identity: share of routers (\%) whose most common top-$1$ is identity.}
\label{tab:their_split_menu}
\begin{tabular}{llccc}
\toprule
Model & Prompt & \shortstack{Top-$5$\\menu} & \shortstack{Most common\\top-$1$ (\%)} & \shortstack{Identity\\(\%)} \\
\midrule
\multirow{2}{*}{Qwen1.5-MoE-A2.7B} & \texttt{polar\_default} & \mn{1em}{31}--\ml{1.5em}{45} & \mn{1em}{61} [\mn{1em}{27}--\mn{1.5em}{100}] & \mn{1.5em}{50} \\
 & \texttt{oneshot} & \mn{1em}{30}--\ml{1.5em}{44} & \mn{1em}{70} [\mn{1em}{31}--\mn{1.5em}{100}] & \mn{1.5em}{67} \\
\midrule
\multirow{2}{*}{Qwen2.5-3B} & \texttt{polar\_default} & \mn{1em}{57}--\ml{1.5em}{101} & \mn{1em}{55} [\mn{1em}{21}--\mn{1.5em}{93}] & \mn{1.5em}{40} \\
 & \texttt{oneshot} & \mn{1em}{42}--\ml{1.5em}{63} & \mn{1em}{74} [\mn{1em}{35}--\mn{1.5em}{100}] & \mn{1.5em}{68} \\
\midrule
\multirow{2}{*}{Qwen3-8B} & \texttt{polar\_default} & \mn{1em}{55}--\ml{1.5em}{65} & \mn{1em}{57} [\mn{1em}{26}--\mn{1.5em}{91}] & \mn{1.5em}{52} \\
 & \texttt{oneshot} & \mn{1em}{40}--\ml{1.5em}{53} & \mn{1em}{63} [\mn{1em}{39}--\mn{1.5em}{95}] & \mn{1.5em}{52} \\
\midrule
\multirow{2}{*}{LLaMA-3.2-3B} & \texttt{polar\_default} & \mn{1em}{26}--\ml{1.5em}{48} & \mn{1em}{95} [\mn{1em}{72}--\mn{1.5em}{100}] & \mn{1.5em}{100} \\
 & \texttt{oneshot} & \mn{1em}{30}--\ml{1.5em}{41} & \mn{1em}{95} [\mn{1em}{53}--\mn{1.5em}{100}] & \mn{1.5em}{100} \\
\bottomrule
\end{tabular}
\end{table}

\subsection{Run-to-Run Variance}
\label{sec:appendix_their_split_variance}

Under their setup, the router's gain varies with two random seeds: the seed of the search, which produces the training data, and the seed of the router's own training.
With the search seed changed, the router gain lies between $+0.7$ and $+13.5$pp on Qwen2.5-3B DM-2, and between $-6.9$ and $+1.6$ on Qwen3-8B DM-1 (\cref{tab:their_split_results}).
It varies this much because their search starts every question from the same seed, so a few edits dominate the router's training labels, and which edits dominate depends on the search seed.
With the search data fixed (Qwen2.5-3B DM-2, run 2) and only the router's training seed changed, three seeds give gains between $+7.1$ and $+13.5$pp.
On both models, the fixed program, by contrast, gains on every run (\cref{tab:their_split_results}).

\subsection{Out-of-Distribution Results on the PoLar Split}
\label{sec:appendix_their_split_ood}

\textbf{Setup.}
As \citeauthor{li2026polar} do, we train one \Router\ on DM-1 to DM-5 together (PoLar split, \texttt{polar\_default}, search run 1) and evaluate it zero-shot on ASDiv ($2{,}305$ questions), MAWPS ($520$) and MMLU-Pro ($11{,}622$ questions in the $13$ subjects of their tables).
The fixed program is picked on the validation questions of all five levels.
ASDiv and MAWPS use a generation cap of $50$ tokens, $\tau{=}0$ and no chat template; MMLU-Pro is scored from the logits of the answer letters, without generation.

\begin{table}[tb]
\centering
\scriptsize
\caption{Out of distribution, only Qwen2.5-3B gains much, on ASDiv and MAWPS, by repairing answers cut off before the box; on MMLU-Pro, no router gains.
Accuracy (\%) with \texttt{polar\_default} and search run 1: \Base($\tau{=}0$), \Router, and fixed program, without and with the chat template of their evaluation code. LLaMA-3.2-3B has no chat template, and MMLU-Pro was run without one.}
\label{tab:their_split_ood}
\setlength{\tabcolsep}{3pt}
\begin{tabular}{ll ccc ccc ccc}
\toprule
 & & \multicolumn{3}{c}{ASDiv} & \multicolumn{3}{c}{MAWPS} & \multicolumn{3}{c}{MMLU-Pro} \\
\cmidrule(lr){3-5}\cmidrule(lr){6-8}\cmidrule(lr){9-11}
Model & Setting & \Base & \Router & Fixed & \Base & \Router & Fixed & \Base & \Router & Fixed \\
\midrule
\multirow{2}{*}{Qwen1.5-MoE-A2.7B} & no template & 48.9 & 49.0 & 51.0 & 58.3 & 58.3 & 57.7 & 24.4 & 24.4 & 25.5 \\
 & chat template & 46.3 & 46.3 & 48.9 & 41.5 & 41.5 & 50.8 & -- & -- & -- \\
\midrule
\multirow{2}{*}{Qwen2.5-3B} & no template & 0.4 & 6.8 & 56.8 & 0.6 & 9.0 & 52.5 & 35.4 & 33.6 & 29.7 \\
 & chat template & 40.1 & 61.8 & 59.6 & 36.3 & 54.2 & 52.5 & -- & -- & -- \\
\midrule
\multirow{2}{*}{Qwen3-8B} & no template & 70.6 & 71.8 & 71.2 & 73.8 & 72.7 & 64.0 & 42.1 & 40.2 & 41.1 \\
 & chat template & 47.9 & 66.6 & 43.3 & 52.3 & 70.8 & 36.7 & -- & -- & -- \\
\midrule
\multirow{1}{*}{LLaMA-3.2-3B} & no template & 62.5 & 62.5 & 46.9 & 72.1 & 72.1 & 42.5 & 31.5 & 31.5 & 31.0 \\
\bottomrule
\end{tabular}
\end{table}

\Cref{tab:their_split_ood} gives the out-of-distribution results.
With \texttt{polar\_default}, only Qwen2.5-3B gains much on ASDiv and MAWPS.
Its unmodified model ignores the instruction to answer with a box only, starts to reason and is cut off after $50$ tokens, so almost no answer is graded.
Every question that the router or the fixed program rescues is such a case: they make the model answer first, not solve more.
The routers of Qwen2.5-3B and Qwen3-8B predict the same program, repeating layer $17$ once, for $92$--$97\%$ of the ASDiv and MAWPS questions and $77$--$82\%$ of the MMLU-Pro questions,
so they act as a fixed program, though not the one picked on validation.
The Qwen1.5-MoE-A2.7B and LLaMA-3.2-3B routers predict identity for almost every question ($\geq 99.7\%$).
For the other models the changes are small or negative.
MMLU-Pro needs no generated answer, so there is no output format to repair, and the routers do not gain there.

Other fixes remove the same failure.
The chat template of their evaluation code also makes Qwen2.5-3B answer, but it lowers Qwen3-8B and Qwen1.5-MoE-A2.7B, whose router or fixed program then recovers only part of the loss.
\texttt{oneshot} fixes it cheaply (\cref{tab:their_split_ood_protocol}).
With $512$ instead of $50$ tokens, the model has room to reason first and still write the box, so \Base\ also profits from the reasoning: it is then above what \citeauthor{li2026polar} report for \gls{polar}.
The gain at $50$ tokens can therefore be a repair of answers that are cut off before the box.

\begin{table}[tb]
\centering
\small
\caption{Qwen2.5-3B's near-zero \Base\ at $50$ tokens is a format failure: with \texttt{oneshot} or $512$ tokens it is far higher, and with $512$ tokens above what \citeauthor{li2026polar} report for \gls{polar}.
\Base($\tau{=}0$) accuracy (\%) of Qwen2.5-3B on ASDiv and MAWPS. \citeauthor{li2026polar}: the numbers they report.}
\label{tab:their_split_ood_protocol}
\begin{tabular}{lcc}
\toprule
Setting & ASDiv & MAWPS \\
\midrule
\texttt{polar\_default}, $50$ tokens & 0.4 & 0.6 \\
\texttt{polar\_default}, $512$ tokens & 89.8 & 69.6 \\
\texttt{oneshot}, $50$ tokens & 68.1 & 62.9 \\
Chat template, $50$ tokens & 40.1 & 36.3 \\
Chat template, $512$ tokens & 83.3 & 63.3 \\
\citeauthor{li2026polar}, \Base & 49.5 & 36.2 \\
\citeauthor{li2026polar}, \gls{polar} & 78.1 & 57.7 \\
\bottomrule
\end{tabular}
\end{table}

\subsection{Summary}
\label{sec:appendix_their_split_summary}

Under their setup, we cannot reproduce the router gains that \citeauthor{li2026polar} report on every configuration.
For some models and levels they do not appear, for others they depend on the run.
The gains that do appear come from repairing an output format, not from better reasoning: with \texttt{polar\_default}, some models start to reason instead of answering with a box, and either run out of tokens or write the placeholder.
Once the format is correct (\texttt{oneshot}), router and fixed program no longer gain.
Unlike on our split, the router does not collapse to identity (except for LLaMA-3.2-3B; out of distribution, also Qwen1.5-MoE-A2.7B).
Their search digs deep and starts every question from the same seed, which together make a few edits dominate the training labels.
The router's most common output is the average of its labels, one of these edits, which it gives to about $40\%$ of the questions, and identity to most of the rest.
On our split, our search spreads its edits over many layers, so at every layer \textsf{keep} dominates the labels and the average is identity.
Their search concentrates its edits on a few layers, so there one edit dominates.
Out of distribution, only Qwen2.5-3B gains much, and this is the same repair.
In total, the gains come from repairing this format error, and several fixes do so:
a fixed program that makes the model box its answer first, a router that mostly predicts such a program, the one-shot example of the \texttt{oneshot} prompt, the chat template (out of distribution),
or a budget of $512$ tokens.
The one-shot example is the cheapest and most robust;
the $512$-token budget is not a like-for-like comparison, since it also leaves the model room to reason.
As in our setup, the programs exist, but a router that picks among them per input remains to be found (\cref{sec:conclusion}).

\section{Compute Infrastructure and the Error-Tagging Pipeline}
\glsresetall

\subsection{Compute Infrastructure and Hardware}
\label{sec:appendix_hardware}

Experiments ran on a shared university GPU cluster spanning A100, H100, H200, and B200 nodes.
Each \gls{mcts} run of the main table is pinned to a single GPU architecture, since bf16 non-associativity makes greedy decoding diverge slightly across architectures (\cref{sec:appendix_reward_noise}).
In total, this paper's reported results used roughly 5,400 GPU-hours (see \cref{tab:hardware-by-phase}), plus separate \gls{llm} usage detailed below.

\begin{table}[h]
\centering
\small
\caption{GPU-hours by what the compute went toward (rounded).}
\label{tab:hardware-by-phase}
\begin{tabular}{p{12cm}r}
\toprule
Stage & GPU-hours \\
\midrule
\multicolumn{2}{l}{\emph{Main track: our split, our search}} \\
\gls{mcts} (6 models, 5 difficulties, 3 seeds) & $\sim$2{,}050 \\
Routers, fixed programs and recipe selection (\cref{sec:appendix_router_recipe}) & $\sim$940 \\
Menu evaluation (\hyperref[finding:5]{\ftag{F5}}) & $\sim$115 \\
\midrule
\multicolumn{2}{l}{\emph{PoLar split: their search (\cref{sec:appendix_their_split})}} \\
\gls{mcts} (4 models, 2 prompts, 3 seeds) & $\sim$1{,}120 \\
Routers, fixed programs, base evaluations and split construction & $\sim$560 \\
Cross-checks: their search on our split, our search on the PoLar split (Qwen3-8B, LLaMA-3.2-3B), with routers & $\sim$160 \\
\midrule
\multicolumn{2}{l}{\emph{MMLU-Pro track (\cref{sec:mmlu_pro_track})}} \\
\gls{mcts} (14 domains, Qwen3-8B) & $\sim$40 \\
Routers and recipe selection, letter-bias checks (\cref{sec:appendix_mmlu_letter_bias}) & $\sim$150 \\
\midrule
\multicolumn{2}{l}{\emph{Supporting analyses}} \\
Baseline reproduction and \gls{ood} transfer (\cref{sec:baseline_reproduction}) & $\sim$130 \\
Prompt-format sweep (\cref{sec:appendix_placeholder_echo}) & $\sim$60 \\
Other checks (\gls{ood} on the PoLar split, Qwen3-32B token budget, router seed, protocol, chat template, \gls{mcts} engine, further small checks) & $\sim$100 \\
\bottomrule
\end{tabular}
\end{table}

The \gls{llm} used for qualitative error tagging runs entirely off this cluster on separate internally hosted infrastructure (one DGX-class node, 8 $\times$ A100 40GB) and is not included in the GPU-hours.
Across its error-categorization use and a separate solver competence check (\cref{sec:appendix_bht_large}), it used roughly 90M tokens in total.

\subsection{\texttt{MiniMax-M2.7-AWQ} as \gls{llm}-Assisted Error Tagger}
\label{sec:appendix_bht_large}

The qualitative error analysis in \cref{sec:analysis} uses an internally hosted \gls{llm} (a locally served, AWQ-4-bit-quantized MiniMax-M2.7 checkpoint, \texttt{cyankiwi/MiniMax-M2.7-AWQ-4bit} on Hugging Face, abbreviated \texttt{MiniMax-M2.7-AWQ} here) for one narrow purpose: labeling how close a wrong answer already is to the ground truth, \errtag{related} or \errtag{unrelated} (\emph{Closeness classification} below).
It first writes a short explanation of the base program's error as scratch reasoning; only its \errtag{related}/\errtag{unrelated} label is used.
\texttt{MiniMax-M2.7-AWQ} never decides correctness itself, and takes no part in the mechanical error taxonomy below.

The tagger is only ever shown the exact text the grader scored (by design, so it cannot critique content that was never causal to the grade), and base models frequently emit the boxed answer \emph{before} any reasoning, so there is usually no genuine derivation left to show once trailing hallucinated continuation is stripped.

\paragraph{Taxonomy.} \texttt{MiniMax-M2.7-AWQ} plays no role in this step.
We derive the error taxonomy mechanically, in code, from the same boxed-answer extraction our strict grader uses:
\errtag{unboxed} (no \verb|\boxed{}| at all), \errtag{malformed boxed} (an empty \verb|\boxed{}|), and \errtag{wrong final answer} (a non-empty boxed answer that simply differs from ground truth), plus a deterministic \errtag{prompt template echo} pre-filter for answers that parrot our own prompt's unfilled \verb|\boxed{ANSWER}| placeholder (\cref{sec:appendix_placeholder_echo} covers this failure mode in more depth).
The paper's full-corpus tagging pass (all 6 models, all 5 DART-Math difficulties) resolves to $90.4\%$ \errtag{wrong final answer} overall, with \errtag{unboxed} concentrated almost entirely in one model (Qwen3-32B, $59.3\%$ of its own errors, vs.\ $\le 1.1\%$ for every other model); \cref{tab:error_categories} below gives the full per-model breakdown, and \cref{sec:analysis_beyond_polar} discusses what it means for \hyperref[finding:6]{\ftag{F6}}.

\paragraph{Closeness classification.} For the \errtag{wrong final answer} category specifically, \texttt{MiniMax-M2.7-AWQ}
classifies how close the model's wrong boxed answer is to the ground-truth answer, shown only the two answers: \errtag{related}
if there is any discernible connection, such as close in value, same order of magnitude, a clean scaling or sign relationship (e.g.\ off by a constant factor, or a sign flip), or the same expression type with a different coefficient.
It is \errtag{unrelated} otherwise.

\paragraph{Solver competence check.} As a sanity check on whether \texttt{MiniMax-M2.7-AWQ} has any real competence to inform its labels (not proof that they are correct), we separately benchmark it as a \emph{solver} on our own DART-Math split (\cref{sec:appendix_dart_math}).
Over the full $10{,}000$-question run (all five difficulty levels) under \texttt{oneshot} (\cref{sec:appendix_placeholder_echo}), with generation capped at $32{,}000$ tokens since the model runs in thinking mode (our usual $50$-token budget, sized for non-reasoning models, would not fit a thinking trace at all), it reaches $96.15\%$ accuracy ($99.4\%$ of responses contain a \verb|\boxed{}| at all, $0.05\%$ an empty one, $0.66\%$ a prompt-template-echo placeholder).

\paragraph{Truncation and rescue.} Among the models being tagged, Qwen3-32B starts to reason far more often than the other five and is then cut off after $50$ tokens before any \verb|\boxed{}|, the same \errtag{unboxed} failure as in \cref{sec:appendix_placeholder_echo}, which also explains its fixed-program gain (\cref{sec:appendix_q32b_budget}).
None of these truncated generations contain a literal \texttt{<think>} tag.
\Cref{tab:closeness_rescue} reports, for every \errtag{wrong final answer} record, whether \gls{mcts} found at least one valid fixing program (rescued), against the closeness label: programs correct a numerically close miss more easily than a structurally different one, in every model ($57.9$--$83.1\%$ vs.\ $36.9$--$62.7\%$).

\begin{table}[tb]
\centering
\footnotesize
\caption{Nearly all errors are wrong final answers, except on Qwen3-32B, where $59\%$ are unboxed answers cut off at the $50$-token cap.
Error categories of every identity-wrong record, all 5 DART-Math difficulties pooled, per model; taxonomy above.}
\label{tab:error_categories}
\begin{tabular}{lccccc}
\toprule
Model & \errtag{wrong final answer} & \errtag{unboxed} & \errtag{prompt template echo} & \errtag{malformed boxed} & Total \\
\midrule
Qwen1.5-MoE-A2.7B & 7{,}953 (98.7\%) & 87 (1.1\%) & 1 (0.0\%) & 16 (0.2\%) & 8{,}057 \\
Qwen2.5-3B & 7{,}668 (99.0\%) & 74 (1.0\%) & 0 (0.0\%) & 0 (0.0\%) & 7{,}742 \\
Qwen2.5-7B & 6{,}299 (100.0\%) & 0 (0.0\%) & 0 (0.0\%) & 3 (0.0\%) & 6{,}302 \\
Qwen3-8B & 6{,}599 (99.9\%) & 0 (0.0\%) & 2 (0.0\%) & 4 (0.1\%) & 6{,}605 \\
Qwen3-32B & 2{,}548 (39.1\%) & 3{,}857 (59.3\%) & 102 (1.6\%) & 2 (0.0\%) & 6{,}509 \\
LLaMA-3.2-3B & 8{,}073 (99.9\%) & 0 (0.0\%) & 0 (0.0\%) & 5 (0.1\%) & 8{,}078 \\
\midrule
All models & 39{,}140 (90.4\%) & 4{,}018 (9.3\%) & 105 (0.2\%) & 30 (0.1\%) & 43{,}293 \\
\bottomrule
\end{tabular}
\end{table}

\begin{table}[tb]
\centering
\footnotesize
\caption{Programs rescue a wrong answer that is close to the ground truth more often than an unrelated one, on every model ($57.9$--$83.1\%$ vs.\ $36.9$--$62.7\%$).
Rescue rate (share of \errtag{wrong final answer} records for which \gls{mcts}, search run seed $42$, found at least one valid fixing program) by closeness label, all 5 difficulties pooled, per model.
Related and unrelated $n$ add up to $38{,}802$ of the $39{,}140$ wrong final answers in \cref{tab:error_categories};
the other $338$ have no closeness label, because the tagger ran out of its $4{,}000$-token reasoning budget ($336$) or timed out ($2$).}
\label{tab:closeness_rescue}
\begin{tabular}{lcccc}
\toprule
Model & Related $n$ & Related rescued & Unrelated $n$ & Unrelated rescued \\
\midrule
Qwen1.5-MoE-A2.7B & 6{,}653 & 57.9\% & 1{,}215 & 41.3\% \\
Qwen2.5-3B & 6{,}822 & 64.3\% & 784 & 36.9\% \\
Qwen2.5-7B & 5{,}661 & 66.3\% & 587 & 37.6\% \\
Qwen3-8B & 5{,}841 & 63.5\% & 700 & 43.6\% \\
Qwen3-32B & 2{,}355 & 83.1\% & 185 & 62.7\% \\
LLaMA-3.2-3B & 6{,}907 & 59.8\% & 1{,}092 & 38.4\% \\
\midrule
All models & 34{,}239 & 63.7\% & 4{,}563 & 40.6\% \\
\bottomrule
\end{tabular}
\end{table}

\section{Additional \gls{mcts} Program-Analysis Results}
\label{sec:appendix_analysis_extra}
\glsresetall

This section collects the rest of our own \gls{mcts} reproduction results that did not fit in \cref{sec:analysis}, with our search on our split, on the same 6 models and 5 DART-Math difficulties.
It has two parts: reproductions of several of \citet{li2026polar}'s own diagnostic figures, matching what \cref{sec:analysis} reports numerically (\cref{sec:appendix_analysis_polar_figs}), and additional analyses with no equivalent in \gls{polar}'s paper, extending several main-body findings to their full per-model, per-difficulty detail (\cref{sec:appendix_analysis_no_polar}).

\subsection{Li et al.'s Figure Reproductions}
\label{sec:appendix_analysis_polar_figs}

\begin{figure}[htb]
\centering
\footnotesize
\legenditem{polQwen15moe}{Qwen1.5-MoE-A2.7B}\quad\legenditem{polQwen253b}{Qwen2.5-3B}\quad
\legenditem{polQwen257b}{Qwen2.5-7B}\quad\legenditem{polQwen38b}{Qwen3-8B}\quad
\legenditem{polQwen332b}{Qwen3-32B}\quad\legenditem{polLlama32}{LLaMA-3.2-3B}\\[0.3em]
\normalsize
\begin{subfigure}[t]{0.3816\linewidth}\centering
\includegraphics[width=1.0000\linewidth]{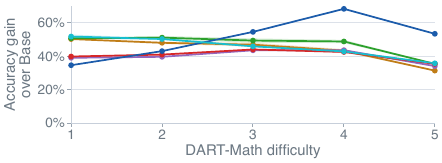}
\caption{Skip\&Repeat gain over \Base.}
\label{fig:accuracy_gain_over_base_mcts}
\end{subfigure}%
\hspace{38.3pt}%
\begin{subfigure}[t]{0.3816\linewidth}\centering
\includegraphics[width=1.0000\linewidth]{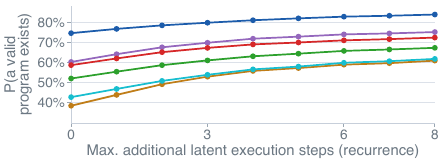}
\caption{Share of questions with $\geq\!1$ valid program as the recurrence budget grows.}
\label{fig:analysis_fig5a}
\end{subfigure}
\caption{Skip\&Repeat gains at every difficulty on every model (a), and each extra recurrence step makes more questions solvable (b).
(a) Accuracy gained by allowing skip and repeat together, over the standard
forward pass alone, at each DART-Math difficulty level (1 = easiest, 5 = hardest); same
Gain column as \cref{tab:analysis_table1} above. (b) Share of questions with at least one
valid program, as more extra latent execution steps (recurrence) are allowed beyond the
standard forward pass depth, pooled over all difficulties, the recurrence-budget half of
\hyperref[finding:3]{\ftag{F3}}. One line per model in both
panels. Lines are the mean over the three search runs, bands are $\pm$ 1 std.}
\label{fig:analysis_f1}
\end{figure}

\begin{figure}[tb]
\centering
\scriptsize
\legenditem{polOrigPath}{Original program}\,\legenditem{polDepth90}{$\le$90\% depth}\,
\legenditem{polDepth95}{$\le$95\% depth}\,\legenditem{polDepth100}{$\le$100\% depth}\,
\legenditem{polDepth105}{$\le$105\% depth}\,\legenditem{polDepth110}{$\le$110\% depth}\,
\legenditem{polDepth115}{$\le$115\% depth}\\[0.3em]
\normalsize
\begin{subfigure}{0.1996\linewidth}\centering
\includegraphics[width=0.9948\linewidth]{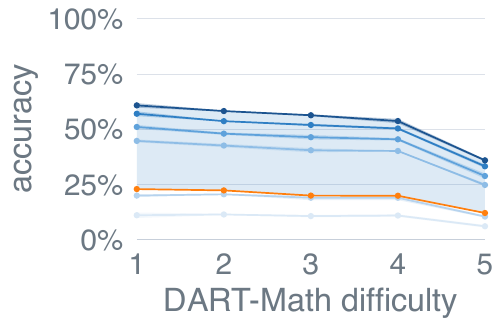}
\caption{\scriptsize\mbox{Qwen1.5-MoE-A2.7B}}\end{subfigure}%
\hspace{4.4pt}%
\begin{subfigure}{0.1490\linewidth}\centering
\includegraphics[width=1.0000\linewidth]{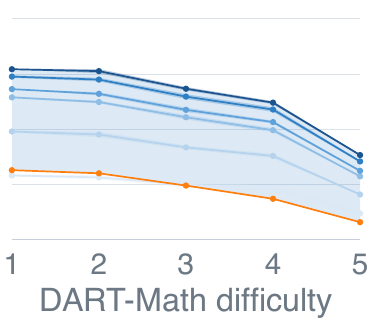}
\caption{\scriptsize\mbox{Qwen2.5-3B}}\end{subfigure}%
\hspace{4.6pt}%
\begin{subfigure}{0.1490\linewidth}\centering
\includegraphics[width=1.0000\linewidth]{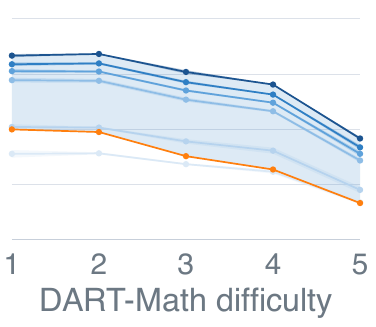}
\caption{\scriptsize\mbox{Qwen2.5-7B}}\end{subfigure}%
\hspace{4.6pt}%
\begin{subfigure}{0.1490\linewidth}\centering
\includegraphics[width=1.0000\linewidth]{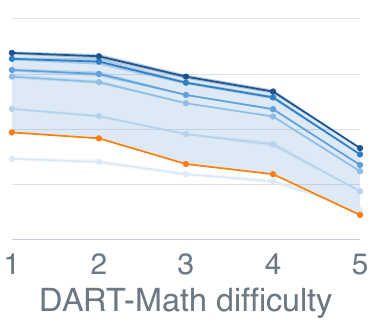}
\caption{\scriptsize\mbox{Qwen3-8B}}\end{subfigure}%
\hspace{4.6pt}%
\begin{subfigure}{0.1490\linewidth}\centering
\includegraphics[width=1.0000\linewidth]{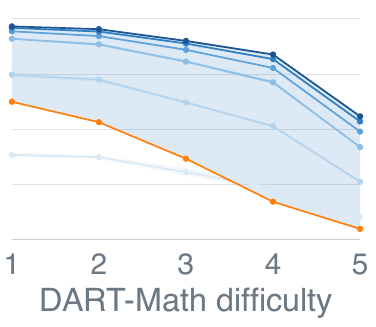}
\caption{\scriptsize\mbox{Qwen3-32B}}\end{subfigure}%
\hspace{3.8pt}%
\begin{subfigure}{0.1526\linewidth}\centering
\includegraphics[width=0.9763\linewidth]{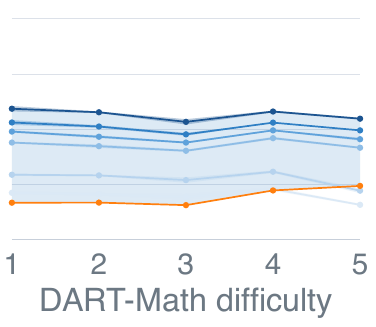}
\caption{\scriptsize\mbox{LLaMA-3.2-3B}}\end{subfigure}
\vspace{-4pt}
\caption{\hyperref[finding:3]{\ftag{F3}}\ Allowing programs up to $115\%$ of the standard depth gains $34$--$49$pp over identity on average, on every model.
The recurrence-budget evidence, from a depth-budget angle: best accuracy achievable using only a program (identity or found) whose executed length is at most the given budget, as a percentage of the standard forward pass depth (DART-Math difficulty 1 = easiest to 5 = hardest). Lines are the mean over the three search runs, bands are $\pm$ 1 std. The Original program (identity) does not depend on the search, so it has no band.
The shaded area is the gap between the Original program and the $\le$115\% line.}
\label{fig:appendix_fig3}
\end{figure}

\Cref{tab:analysis_table1,fig:analysis_f1} reproduce \citet{li2026polar}'s own
skip-and-repeat gain and recurrence-scaling results; their recurrence-vs-skip figure is
instead embedded in \cref{sec:analysis} as \cref{fig:appendix_fig5b}, alongside
\hyperref[finding:3]{\ftag{F3}}.
\Cref{fig:appendix_fig3} extends this with the full execution-depth-budget breakdown, and
\cref{tab:analysis_f4} reports the segment-length share behind
\hyperref[finding:4]{\ftag{F4}}.

\begin{table}[htb]
\centering
\footnotesize
\caption{Combining skip and repeat always beats either alone (\hyperref[finding:1]{\ftag{F1}}):
Skip\&Repeat's gain over \Base ranges from $+31.4$ to $+68.4$pp across all 6 models and every DART-Math difficulty.
\Base\ = identity already correct. Skip = identity or a skip-only valid program exists. Repeat = identity or a repeat-only valid program exists.
Skip\&Repeat = identity or any valid program exists. Gain = Skip\&Repeat $-$ \Base.
Skip, Repeat, and Skip\&Repeat: mean $\pm$ sample standard deviation over the three search runs.}
\label{tab:analysis_table1}
\begin{tabular}{ll r *{3}{r@{\,}c@{\,}l} r}
\toprule
Model & Difficulty & \Base & \multicolumn{3}{c}{Skip} & \multicolumn{3}{c}{Repeat} & \multicolumn{3}{c}{Skip\&Repeat} & Gain \\
\midrule
\multirow{5}{*}{Qwen1.5-MoE-A2.7B} & DM-1 & 22.7\% & 35.6\% & {\scriptsize$\pm$} & {\scriptsize 0.7} & \underline{67.0\%} & {\scriptsize$\pm$} & {\scriptsize 0.4} & \textbf{73.3\%} & {\scriptsize$\pm$} & {\scriptsize 0.4} & \textcolor{polgreen}{+50.6} \\
 & DM-2 & 22.8\% & 34.9\% & {\scriptsize$\pm$} & {\scriptsize 1.0} & \underline{64.5\%} & {\scriptsize$\pm$} & {\scriptsize 0.7} & \textbf{70.5\%} & {\scriptsize$\pm$} & {\scriptsize 0.8} & \textcolor{polgreen}{+47.7} \\
 & DM-3 & 19.6\% & 32.8\% & {\scriptsize$\pm$} & {\scriptsize 0.9} & \underline{60.6\%} & {\scriptsize$\pm$} & {\scriptsize 0.8} & \textbf{67.0\%} & {\scriptsize$\pm$} & {\scriptsize 0.8} & \textcolor{polgreen}{+47.4} \\
 & DM-4 & 19.9\% & 32.2\% & {\scriptsize$\pm$} & {\scriptsize 0.3} & \underline{58.0\%} & {\scriptsize$\pm$} & {\scriptsize 0.9} & \textbf{63.3\%} & {\scriptsize$\pm$} & {\scriptsize 0.9} & \textcolor{polgreen}{+43.4} \\
 & DM-5 & 12.2\% & 19.6\% & {\scriptsize$\pm$} & {\scriptsize 0.4} & \underline{39.2\%} & {\scriptsize$\pm$} & {\scriptsize 0.4} & \textbf{43.6\%} & {\scriptsize$\pm$} & {\scriptsize 0.3} & \textcolor{polgreen}{+31.4} \\
\midrule
\multirow{5}{*}{Qwen2.5-3B} & DM-1 & 31.6\% & 54.7\% & {\scriptsize$\pm$} & {\scriptsize 0.3} & \underline{75.5\%} & {\scriptsize$\pm$} & {\scriptsize 0.8} & \textbf{82.3\%} & {\scriptsize$\pm$} & {\scriptsize 0.3} & \textcolor{polgreen}{+50.6} \\
 & DM-2 & 30.1\% & 52.1\% & {\scriptsize$\pm$} & {\scriptsize 1.3} & \underline{73.5\%} & {\scriptsize$\pm$} & {\scriptsize 0.4} & \textbf{81.3\%} & {\scriptsize$\pm$} & {\scriptsize 1.0} & \textcolor{polgreen}{+51.2} \\
 & DM-3 & 24.6\% & 45.7\% & {\scriptsize$\pm$} & {\scriptsize 1.0} & \underline{65.6\%} & {\scriptsize$\pm$} & {\scriptsize 0.4} & \textbf{74.0\%} & {\scriptsize$\pm$} & {\scriptsize 1.3} & \textcolor{polgreen}{+49.4} \\
 & DM-4 & 18.6\% & 40.3\% & {\scriptsize$\pm$} & {\scriptsize 0.3} & \underline{60.0\%} & {\scriptsize$\pm$} & {\scriptsize 0.6} & \textbf{67.5\%} & {\scriptsize$\pm$} & {\scriptsize 1.0} & \textcolor{polgreen}{+48.9} \\
 & DM-5 & 8.0\% & 21.7\% & {\scriptsize$\pm$} & {\scriptsize 0.4} & \underline{36.4\%} & {\scriptsize$\pm$} & {\scriptsize 0.4} & \textbf{43.5\%} & {\scriptsize$\pm$} & {\scriptsize 0.4} & \textcolor{polgreen}{+35.5} \\
\midrule
\multirow{5}{*}{Qwen2.5-7B} & DM-1 & 50.0\% & 65.2\% & {\scriptsize$\pm$} & {\scriptsize 1.2} & \underline{84.6\%} & {\scriptsize$\pm$} & {\scriptsize 0.3} & \textbf{89.0\%} & {\scriptsize$\pm$} & {\scriptsize 0.2} & \textcolor{polgreen}{+39.1} \\
 & DM-2 & 48.6\% & 65.0\% & {\scriptsize$\pm$} & {\scriptsize 1.0} & \underline{83.9\%} & {\scriptsize$\pm$} & {\scriptsize 0.3} & \textbf{88.5\%} & {\scriptsize$\pm$} & {\scriptsize 0.2} & \textcolor{polgreen}{+39.9} \\
 & DM-3 & 38.0\% & 55.9\% & {\scriptsize$\pm$} & {\scriptsize 0.8} & \underline{75.8\%} & {\scriptsize$\pm$} & {\scriptsize 0.5} & \textbf{81.4\%} & {\scriptsize$\pm$} & {\scriptsize 0.2} & \textcolor{polgreen}{+43.5} \\
 & DM-4 & 31.8\% & 50.9\% & {\scriptsize$\pm$} & {\scriptsize 1.0} & \underline{69.4\%} & {\scriptsize$\pm$} & {\scriptsize 0.9} & \textbf{75.5\%} & {\scriptsize$\pm$} & {\scriptsize 0.4} & \textcolor{polgreen}{+43.7} \\
 & DM-5 & 16.6\% & 29.6\% & {\scriptsize$\pm$} & {\scriptsize 1.0} & \underline{44.0\%} & {\scriptsize$\pm$} & {\scriptsize 0.2} & \textbf{50.8\%} & {\scriptsize$\pm$} & {\scriptsize 0.4} & \textcolor{polgreen}{+34.1} \\
\midrule
\multirow{5}{*}{Qwen3-8B} & DM-1 & 48.6\% & 66.7\% & {\scriptsize$\pm$} & {\scriptsize 0.9} & \underline{83.5\%} & {\scriptsize$\pm$} & {\scriptsize 0.3} & \textbf{88.5\%} & {\scriptsize$\pm$} & {\scriptsize 0.4} & \textcolor{polgreen}{+39.9} \\
 & DM-2 & 45.9\% & 64.0\% & {\scriptsize$\pm$} & {\scriptsize 1.1} & \underline{81.6\%} & {\scriptsize$\pm$} & {\scriptsize 0.1} & \textbf{86.9\%} & {\scriptsize$\pm$} & {\scriptsize 0.5} & \textcolor{polgreen}{+41.0} \\
 & DM-3 & 34.3\% & 53.7\% & {\scriptsize$\pm$} & {\scriptsize 0.5} & \underline{71.7\%} & {\scriptsize$\pm$} & {\scriptsize 0.1} & \textbf{78.4\%} & {\scriptsize$\pm$} & {\scriptsize 0.5} & \textcolor{polgreen}{+44.1} \\
 & DM-4 & 29.7\% & 48.9\% & {\scriptsize$\pm$} & {\scriptsize 0.9} & \underline{65.5\%} & {\scriptsize$\pm$} & {\scriptsize 0.7} & \textbf{72.3\%} & {\scriptsize$\pm$} & {\scriptsize 0.6} & \textcolor{polgreen}{+42.6} \\
 & DM-5 & 11.2\% & 25.1\% & {\scriptsize$\pm$} & {\scriptsize 0.4} & \underline{40.4\%} & {\scriptsize$\pm$} & {\scriptsize 0.6} & \textbf{46.7\%} & {\scriptsize$\pm$} & {\scriptsize 0.4} & \textcolor{polgreen}{+35.5} \\
\midrule
\multirow{5}{*}{Qwen3-32B} & DM-1 & 62.5\% & 84.6\% & {\scriptsize$\pm$} & {\scriptsize 0.3} & \underline{95.1\%} & {\scriptsize$\pm$} & {\scriptsize 0.3} & \textbf{97.1\%} & {\scriptsize$\pm$} & {\scriptsize 0.2} & \textcolor{polgreen}{+34.6} \\
 & DM-2 & 53.2\% & 81.7\% & {\scriptsize$\pm$} & {\scriptsize 0.4} & \underline{93.8\%} & {\scriptsize$\pm$} & {\scriptsize 0.3} & \textbf{96.3\%} & {\scriptsize$\pm$} & {\scriptsize 0.2} & \textcolor{polgreen}{+43.1} \\
 & DM-3 & 36.7\% & 71.8\% & {\scriptsize$\pm$} & {\scriptsize 0.4} & \underline{87.3\%} & {\scriptsize$\pm$} & {\scriptsize 0.6} & \textbf{91.3\%} & {\scriptsize$\pm$} & {\scriptsize 0.4} & \textcolor{polgreen}{+54.6} \\
 & DM-4 & 17.2\% & 60.0\% & {\scriptsize$\pm$} & {\scriptsize 0.2} & \underline{80.2\%} & {\scriptsize$\pm$} & {\scriptsize 0.4} & \textbf{85.7\%} & {\scriptsize$\pm$} & {\scriptsize 0.4} & \textcolor{polgreen}{+68.4} \\
 & DM-5 & 5.0\% & 31.5\% & {\scriptsize$\pm$} & {\scriptsize 0.6} & \underline{52.0\%} & {\scriptsize$\pm$} & {\scriptsize 0.5} & \textbf{58.4\%} & {\scriptsize$\pm$} & {\scriptsize 0.3} & \textcolor{polgreen}{+53.5} \\
\midrule
\multirow{5}{*}{LLaMA-3.2-3B} & DM-1 & 16.8\% & 33.6\% & {\scriptsize$\pm$} & {\scriptsize 0.7} & \underline{59.1\%} & {\scriptsize$\pm$} & {\scriptsize 0.8} & \textbf{68.7\%} & {\scriptsize$\pm$} & {\scriptsize 1.2} & \textcolor{polgreen}{+51.9} \\
 & DM-2 & 16.9\% & 33.1\% & {\scriptsize$\pm$} & {\scriptsize 0.5} & \underline{57.4\%} & {\scriptsize$\pm$} & {\scriptsize 0.8} & \textbf{67.2\%} & {\scriptsize$\pm$} & {\scriptsize 0.5} & \textcolor{polgreen}{+50.3} \\
 & DM-3 & 15.7\% & 31.4\% & {\scriptsize$\pm$} & {\scriptsize 1.0} & \underline{52.4\%} & {\scriptsize$\pm$} & {\scriptsize 0.7} & \textbf{61.7\%} & {\scriptsize$\pm$} & {\scriptsize 0.7} & \textcolor{polgreen}{+46.0} \\
 & DM-4 & 22.4\% & 38.5\% & {\scriptsize$\pm$} & {\scriptsize 0.4} & \underline{58.1\%} & {\scriptsize$\pm$} & {\scriptsize 1.4} & \textbf{65.3\%} & {\scriptsize$\pm$} & {\scriptsize 0.9} & \textcolor{polgreen}{+42.9} \\
 & DM-5 & 24.4\% & 36.4\% & {\scriptsize$\pm$} & {\scriptsize 0.2} & \underline{55.3\%} & {\scriptsize$\pm$} & {\scriptsize 0.4} & \textbf{60.2\%} & {\scriptsize$\pm$} & {\scriptsize 0.1} & \textcolor{polgreen}{+35.8} \\
\bottomrule
\end{tabular}
\end{table}

\begin{table}[tb]
\centering
\footnotesize
\caption{\hyperref[finding:4]{\ftag{F4}}\ With our search, single-layer segments are below \citeauthor{li2026polar}'s $57.7\%$ on all 6 models ($21$--$46\%$).
Segment-length share of the shortest valid program per solved question, pooled over all 5 DART-Math difficulties, per model.
\citet{li2026polar} report $57.7\%$ single-layer and over two-thirds at most two layers.}
\label{tab:analysis_f4}
\begin{tabular}{lr@{\,}c@{\,}rr@{\,}c@{\,}lr@{\,}c@{\,}l}
\toprule
Model & \multicolumn{3}{c}{$n$ segments} & \multicolumn{3}{c}{Single-layer} & \multicolumn{3}{c}{At-most-two-layers} \\
\midrule
Qwen1.5-MoE-A2.7B & $9{,}014$ & {\scriptsize$\pm$} & {\scriptsize 103} & $45.5\%$ & {\scriptsize$\pm$} & {\scriptsize 0.8} & $73.5\%$ & {\scriptsize$\pm$} & {\scriptsize 0.2} \\
Qwen2.5-3B & $12{,}512$ & {\scriptsize$\pm$} & {\scriptsize 84} & $32.1\%$ & {\scriptsize$\pm$} & {\scriptsize 0.7} & $58.1\%$ & {\scriptsize$\pm$} & {\scriptsize 0.3} \\
Qwen2.5-7B & $12{,}483$ & {\scriptsize$\pm$} & {\scriptsize 71} & $37.8\%$ & {\scriptsize$\pm$} & {\scriptsize 0.4} & $64.0\%$ & {\scriptsize$\pm$} & {\scriptsize 0.8} \\
Qwen3-8B & $13{,}491$ & {\scriptsize$\pm$} & {\scriptsize 16} & $32.6\%$ & {\scriptsize$\pm$} & {\scriptsize 0.3} & $58.5\%$ & {\scriptsize$\pm$} & {\scriptsize 0.1} \\
Qwen3-32B & $18{,}774$ & {\scriptsize$\pm$} & {\scriptsize 71} & $21.2\%$ & {\scriptsize$\pm$} & {\scriptsize 0.1} & $43.0\%$ & {\scriptsize$\pm$} & {\scriptsize 0.3} \\
LLaMA-3.2-3B & $10{,}644$ & {\scriptsize$\pm$} & {\scriptsize 91} & $36.5\%$ & {\scriptsize$\pm$} & {\scriptsize 0.4} & $64.2\%$ & {\scriptsize$\pm$} & {\scriptsize 0.3} \\
\midrule
\gls{polar} (reference) & \multicolumn{3}{c}{--} & \multicolumn{3}{r}{$57.7\%$} & \multicolumn{3}{r}{over two-thirds} \\
\bottomrule
\end{tabular}
\end{table}

\FloatBarrier

\subsection{Results with No Equivalent in Li et al.}
\label{sec:appendix_analysis_no_polar}

\begin{figure}[htbp]
\centering
\footnotesize
\legenditem{polD1}{DM-1}\quad\legenditem{polD2}{DM-2}\quad
\legenditem{polD3}{DM-3}\quad\legenditem{polD4}{DM-4}\quad
\legenditem{polD5}{DM-5}\\[0.3em]
\normalsize
\begin{subfigure}{0.2950\linewidth}\centering
\includegraphics[width=1.0000\linewidth]{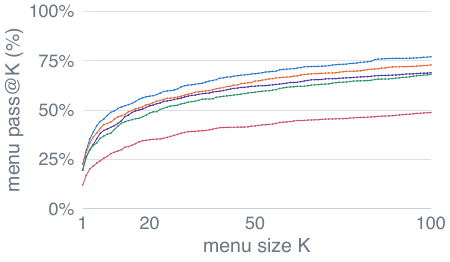}
\caption{\scriptsize\mbox{Qwen1.5-MoE-A2.7B}}\end{subfigure}%
\hspace{16.9pt}%
\begin{subfigure}{0.2950\linewidth}\centering
\includegraphics[width=1.0000\linewidth]{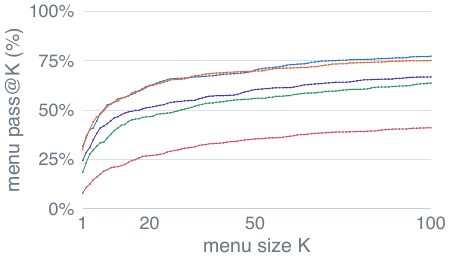}
\caption{\scriptsize\mbox{Qwen2.5-3B}}\end{subfigure}%
\hspace{16.9pt}%
\begin{subfigure}{0.2950\linewidth}\centering
\includegraphics[width=1.0000\linewidth]{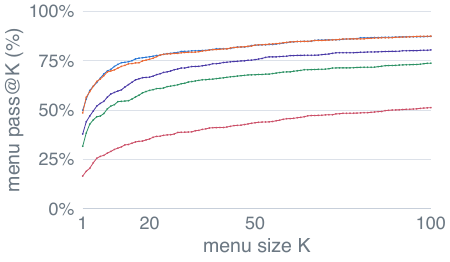}
\caption{\scriptsize\mbox{Qwen2.5-7B}}\end{subfigure}\par\vspace{4pt}
\begin{subfigure}{0.2950\linewidth}\centering
\includegraphics[width=1.0000\linewidth]{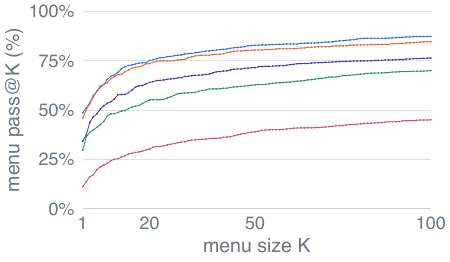}
\caption{\scriptsize\mbox{Qwen3-8B}}\end{subfigure}%
\hspace{16.9pt}%
\begin{subfigure}{0.2950\linewidth}\centering
\includegraphics[width=1.0000\linewidth]{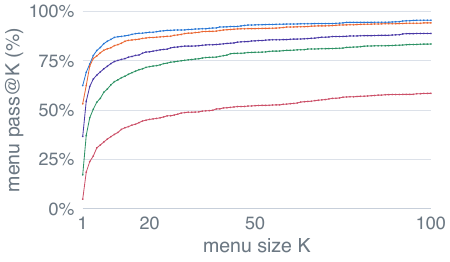}
\caption{\scriptsize\mbox{Qwen3-32B}}\end{subfigure}%
\hspace{16.9pt}%
\begin{subfigure}{0.2950\linewidth}\centering
\includegraphics[width=1.0000\linewidth]{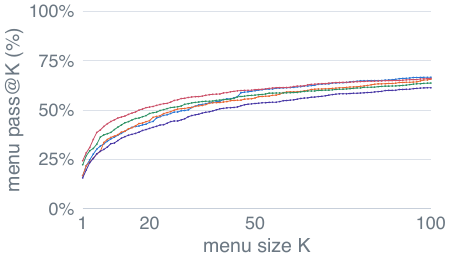}
\caption{\scriptsize\mbox{LLaMA-3.2-3B}}\end{subfigure}
\vspace{-4pt}
\caption{\hyperref[finding:5]{\ftag{F5}}\ A larger menu solves more questions, up to about $2.6\times$ identity's accuracy at $K{=}100$, on average over models and difficulties.
Menu pass@$K$ (share of questions that at least one of its $K$ programs solves) as the menu size $K$ grows (one search run, seed 42), one line per
DART-Math difficulty, one panel per model, with $K$ on a log scale.}
\label{fig:menu_topk}
\end{figure}

\begin{figure}[htbp]
\centering
\begin{subfigure}{0.2888\linewidth}\centering
\includegraphics[width=1.0000\linewidth]{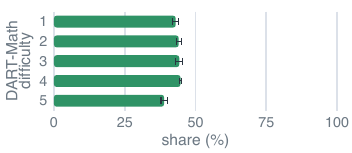}
\caption{\scriptsize\mbox{Qwen1.5-MoE-A2.7B}}\end{subfigure}%
\hspace{30.8pt}%
\begin{subfigure}{0.2888\linewidth}\centering
\includegraphics[width=1.0000\linewidth]{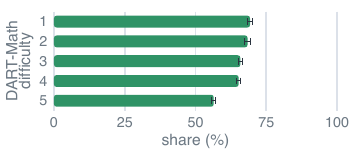}
\caption{\scriptsize\mbox{Qwen2.5-3B}}\end{subfigure}%
\hspace{30.8pt}%
\begin{subfigure}{0.2888\linewidth}\centering
\includegraphics[width=1.0000\linewidth]{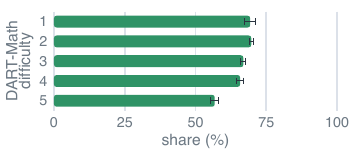}
\caption{\scriptsize\mbox{Qwen2.5-7B}}\end{subfigure}%
\par\vspace{4pt}
\begin{subfigure}{0.2888\linewidth}\centering
\includegraphics[width=1.0000\linewidth]{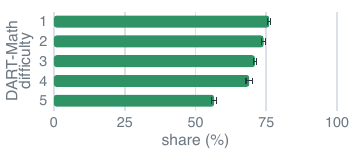}
\caption{\scriptsize\mbox{Qwen3-8B}}\end{subfigure}%
\hspace{30.8pt}%
\begin{subfigure}{0.2888\linewidth}\centering
\includegraphics[width=1.0000\linewidth]{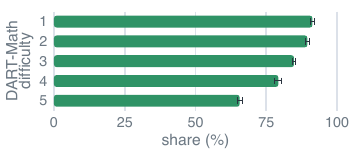}
\caption{\scriptsize\mbox{Qwen3-32B}}\end{subfigure}%
\hspace{30.8pt}%
\begin{subfigure}{0.2888\linewidth}\centering
\includegraphics[width=1.0000\linewidth]{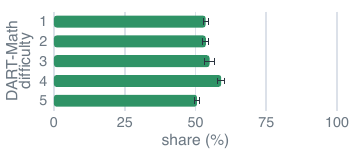}
\caption{\scriptsize\mbox{LLaMA-3.2-3B}}\end{subfigure}
\caption{\hyperref[finding:2]{\ftag{F2}}\ On 5 of 6 models, most solved questions at every difficulty admit a valid program shorter than the standard forward pass;
Qwen1.5-MoE-A2.7B stays below half.
Share of solved questions admitting a shorter program, per model and DART-Math difficulty.
\citet{li2026polar} instead report this share pooled over difficulty and split by whether the standard pass was correct (C$\to$C) or wrong (W$\to$C), so their numbers are not directly comparable.
One panel per model. Rows are the difficulty, 1 (easiest) to 5 (hardest). Bars are the mean over the three search runs, whiskers are $\pm$ 1 std.
Same 0--100\% scale in all panels.}
\label{fig:appendix_shorteridentity_diff_rate}
\end{figure}

\begin{figure}[htbp]
\centering
\begin{subfigure}{0.2956\linewidth}\centering
\includegraphics[width=1.0000\linewidth]{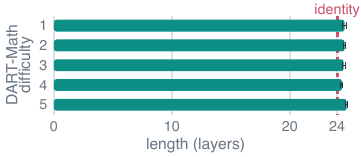}
\caption{\scriptsize\mbox{Qwen1.5-MoE-A2.7B}}\end{subfigure}%
\hspace{26.0pt}%
\begin{subfigure}{0.2956\linewidth}\centering
\includegraphics[width=1.0000\linewidth]{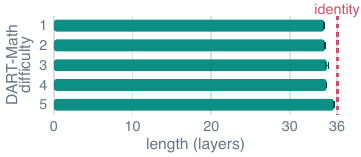}
\caption{\scriptsize\mbox{Qwen2.5-3B}}\end{subfigure}%
\hspace{26.0pt}%
\begin{subfigure}{0.2956\linewidth}\centering
\includegraphics[width=1.0000\linewidth]{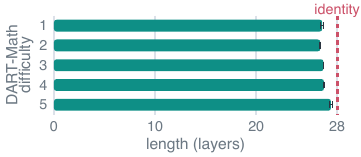}
\caption{\scriptsize\mbox{Qwen2.5-7B}}\end{subfigure}%
\par\vspace{4pt}
\begin{subfigure}{0.2956\linewidth}\centering
\includegraphics[width=1.0000\linewidth]{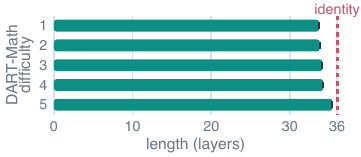}
\caption{\scriptsize\mbox{Qwen3-8B}}\end{subfigure}%
\hspace{26.0pt}%
\begin{subfigure}{0.2956\linewidth}\centering
\includegraphics[width=1.0000\linewidth]{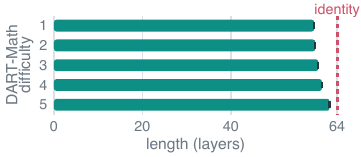}
\caption{\scriptsize\mbox{Qwen3-32B}}\end{subfigure}%
\hspace{26.0pt}%
\begin{subfigure}{0.2956\linewidth}\centering
\includegraphics[width=1.0000\linewidth]{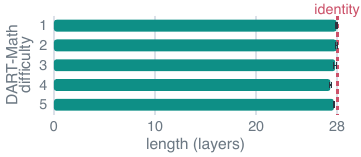}
\caption{\scriptsize\mbox{LLaMA-3.2-3B}}\end{subfigure}
\caption{\hyperref[finding:2]{\ftag{F2}}\ On the four Qwen2.5 and Qwen3 models, the shortest valid program is shorter than the standard forward pass on average at every difficulty, and longest at DM-5; on Qwen1.5-MoE-A2.7B and LLaMA-3.2-3B it is about as long.
Program-length counterpart of \cref{fig:appendix_shorteridentity_diff_rate}: the shortest valid program's executed length, per model and DART-Math difficulty. The dotted line marks the identity (standard forward pass) length $D$. One panel per model. Rows are the difficulty, 1 (easiest) to 5 (hardest). Bars are the mean over the three search runs, whiskers are $\pm$ 1 std.}
\label{fig:appendix_shorteridentity_diff_len}
\end{figure}

\begin{figure}[htbp]
\centering
\footnotesize
\legenditem{polBaseline}{baseline (identity)}\quad\legenditem{polRescued}{rescued}\quad\legenditem{polUnsolved}{unsolved}\\[0.3em]
\normalsize
\begin{subfigure}{0.2103\linewidth}\centering
\includegraphics[width=1.0000\linewidth]{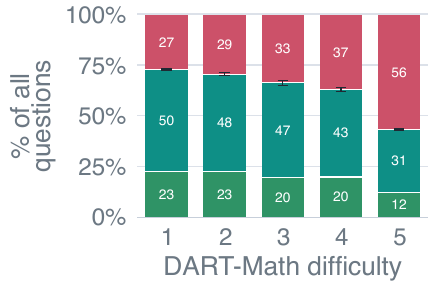}
\caption{\scriptsize\mbox{Qwen1.5-MoE-A2.7B}}\end{subfigure}%
\hspace{5.8pt}%
\begin{subfigure}{0.1442\linewidth}\centering
\includegraphics[width=1.0000\linewidth]{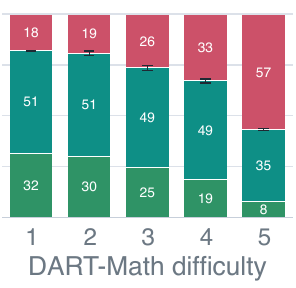}
\caption{\scriptsize\mbox{Qwen2.5-3B}}\end{subfigure}%
\hspace{5.8pt}%
\begin{subfigure}{0.1442\linewidth}\centering
\includegraphics[width=1.0000\linewidth]{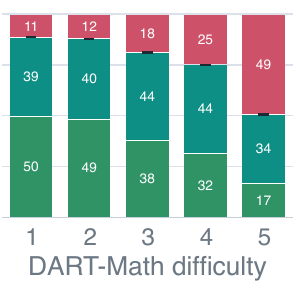}
\caption{\scriptsize\mbox{Qwen2.5-7B}}\end{subfigure}%
\hspace{5.8pt}%
\begin{subfigure}{0.1442\linewidth}\centering
\includegraphics[width=1.0000\linewidth]{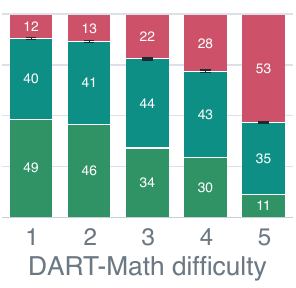}
\caption{\scriptsize\mbox{Qwen3-8B}}\end{subfigure}%
\hspace{5.8pt}%
\begin{subfigure}{0.1442\linewidth}\centering
\includegraphics[width=1.0000\linewidth]{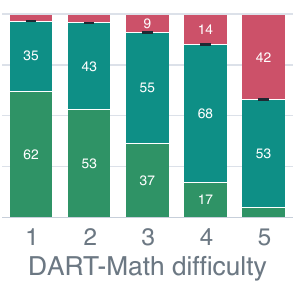}
\caption{\scriptsize\mbox{Qwen3-32B}}\end{subfigure}%
\hspace{3.8pt}%
\begin{subfigure}{0.1526\linewidth}\centering
\includegraphics[width=0.9449\linewidth]{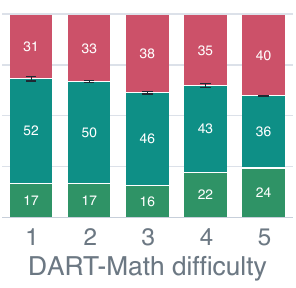}
\caption{\scriptsize\mbox{LLaMA-3.2-3B}}\end{subfigure}
\vspace{-4pt}
\caption{At every difficulty, the search rescues many of the questions the standard pass gets wrong, but the unsolved share grows with difficulty.
The baseline/rescued/unsolved split of the questions.
Of every question: the share already answered correctly by the standard forward pass (baseline); the share the standard forward pass got wrong but for which the search found at least one valid program (rescued); the share for which nothing valid was found (unsolved). One panel per model, by DART-Math difficulty (1 easiest, 5 hardest); the y axis is shown on the left-most panel only. Numbers are shares in \%, shown for blocks of 8\% or more. Baseline does not depend on the search; rescued is the mean over the three search runs, and its whisker is $\pm$ 1 std (the unsolved share is what remains, so it moves by the same amount).}
\label{fig:appendix_rescue}
\end{figure}

\begin{figure}[htbp]
\centering
\footnotesize
\legenditem{polIdentityGray}{identity}\quad\legenditem{polSkip}{skip-only}\quad\legenditem{polRepeat}{repeat-only}\quad\legenditem{polBoth}{both}\\[0.3em]
\normalsize
\begin{subfigure}{0.2103\linewidth}\centering
\includegraphics[width=1.0000\linewidth]{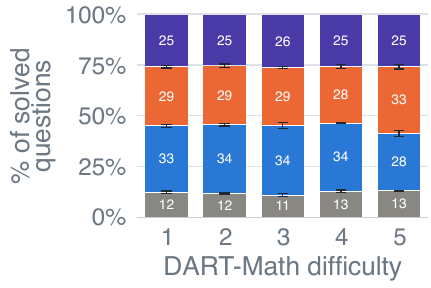}
\caption{\scriptsize\mbox{Qwen1.5-MoE-A2.7B}}\end{subfigure}%
\hspace{5.8pt}%
\begin{subfigure}{0.1442\linewidth}\centering
\includegraphics[width=1.0000\linewidth]{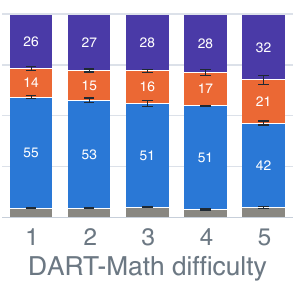}
\caption{\scriptsize\mbox{Qwen2.5-3B}}\end{subfigure}%
\hspace{5.8pt}%
\begin{subfigure}{0.1442\linewidth}\centering
\includegraphics[width=1.0000\linewidth]{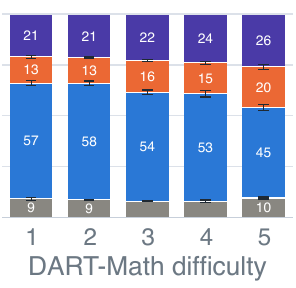}
\caption{\scriptsize\mbox{Qwen2.5-7B}}\end{subfigure}%
\hspace{5.8pt}%
\begin{subfigure}{0.1442\linewidth}\centering
\includegraphics[width=1.0000\linewidth]{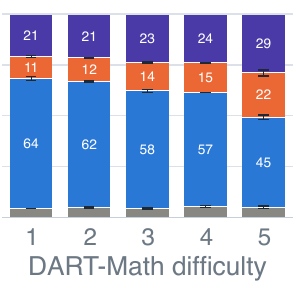}
\caption{\scriptsize\mbox{Qwen3-8B}}\end{subfigure}%
\hspace{5.8pt}%
\begin{subfigure}{0.1442\linewidth}\centering
\includegraphics[width=1.0000\linewidth]{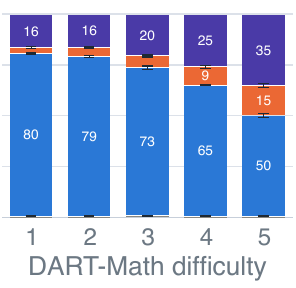}
\caption{\scriptsize\mbox{Qwen3-32B}}\end{subfigure}%
\hspace{3.8pt}%
\begin{subfigure}{0.1526\linewidth}\centering
\includegraphics[width=0.9449\linewidth]{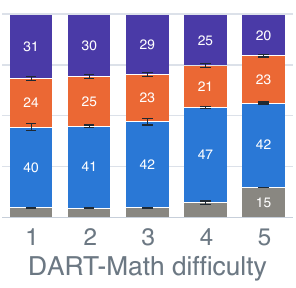}
\caption{\scriptsize\mbox{LLaMA-3.2-3B}}\end{subfigure}
\vspace{-4pt}
\caption{\hyperref[finding:1]{\ftag{F1}}\ The shortest valid program most often only skips, except on Qwen1.5-MoE-A2.7B at DM-5; on the Qwen models, programs that repeat become more common at harder levels.
The solvability split by operation class. Each solved question is counted once, by its shortest valid program: the standard forward pass already works (identity), the program uses only skips, only repeats, or both. The four shares add up to 100\%. One panel per model, by DART-Math difficulty (1 easiest, 5 hardest); the y axis is shown on the left-most panel only. Bars are the mean over the three search runs, whiskers are $\pm$ 1 std of the cumulative share up to that edge. Numbers are shares in \%, shown for blocks of 8\% or more.}
\label{fig:appendix_opclass}
\end{figure}

\begin{figure}[htbp]
\centering
\begin{subfigure}{0.4908\linewidth}\centering
\includegraphics[width=1.0000\linewidth]{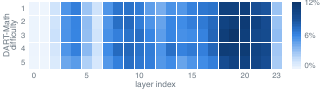}
\caption{Qwen1.5-MoE-A2.7B}\end{subfigure}%
\hspace{5.0pt}%
\begin{subfigure}{0.4607\linewidth}\centering
\includegraphics[width=1.0000\linewidth]{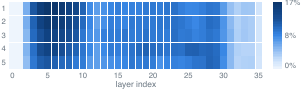}
\caption{Qwen2.5-3B}\end{subfigure}%
\\[0.4em]%
\begin{subfigure}{0.4908\linewidth}\centering
\includegraphics[width=1.0000\linewidth]{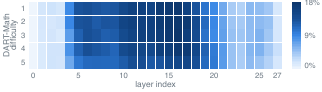}
\caption{Qwen2.5-7B}\end{subfigure}%
\hspace{5.0pt}%
\begin{subfigure}{0.4607\linewidth}\centering
\includegraphics[width=1.0000\linewidth]{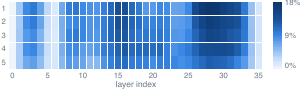}
\caption{Qwen3-8B}\end{subfigure}%
\\[0.4em]%
\begin{subfigure}{0.4908\linewidth}\centering
\includegraphics[width=1.0000\linewidth]{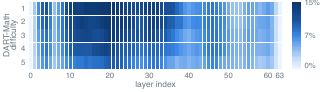}
\caption{Qwen3-32B}\end{subfigure}%
\hspace{5.0pt}%
\begin{subfigure}{0.4607\linewidth}\centering
\includegraphics[width=1.0000\linewidth]{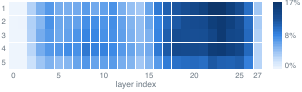}
\caption{LLaMA-3.2-3B}\end{subfigure}
\caption{\hyperref[finding:3]{\ftag{F3}}\ The earliest layers are almost never skipped in any model, consistent with the early load-bearing region in \cref{fig:motivation}.
Share of shortest-valid programs that skip each layer, by layer index and DART-Math difficulty. Each cell is the mean over the three search runs; each panel has its own color scale.}
\label{fig:appendix_layerops_skip}
\end{figure}

\begin{figure}[htbp]
\centering
\begin{subfigure}{0.4851\linewidth}\centering
\includegraphics[width=1.0000\linewidth]{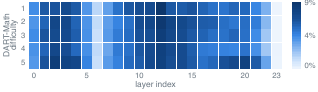}
\caption{Qwen1.5-MoE-A2.7B}\end{subfigure}%
\hspace{5.0pt}%
\begin{subfigure}{0.4550\linewidth}\centering
\includegraphics[width=1.0000\linewidth]{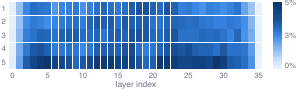}
\caption{Qwen2.5-3B}\end{subfigure}%
\\[0.4em]%
\begin{subfigure}{0.4851\linewidth}\centering
\includegraphics[width=1.0000\linewidth]{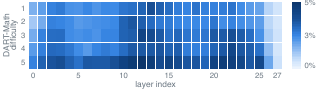}
\caption{Qwen2.5-7B}\end{subfigure}%
\hspace{5.0pt}%
\begin{subfigure}{0.4550\linewidth}\centering
\includegraphics[width=1.0000\linewidth]{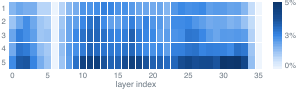}
\caption{Qwen3-8B}\end{subfigure}%
\\[0.4em]%
\begin{subfigure}{0.4851\linewidth}\centering
\includegraphics[width=1.0000\linewidth]{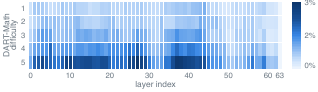}
\caption{Qwen3-32B}\end{subfigure}%
\hspace{5.0pt}%
\begin{subfigure}{0.4550\linewidth}\centering
\includegraphics[width=1.0000\linewidth]{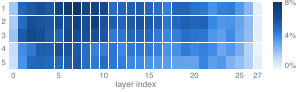}
\caption{LLaMA-3.2-3B}\end{subfigure}
\caption{\hyperref[finding:3]{\ftag{F3}}\ Only on Qwen3-32B do harder levels spread repeats from a few layers to almost the whole stack; the other models repeat across nearly all layers at every level.
The repeat-side counterpart to \cref{fig:appendix_layerops_skip}. Share of shortest-valid programs that repeat each layer, by layer index and DART-Math difficulty. Each cell is the mean over the three search runs; each panel has its own color scale.}
\label{fig:appendix_layerops_repeat}
\end{figure}

This subsection collects additional analyses with no equivalent in \gls{polar}'s
paper.
\Cref{fig:menu_topk} tracks the menu's pass@$K$ as the candidate menu size $K$ grows;
\cref{fig:appendix_shorteridentity_diff_rate,fig:appendix_shorteridentity_diff_len} give
the per-model share and length of shorter-than-identity programs by difficulty, the full detail behind \hyperref[finding:2]{\ftag{F2}}.
\Cref{fig:appendix_rescue,fig:appendix_opclass} give two further results not in the main
body: per-model, per-difficulty breakdowns of the same search output that
\cref{sec:analysis} discusses only at the pooled level, how much of each difficulty is
already solved, rescued, or left unsolved, and which operation class (skip-only,
repeat-only, or both) the working programs use.
\Cref{fig:appendix_layerops_skip,fig:appendix_layerops_repeat} extend
\cref{fig:main_layerops_qwen3_32b}'s per-layer skip/repeat analysis to all six
models.
The single-segment-reversion check behind \hyperref[finding:7]{\ftag{F7}} (\cref{fig:op_transition_robustness}) covers all $30$ model/difficulty combinations:
per search run, about $21{,}900$ of the roughly $317{,}000$ rescuing programs have a reverted variant that the search evaluated;
for about $3{,}700$ of them it is identity (a single-edit program), so about $18{,}200$ programs are tested.

\FloatBarrier

\section{\textsc{PoLar} on MMLU-Pro}
\label{sec:appendix_mmlu_letter_bias}
\glsresetall

MMLU-Pro's own answer keys \citep{wang2024mmlupro} carry no real letter bias.
Conditioned on \texttt{num\_options}, the dominant group in our split ($6{,}542$ of $7{,}845$ train rows, $83\%$, all with $\texttt{num\_options}=10$) is close to uniform (per-letter frequency $9.3$--$10.6\%$ against a $10.0\%$ random baseline)\label{fact:mmlu_dataset_letter}.
We find no discrete-letter bias for any \texttt{num\_options} bucket we checked ($3$--$10$).
So any bias we observe below is a property of the model or its \gls{mcts}-discovered programs, not the data.
The rest of this section gives the full evidence that many \gls{mcts}-discovered programs on this benchmark answer correctly for a specific option arrangement rather than a specific question, and collapse when that arrangement changes (\cref{sec:appendix_mmlu_reshuffle_original,sec:appendix_mmlu_router_random_baseline}).

\subsection{Option-Order Robustness: \gls{mcts}-Discovered Programs Track Position, Not Content}
\label{sec:appendix_mmlu_reshuffle_original}

A program that has genuinely learned to answer a question should still get it right if the question's options are presented in a different order.
A program that has instead keyed on which letter or position the correct answer occupied should not.
We test this directly by re-scoring an \gls{mcts}-discovered program after randomly reshuffling its query's option order, leaving question content untouched, and comparing against the program's original-order accuracy.
This is one of our two main \gls{mcts}-on-MMLU-Pro findings: many \gls{mcts}-discovered programs on this benchmark are unable to survive an option reshuffle, meaning they have effectively overfit to the correct letter rather than solved the question.

We repeat the option-reshuffle check on Qwen3-8B's full search (all $12{,}032$ MMLU-Pro queries, no train/val/test subsetting), to confirm the finding below is not an artifact of subsetting to train/val/test splits.

\paragraph{Identity is close to shuffle-invariant; rescuing programs are not.}
As a control, we reshuffle option order for the \emph{entire} evaluation set (not just rescuing queries) and re-score under the unmodified identity program.
Accuracy is unchanged within noise ($42.2\%\rightarrow41.9\%$, $n=12{,}032$), even though $29.9\%$ of the originally correct answers flip to wrong and $21.3\%$ of the originally wrong ones flip to correct under the identical reshuffle.
Both are about $12\%$ of all questions, so they cancel.
This net-flat accuracy masks substantial per-query churn, consistent with ordinary decision noise rather than systematic position exploitation.
Programs discovered by \gls{mcts} specifically to rescue an identity failure behave completely differently under the same manipulation: on the rescuing population ($n=6{,}866$ questions, each tested with its shortest valid non-identity program; identity wrong, program correct, under the original arrangement), program accuracy collapses from $94.2\%$ to $17.7\%$, a $76.4$pp drop, not far above the uniform-random floor (roughly $10\%$ for this dataset's dominant $10$-option questions, \cref{fact:mmlu_dataset_letter}).
The original arrangement does not score exactly as in the search, because we re-score it in a fresh run, where batch-dependent bf16 noise flips some of the search's labels (\cref{sec:appendix_reward_noise}): the programs reach $94.2\%$ rather than $100\%$, and identity, wrong on all these questions during the search, now answers $0.4\%$ of them correctly.
This noise is small next to the drop under reshuffling.

\paragraph{The fragility is not exclusive to rescuing programs, but rescuing programs are markedly worse.} Repeating the check on every query with any valid non-identity program ($n=11{,}938$, $99.2\%$ of the dataset, almost the whole benchmark) and splitting by whether identity itself was already correct: programs found where identity already succeeds are also reshuffle-fragile ($96.8\%\rightarrow36.6\%$, $-60.2$pp) but recover well above the random-guess floor, while rescuing programs collapse further still, matching the rescuing-only measurement above.
Reshuffle-fragility is therefore a general property of \gls{mcts}-discovered non-identity programs here, not an artifact of the rescue mechanism specifically, though programs built to correct an identity failure lean on the original option arrangement hardest.
In short, the base model is close to shuffle-invariant, but every population of \gls{mcts}-discovered non-identity programs we measure tracks the arrangement a question happened to be presented in, not what it is actually asking.

\subsection{Router Reproduction on MMLU-Pro: A Random-Valid-Program Control}
\label{sec:appendix_mmlu_router_random_baseline}

\Cref{sec:router_collapse_results} documents the router's pass@$1$ identity-collapse on DART-Math and yet a genuine pass@$5$ rescue signal underneath it.
We separately trained and evaluated a router directly on MMLU-Pro, on the $7{,}845$ train and $1{,}387$ validation questions and a $2{,}800$-question held-out test split, to test whether the position-overfitting result in \cref{sec:appendix_mmlu_reshuffle_original} predicts a parallel router-level failure: if what \gls{mcts} labels as ``correct'' on this benchmark is disproportionately a position artifact rather than a stable function of question content, a router trained on that supervision has little content-based signal left to generalize from, and should not be expected to rank real candidates any better than chance.
The split is per domain: $200$ test questions from each of the $14$ domains are drawn first (seed $42$), and the rest is split $85$/$15$ into train and validation, so the three splits together are the whole benchmark.

\paragraph{Pass@$1$ collapses onto identity, exactly as on DART-Math.} Across every recipe we swept (down-weighting, class-balanced focal loss, length-preference reweighting, Drop/Crop repeat-segment parsing, \cref{sec:appendix_router_recipe}), the router's top-$1$ prediction is the identity program for every one of the $2{,}800$ held-out test queries, with identity accuracy $43.9\%$.
This matches \cref{sec:router_collapse_results}'s DART-Math finding and confirms it is not an artifact specific to that dataset's answer format.

\paragraph{A pass@$5$ margin over identity survives, but so does an equally large margin for random guessing.}
Pass@$5$ for Drop-CE (the recipe we use on DART-Math, \cref{sec:appendix_router_recipe}) reaches $62.5\%$, above the $43.9\%$ identity floor (the rest of our recipe sweep lands lower, $53$--$57\%$; \cref{sec:appendix_router_recipe} covers those recipes).
But on a forced-choice benchmark with as few as $3$--$10$ options, trying several candidates and taking the best can look like a real gain with zero learned signal, so we build a control: $5$ programs sampled uniformly from the router's own action grammar, independent of training, and applied as one fixed menu to every test question, graded identically, repeated over $5$ seeds.
This control averages $45.3\%\pm8.8\%$ at pass@$5$ ($17.0\%\pm9.7\%$ at pass@$1$).
Only Drop-CE clears it, by roughly two standard deviations.
That margin, though, does not behave like a learned one.
Retraining Drop-CE's winning configuration $5$ independent times (weight-init/data-order seed only) reproduces \emph{exactly} the same pass@$5$ figure ($62.5\%$ to full floating-point precision) every time, because every retraining converges on the same menu of $5$ programs,
which it proposes for all test queries in only varying order (\cref{tab:mmlu-router-random-baseline}),
despite the $5$ checkpoints having substantially different weights, validation loss ($1.0920$--$1.0953$), and pre-decode logits.
If the router's non-identity candidates tracked each question's content, different initializations should disagree at least occasionally on which candidates to propose; that five runs with genuinely different weights converge on one menu instead points to a fixed, low-content-dependence attractor of training, not evidence the router has learned to read questions.
The menu consists of identity and skipping one of layers 20--23, 24--27, 28--31 or 32--35.
It is the same in all $14$ domains, with identity always ranked first and skipping layers 32--35 almost always second.
Three of the four skips answer a near-constant letter (\cref{tab:mmlu-router-menu-letters}):
skipping 20--23 answers C on $76\%$ of the questions, 24--27 F on $59\%$, and 28--31 A on $74\%$, in every domain, while the gold letters are close to uniform ($12.8\%$ to $7.8\%$).
They keep their letter when the options are reshuffled,
while skipping 32--35 behaves like identity.
Identity plus four fixed letters (A--D), without any layer edit, matches, and slightly exceeds, the router's pass@$5$ ($62.8\%$ vs.\ $62.5\%$).
The margin over random therefore comes from trying several fixed answer letters, not from a per-question, content-aware prediction.

\begin{table}[tb]
\centering
\small
\caption{The router beats a control of $5$ random valid programs at pass@$5$ ($62.5\%$ vs.\ $45.3\%\pm8.8\%$), but identically in all $5$ retrainings, which converge on the same $5$-program menu.
Qwen3-8B, $2{,}800$-question held-out test split, online execution.
Random baseline: one fixed menu of $5$ programs, drawn independently of training, $5$ draws (mean$\pm$std).
Router: Drop-CE's best configuration retrained with $5$ seeds (initialization and data order); values without $\pm$ are identical across all $5$ retrainings.}
\label{tab:mmlu-router-random-baseline}
\begin{tabular}{lccccc}
\toprule
& pass@$1$ & pass@$2$ & pass@$3$ & pass@$4$ & pass@$5$ \\
\midrule
Identity & $43.9\%$ & -- & -- & -- & -- \\
Random baseline & $17.0\%\pm9.7\%$ & $25.9\%\pm8.8\%$ & $36.1\%\pm6.4\%$ & $40.4\%\pm7.8\%$ & $45.3\%\pm8.8\%$ \\
\Router\ (Drop-CE) & $43.9\%$ & $50.1\%$ & $54.4\%\pm0.9\%$ & $58.5\%$ & $62.5\%$ (identical, all 5) \\
\bottomrule
\end{tabular}
\end{table}

\begin{table}[tb]
\centering
\small
\caption{Three of the four skips in the router's menu answer a near-constant letter, and identity plus four fixed letters matches, and slightly exceeds, the router's pass@$5$ ($62.8\%$ vs.\ $62.5\%$).
What the $5$ programs of the MMLU-Pro router's fixed menu answer. Qwen3-8B, $2{,}800$-question held-out test split, same grading as the router's pass@$5$.
Rescues: questions identity gets wrong and the program right.
Reshuffled: options reshuffled as in \cref{sec:appendix_mmlu_reshuffle_original}; share of questions on which the program answers the same letter as before, and on which it picks the same option content.
With the four letters the skips predict most (A, C, F, H) instead of the four most common gold letters (A--D), it reaches $64.1\%$.}
\label{tab:mmlu-router-menu-letters}
\begin{tabular}{lccccc}
\toprule
& Most frequent & & & \multicolumn{2}{c}{Reshuffled} \\
\cmidrule(lr){5-6}
Program & letter (share) & Accuracy & Rescues & Same letter & Same content \\
\midrule
Identity & A ($30.5\%$) & $43.9\%$ & -- & $25.3\%$ & $48.9\%$ \\
\textsf{skip} 20--23 & C ($75.9\%$) & $13.1\%$ & $124$ & $76.5\%$ & $13.1\%$ \\
\textsf{skip} 24--27 & F ($59.4\%$) & $17.5\%$ & $203$ & $63.3\%$ & $15.1\%$ \\
\textsf{skip} 28--31 & A ($73.5\%$) & $27.2\%$ & $117$ & $62.5\%$ & $21.1\%$ \\
\textsf{skip} 32--35 & A ($30.6\%$) & $43.5\%$ & $175$ & $23.7\%$ & $46.5\%$ \\
\bottomrule
\end{tabular}
\end{table}

Read against \cref{sec:appendix_mmlu_reshuffle_original}, this closes the loop the option-order finding opens.
\gls{mcts} supervision on MMLU-Pro is dominated by the answer's position rather than its content, and a router trained on it does not recover the content either.
Its pass@$5$ margin is real and reproducible, but it comes from a fixed menu, not from per-question choices.
This is the failure the reshuffle result predicts, since a multiple-choice format gives \gls{mcts}-discovered programs a shortcut, the answer's position, that they can overfit to instead of the question's content.

\section{Further Brain-Inspired Machine Learning Approaches}
\label{sec:appendix_brain_ml}
\glsresetall

\textbf{Modular specialization and global coordination.}
Sparse mixture-of-experts (MoE) architectures introduce conditional computation by routing each token representation to only a subset of feed-forward experts within a layer:
selective, but locally scoped, since it does not alter the model's overall layer-level execution path.
Mixture of Cognitive Reasoners (MiCRo) extends the idea of specialization by organizing expert modules around distinct cognitive networks \citep{alkhamissi2025micro}.
Programs-of-layers instead condition on the whole execution path,
which layers run at all and how many times,
rather than which experts fire within a fixed layer sequence.
\citet{goyal2021globalworkspace} propose a complementary piece:
a shared, bandwidth-limited communication channel through which otherwise-specialized neural modules coordinate,
inspired by global workspace theory.
We invoke this same idea in \cref{sec:cortical_analogies}: a program-of-layers' path-level coordination, deciding a sequence of engagements over the whole forward pass, is closer in spirit to this global channel than to MoE's per-token, per-layer routing.
MoE routing stays local and never sees the full path at once.

\textbf{Brain mechanisms as design inspiration.}
\citet{hawkins2019gridcells} argue that grid-cell-like location representations, neurons that encode spatial position, run throughout the neocortex and that individual cortical columns learn complete models of objects by combining external sensory input with this location signal.
Their picture of many largely independent columns, each building its own model rather than passing everything through one fixed hierarchy, is consistent with program skipping or repeating layers per input (\cref{sec:analysis}).
Unlike \citet{li2026polar}, \citeauthor{hawkins2019gridcells} invite building computationally based cortical columns rather than routing inputs over existing LLM architectures.
\citet{darlow2025ctm} build neural synchronization and neuron-level temporal processing directly into the architecture (the Continuous Thought Machine),
and show it supports adaptive computation, allocating more processing to harder inputs.
This is a fundamental biology-based architecture implementation that differs from \citeauthor{li2026polar},
since \citeauthor{li2026polar} reuse a standard transformer's existing layers and change only \emph{when} and \emph{how often} they fire,
rather than changing what a unit computes internally.
\citet{adeel2026mentalstate} link biophysical models of pyramidal neurons (a common type of cortical neuron) to modulation loops among attention components,
and report faster learning at roughly linear complexity on a standard vision-transformer benchmark.
Like the Continuous Thought Machine, this operates at a finer grain than \citeauthor{li2026polar}'s layer-level programs,
but it is further evidence that attention specifically, not just depth or routing, is an active target for brain-inspired redesign.

\textbf{Measuring the analogy, and borrowing brain subsystems rather than whole architectures.}
\citet{shah2025mot} introduce Multi-Level Optimal Transport (MOT),
a framework that jointly aligns layers and individual units across networks of different depth,
and apply it to align model layers not only against each other but directly against human visual-cortex recordings.
Unlike our own analogy in \cref{sec:cortical_analogies}, which stays at the level of a coordinating role rather than a specific mapping,
MOT gives a concrete recipe for actually testing whether the layers a program-of-layers revisits align with specific cortical regions,
an empirical test we only gesture at here, not one we run.
\citet{tran2025tmcl} use top-down modulation, in the style of predictive coding, a theory that the brain constantly predicts its own input and corrects for the difference, to consolidate representations across tasks from very little labeled data,
motivated by cortical feedback circuits.
The connection to \citeauthor{li2026polar}'s work is looser here,
and we include it as a further example of the same general pattern:
cortical feedback as a design principle for a \gls{ml} system, rather than as a direct precedent.
\citet{gutierrez2024hipporag} structure a retrieval-augmented \gls{llm}'s long-term memory after the hippocampal-neocortical memory system with their method, HippoRAG,
improving multi-hop retrieval accuracy and speed over iterative retrieval baselines.
Memory is orthogonal to the execution-path question this paper asks,
but the move is the same: borrow a specific, well-characterized brain system's computational structure, not just its name, as a design template.
\citet{berges2024memorylayers} scale a complementary, non-brain-inspired mechanism, their \emph{memory layers} (sparse trainable key-value memory), to large models.
We include their work as a contrast case: it treats memory as a separate subsystem too, but without any neuroscience motivation, showing the same architectural idea can arise independently of brain inspiration.

\section{Communication with the Original Authors}
\label{sec:author_communication}
\glsresetall

Reproducing \citet{li2026polar} was difficult throughout.
The released repository\footnote{\url{https://github.com/tianyi-lab/PoLar}} originally covered only router training and evaluation given precomputed \gls{mcts} programs for supervision, which are themselves not released.
Three things therefore had to be reconstructed from underspecified prose rather than checked against a reference implementation: the search that produces that supervision (described only as a seven-line pseudocode box), the DM-1--DM-5 difficulty split every reported table is organized around, and the code that produced the paper's own baseline tables (\cref{sec:appendix_mcts_design,sec:appendix_dart_math,sec:baseline_reproduction}).

Our attempts to contact \citeauthor{li2026polar} to inquire about the then-unpublished \gls{mcts} components required to reproduce their work (emails on July 23 and August 18, 2026) were unsuccessful.
Correspondence with the original authors could have resolved many of the reconstruction challenges quickly;
instead, we undertook an independent reconstruction and validation effort, documented throughout this paper's appendix.

\ifdeanonymized
After we asked for the missing search code in a public comment on their OpenReview page\footnote{\url{https://openreview.net/forum?id=pl10b6EQAN}}, \citeauthor{li2026polar} released their \gls{mcts} code,
\else
\citeauthor{li2026polar} later released their \gls{mcts} code,
\fi
still without its search results or the code for their baselines and \gls{ood} evaluation.
We now port their search, check it program for program against their code, and run it on our reconstruction of their split (\cref{sec:appendix_their_split}).
\Cref{sec:appendix_mcts_design} explains why our main results keep our own search.

\section{Use of Large Language Models}
\label{sec:appendix_ai_use}

The authors used general-purpose \gls{llm} coding assistants (Claude Code) during development: writing and debugging experiment code in the companion repository, and drafting/editing prose and \LaTeX{} in this paper.
All generated code and text were reviewed and edited by the authors before use.
All reported numbers, claims, and conclusions are the authors' own and were verified against the underlying experiment outputs.
Separately, an open-weight \gls{llm} (\texttt{MiniMax-M2.7-AWQ}) serves as a tool in our analysis, labeling how close wrong answers are to the ground truth;
its labels are not checked against human labels, but \cref{sec:appendix_bht_large} describes a check of its competence as a solver.

\end{document}